\documentclass[11pt,letterpaper, onecolumn]{article}
\usepackage[labelfont=bf,labelsep=period]{caption}
\usepackage{subcaption}
\usepackage{fullpage}
\usepackage{times}
\usepackage[percent]{overpic} 
\usepackage{array}
\usepackage{ragged2e}
\usepackage{multirow}
\usepackage{soul}

\usepackage{stackengine}    
\usepackage{diagbox}        
\usepackage{booktabs}       
\usepackage{fancyhdr, amssymb, mathtools}
\usepackage[ruled,vlined]{algorithm2e}
\usepackage[dvipsnames]{xcolor}
\usepackage[margin = 1 in]{geometry}
\usepackage{graphicx}
\usepackage{outlines}
\usepackage{amsmath}

\DeclareMathOperator*{\argmin}{argmin}
\usepackage{graphicx}
\usepackage{dirtytalk}

\usepackage{xcolor}
\usepackage{color}

\usepackage{multirow}
\usepackage{arydshln}

\usepackage[english]{babel} 
\usepackage{blindtext}
\usepackage{amsfonts}
\usepackage{amssymb}
\usepackage{bbm}
\usepackage{bm}
\usepackage{upgreek}
\usepackage{mathtools}

\usepackage{xltabular}
\usepackage{wrapfig}
\usepackage{enumitem}
\usepackage{floatrow}
\usepackage{xfrac}
\usepackage[font={footnotesize}, labelfont={bf}, labelsep=space, justification=justified, skip=0pt]{caption}
\usepackage[hang,flushmargin]{footmisc} 
\usepackage[colorlinks=true, allcolors=blue]{hyperref}

\usepackage{diagbox} 
\usepackage{multirow} 
\usepackage{cleveref} 
\Crefname{figure}{Fig.}{Figs.}

\newcommand{\cmt}[1]{} 

\newcommand{\fb}{\boldsymbol{f}}

\newcommand{\mb}{\boldsymbol{m}}
\newcommand{\nb}{\boldsymbol{n}}
\newcommand{\rb}{\boldsymbol{r}}
\newcommand{\tb}{\boldsymbol{t}}
\newcommand{\ub}{\boldsymbol{u}}

\newcommand{\xb}{\boldsymbol{x}}

\newcommand{\zb}{\boldsymbol{z}}

\newcommand{\Ab}{\boldsymbol{A}}

\newcommand{\Cb}{\boldsymbol{C}}

\newcommand{\Fb}{\boldsymbol{F}}

\newcommand{\Ub}{\boldsymbol{U}}
\newcommand{\Vb}{\boldsymbol{V}}

\newcommand{\Xb}{\boldsymbol{X}}

\newcommand{\epsilonb}{\boldsymbol{\varepsilon}}
\newcommand{\sigmab}{\boldsymbol{\sigma}}
\newcommand{\thetab}{\boldsymbol{\theta}}

\newcommand{\rhob}{\boldsymbol{\rho}}
\newcommand{\phib}{\boldsymbol{\phi}}

\newcommand{\lambdab}{\boldsymbol{\lambda}}
\newcommand{\Sigmab}{\boldsymbol{\Sigma}}

\newcommand{\zetab}{\boldsymbol{\zeta}}

\newcommand{\niKb}{\boldsymbol{\mathrm{K}}} 
\newcommand{\niUb}{\boldsymbol{\mathrm{U}}} 
\newcommand{\niFb}{\boldsymbol{\mathrm{F}}} 
\newcommand{\niN}{\mathrm{N}} 
\newcommand{\niub}{\boldsymbol{\mathrm{u}}} 

\newcommand{\nirho}{\uprho}
\newcommand{\nipsi}{\uppsi} 

\newcommand{\niV}{\mathrm{V}} 

\DeclareMathOperator{\swish}{swish}

\DeclareMathOperator{\Res}{Res}
\DeclareMathOperator{\Filter}{Filter}

\usepackage{titling}
\thanksheadextra{}{}
\thanksheadextra{}{} 
\usepackage{lipsum} 

\usepackage{amsmath}
 
\usepackage{natbib}

\title{Compliance Minimization via Physics-Informed Gaussian Processes}
\date{\vspace{-5ex}}
\usepackage{authblk}

\author[1]{Xiangyu Sun}
\author[1]{Amin Yousefpour}
\author[1]{Shirin Hosseinmardi}
\author[1,2]{Ramin Bostanabad \thanks{ Corresponding Author: raminb@uci.edu \\\href{https://github.com/Bostanabad-Research-Group/}{GitHub Repository}}}
\affil[1]{Department of Mechanical and Aerospace Engineering, University of California, Irvine}
\affil[2]{Department of Civil and Environmental Engineering, University of California, Irvine}

\begin{document}
    \pagenumbering{arabic}
    \sloppy
    \maketitle
\noindent \textbf{Abstract}\\

\noindent  
Machine learning (ML) techniques have recently gained significant attention for solving compliance minimization (CM) problems. However, these methods typically provide poor feature boundaries, are very expensive, and lack a systematic mechanism to control the design complexity. Herein, we address these limitations by proposing a mesh-free and simultaneous framework based on physics-informed Gaussian processes (GPs). 
In our approach, we parameterize the design and state variables with GP priors which have independent kernels but share a multi-output neural network (NN) as their mean function. The architecture of this NN is based on Parametric Grid Convolutional Attention Networks (PGCANs) which not only mitigate spectral bias issues, but also provide an interpretable mechanism to control design complexity. We estimate all the parameters of our GP-based representations by simultaneously minimizing the compliance, total potential energy, and residual of volume fraction constraint. Importantly, our loss function exclude all data-based residuals as GPs automatically satisfy them. We also develop computational schemes based on curriculum training and numerical integration to increase the efficiency and robustness of our approach which is shown to (1) produce super-resolution topologies with fast convergence, (2) achieve comparable compliance and relative less gray area fraction compared to traditional numerical methods, (3) provide control over fine-scale features, and (4) outperform competing ML-based methods. 


\section{Introduction} \label{sec intro}

Topology optimization (TO) offers a powerful tool to determine the optimal material distribution and geometric features within a design domain by minimizing a specific objective function under a set of constraints. Since the pioneering work by \citet{bendsoe_generating_1988}, TO has gained significant attention across diverse applications in structural mechanics, fluid dynamics, and materials science \citep{bendsoe_optimization_1995,borrvall_topology_2003,bendsoe_topology_2004,wang_comprehensive_2021}. Among various approaches, the solid isotropic material with penalization (SIMP) \citep{sigmund_99_2001,andreassen_efficient_2011,liu_efficient_2014} and level-set methods \citep{chen_topology_2017} are the most widely adopted for structural optimization problems. In recent years, several advanced TO methods have been introduced to address increasingly complex design scenarios including multi-scale formulations \citep{wu_topology_2021,zhao_design_2021,zhai_topology_2024}, local maximum stress constraints \citep{senhora_topology_2020}, and considerations of fracture or plasticity \citep{jia_controlling_2023}. Among various applications, compliance minimization (CM) represents a foundational application case that provides a benchmark for evaluating and advancing new structural optimization methods. Using the finite element method (FEM) for modeling structural deformation with a linear elastic material, the standard SIMP formulation for CM is given as:
\begin{subequations}\label{eq simp}
    \begin{align}
    \min_{\niub_e,\nirho_e} \quad & c\big(\niub_e,\nirho_e\big) = \sum_{e=1}^{\niN_e} (\nirho_e)^{p} \, \niub_e^{\text{T}} \niKb_0 \niub_e \label{eq simp obj} \\
    \text{subject to} \quad & \frac{\sum_{e=1}^{\niN_e} \nirho_e \niV_e}{\sum_{e=1}^{\niN_e} \niV_e} = \nipsi_f, \label{eq simp constr1} \\
    & \niKb \niUb = \niFb, \label{eq simp constr2} \\
    & \nirho_{\min} \leq \nirho_{e} \leq 1, \label{eq simp constr3}
    \end{align}
\end{subequations}
where $\nirho_e$, $\niub_e$, $\niKb_0$, $\niV_e$, and $\nipsi_f$\footnote{We use non-italic symbols to denote variables, vectors, and matrices in the SIMP formulation based on FEM, while italic symbols are used in our mesh-free framework for differentiation. denote the element density (design variable)}, nodal displacement vector, element stiffness matrix of the solid (i.e., phase $1$), element volume, and prescribed volume fraction, respectively. The volume fraction constraint and the force equilibrium condition are enforced through \Cref{eq simp constr1} and \Cref{eq simp constr2}, respectively. $\niKb$, $\niUb$ and $\niFb$ are the global stiffness matrix, global displacement vector, and global force vector, respectively. While optimizing element densities, a nonzero $\nirho_{\min}$ (typically set to $10^{-3}$) is imposed to avoid numerical instabilities in inverting $\niKb$. 

Most TO methods adopt a \textit{nested} framework to solve the optimization problem in \Cref{eq simp} by coupling an analysis step with a design update step. In the analysis step, the state variables (i.e., nodal displacements) are obtained by solving \Cref{eq simp constr2} using FEM and then used to compute the objective function in \Cref{eq simp obj} and the sensitivity field. The element densities are subsequently updated using gradient-based optimizers such as the optimality criteria (OC) method \citep{bendsoe_optimization_1995} or the method of moving asymptotes (MMA) \citep{svanberg_method_1987}. Under static and linear elastic conditions, the SIMP method enables straightforward and efficient implementations \citep{sigmund_99_2001,andreassen_efficient_2011} but it is prone to issues such as checkerboard patterns and mesh dependency which necessitate the use of filtering techniques such as gradient-based filters to regularize the solution. These filters, however, introduce some gray areas where element densities are not close to either 0 or 1.

Recent advancements in computational graphics have fueled the use of machine learning (ML) in TO \citep{shin_topology_2023}. These ML-based approaches leverage various types of neural networks (NNs) and can be broadly categorized into two groups: $(1)$ supervised data-driven approaches, and $(2)$ unsupervised approaches based on physics-informed machine learning (PIML). Representative data-driven approaches have demonstrated significant acceleration in reducing computational costs by predicting sensitivity fields using trained ML models \citep{chi_universal_2021,senhora_machine_2022}. Generative models that can design optimal topologies in a single forward pass also belong to this category \citep{yamasaki_data_driven_2021,sim_gans_2021,kallioras_mlgen_2021,nie_topologygan_2021}. However, these methods heavily rely on large datasets generated by conventional methods such as SIMP or level-set and the generalization ability of these methods strongly depends on the size and quality of their training samples which are difficult to generate in many cases. Hence in this work, we focus on unsupervised physics-informed approaches. 

PIML refers to a general class of ML models that are trained to surrogate a system's state variables while leveraging domain knowledge which can be in the form of boundary/initial conditions (BCs/ICs) and the governing partial differential equations (PDEs). These BCs/ICs and the PDEs are typically included as loss terms and the training aims to minimize the corresponding residuals. In solid mechanics, the weak form of a PDE system is typically adopted and this practice has led to the development of the deep energy method (DEM) which minimizes the potential energy functional as opposed to the strong form of the force balance equilibrium \citep{nguyen_hyper_DEM_2020,he_DEM_2023,hamel_calibrating_2023}. DEM has also been extended to accommodate nonlinear material behaviors and mixed formulations under large deformation scenarios \citep{nguyen_hyper_DEM_2020,niu_modeling_2023,he_DEM_2023,fuhg_mixed_2022}. Compared to PDE residuals, the variational weak form in DEM improves training stability and convergence rate.

Recent ML-based TO works have produced optimal topologies which are competitive to those obtained via SIMP. These works rely on one of the following three distinct strategies: 
$(1)$ methods that parameterize the density field using NNs while relying on FEM solvers for displacement computations \citep{hoyer_neural_2019,chandrasekhar_tounn_2021,chandrasekhar_approx_2022}; 
$(2)$ methods that estimate displacements using DEM while updating the density field via traditional optimizers such as OC or MMA \citep{he_deep_TO_2023,yin_dynamically_2024,zhao_physics_2024}; and 
$(3)$ fully ML-based frameworks that simultaneously parameterize displacement and density fields with separate NNs and update both minimizing an appropriately defined loss function whose derivatives can be easily obtained via automatic differentiation (AD) \citep{zehnder_Ntop_2021, joglekar_dmf_tonn_2024, jeong_complete_2023}. 
Although these three strategies differ methodologically, they all adopt a \textit{nested} optimization framework in which displacement fields are repeatedly computed using either the FEM or the DEM during each design iteration. This \textit{nested} structure leads to two potential issues: $(1)$ FEM-based nested frameworks inherit limitations similar to those of the traditional SIMP method such as mesh dependency and computational overhead from repeated linear system solves, and $(2)$ DEM-based methods are slow since the displacement field must be obtained by re-training the network (or part of it) at each design iteration. However, we argue that such a \textit{nested} configuration is not strictly necessary: the design update step minimizes compliance and the analysis step minimizes potential energy, so both objectives can be simultaneously optimized. Such a simultaneous approach has the potential to improve computational efficiency while removing mesh sensitivity and gray area. 

While ML-based TO frameworks have shown potential, their effectiveness heavily depends on the design of the NN architecture. The majority of existing works adopt feed-forward fully connected neural networks (also known as multilayer perceptrons or MLPs), which suffer from three major limitations:  
$(1)$ spectral bias, where the network preferentially learns low-frequency components and struggles to capture sharp gradients or fine-scale features such as stress concentrations \citep{cuomo_scientific_2022,kang_pixel_2023,deng_data_2025,chandrasekhar_approx_2022};
$(2)$ vanishing gradients in deep architectures, which reduces convergence rates and hinders effective loss minimization under the adjoint framework; and  
$(3)$ difficulty in strictly enforcing BCs and design constraints on the density field, where reliance on residual loss terms increases training complexity and distorts the optimization trajectory.

In this work we develop a simultaneous and mesh-free framework for CM based on physics-informed Gaussian processes (PIGP). In our framework, we parametrize the displacement and density variables using GP priors defined at a set of collocation points (CPs), with independent kernels but a shared multi-output mean function (the number of outputs equals the total number of design and state variables). We develop a physics-based loss function to estimate the parameters of our formulation and, in turn, obtain a continuous representation of the displacement and density fields. Our framework produces topologies with minimal gray areas and provides interpretable mechanisms for controlling the complexity of the designed topologies or enforcing design constraints.

We design the mean function of our parameterization via a unique NN architecture, i.e., a parametric grid convolution attention network (PGCAN). This architecture mitigates the spectral bias of MLPs and can capture solutions with high gradients (e.g., stress concentrations) which are critical to CM problems. We estimate the parameters of our PGCAN by minimizing an appropriately defined loss function that combines multiple terms related to the compliance, total potential energy, and design constraints (e.g., desired volume fraction). This minimization is conducted robustly and efficiently via a gradient-based optimizer that leverages the adjoint method as well as adaptive numerical approximations and curriculum training. Although our current PIGP framework is conceptually related to \cite{yousefpour_simultaneous_2025}, it incorporates key methodological advances (such as the PGCAN architecture, potential energy functional, adjoint-consistent gradients, and adaptive numerical strategies, etc.) in solving CM problems.

The remainder of this paper is organized as follows. We review the fundamentals of the DEM in \Cref{sec dem} and then introduce our approach in \Cref{sec method}. \Cref{sec results} presents numerical results across various examples and methods including our approach, SIMP, and other ML-based methods. \Cref{sec conclusion} concludes the paper and outlines potential directions for future research.

\section{Deep Energy Method} \label{sec dem}
The DEM offers an alternative to the FEM for solving displacement fields. For an isotropic linear elastic material in the absence of any body force, the governing PDE system is given by:
\begin{equation}\label{eq pde}
    \nabla \cdot \sigmab(\xb) = \mathbf{0}, \quad \forall \xb \in \Omega,
\end{equation}
where $\xb = [x,y]^{\text{T}}$ denotes an arbitrary point in the 2D domain $\Omega$ and $\sigmab$ denotes the Cauchy stress tensor. We assume the boundary of the domain $\partial\Omega$ is subjected to the following displacement and traction BCs:
\begin{subequations} \label{eq bc}
    \begin{align}
    u_i(\Xb_i) &= \tilde{\ub}_i, \quad \forall \Xb_i \in \partial \Omega_{\ub},\quad i = 1,2, \label{eq bc1} \\
    \sigmab(\Xb_F) \cdot \nb(\Xb_F) &= \Fb, \quad \forall \Xb_F \in \partial \Omega_{F}, \label{eq bc2}
    \end{align}
\end{subequations}
where $\partial \Omega_{\ub}$ and $\partial \Omega_F$ denote the portions of the boundary where displacement and traction BCs are applied, respectively, and the two-column matrices $\Xb_i$ and $\Xb_F$ represent sets of sampling points in $\partial \Omega_{\ub}$ and $\partial \Omega_{F}$, respectively. $\ub(\xb) = [u_1(\xb),u_2(\xb)]^{\text{T}}$ is the displacement vector and $\tilde{\ub}_i$ is the vector specifying the displacement BCs. $\nb(\xb) = [n_1(\xb), n_2(\xb)]$ denotes the surface normal vector with two components, and $\Fb$ is a matrix representing the external forces applied at $\Xb_F$. Assuming small deformations, the kinematic relationship between the strain tensor $\epsilonb(\xb)$ and $\ub(\xb)$ is given as:
\begin{equation}\label{eq kinematic}
    \epsilonb(\xb) = \frac{1}{2} \big( \nabla \ub(\xb) + \nabla \ub^{\text{T}}(\xb) \big),
\end{equation}
where $\nabla$ denotes the gradient operator defined as:
\begin{equation}\label{eq nabla}
    \nabla \ub(\xb) = 
    \begin{bmatrix}
        \dfrac{\partial u_1(\xb)}{\partial x} & \dfrac{\partial u_1(\xb)}{\partial y} \\
        \dfrac{\partial u_2(\xb)}{\partial x} & \dfrac{\partial u_2(\xb)}{\partial y}
    \end{bmatrix}.
\end{equation}
The linear elastic constitutive law in tensor form is written as:
\begin{equation}\label{eq constitutive}
    \sigmab(\xb) = \Cb : \epsilonb(\xb),
\end{equation}
where $\Cb$ is the constant fourth-order elastic stiffness tensor, parameterized by the Young's modulus $E$ and Poisson's ratio $\nu$ of the solid phase. 

\begin{figure*}[!h]
    \centering
    \includegraphics[width = 0.9\textwidth]{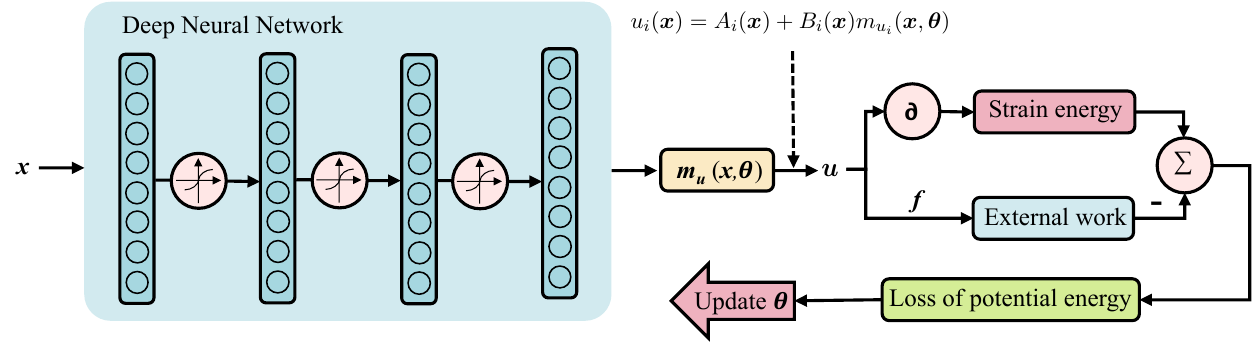}
    \caption{\textbf{Deep energy method:} The displacement field is obtained by minimizing the total potential functional in \Cref{eq pi}. Kinematic admissibility in this architecture is ensured via \Cref{eq ab}.}
    \label{fig dem}
\end{figure*}

DEM is based on the principle of minimum potential energy, which asserts that a solid body in force equilibrium adopts the deformation state that minimizes the total potential energy among all kinematically admissible configurations. The total potential energy $L_P\big(\ub(\xb)\big)$ is formulated as:
\begin{equation}\label{eq pi}
    L_P\big(\ub(\xb)\big) = \frac{1}{2} \int_{\Omega} \sigmab(\xb) : \epsilonb(\xb) \, dV - \int_{\partial \Omega_F} \fb^{\text{T}}(\xb) \ub(\xb) \, dA,
\end{equation}
where $\fb(\xb) = [f_1(\xb),f_2(\xb)]^{\text{T}}$ represents the external forces. The first term on the right-hand side of \Cref{eq pi} quantifies the strain energy and the second term accounts for the external work. So, instead of directly solving the strong form in \Cref{eq pde}, in the DEM method the displacement field is parameterized by an NN whose loss function is based on \Cref{eq pi}. This setup is schematically shown in \Cref{fig dem} with an MLP whose outputs are denoted by $\mb_{\ub}(\xb,\thetab) = [m_{u_1}(\xb,\thetab), m_{u_2}(\xb,\thetab)]^{\text{T}}$ where $\thetab$ denotes the trainable parameters. 

In DEM, to ensure that the solution is kinematically admissible (i.e., satisfies the displacement BCs), either an extra residual loss term is included in the loss function, or a transformation such as the following is employed:
\begin{equation}\label{eq ab}
    u_i(\xb) = A_i(\xb) + B_i(\xb) m_{u_i}(\xb,\thetab), i = 1,2,
\end{equation}
where $A_i(\xb)$ and $B_i(\xb)$ are smooth and differentiable functions that satisfy $A_i(\Xb_i) = \tilde{\ub}_i$ and $B_i(\Xb_i) = 0$ for all $\Xb_i \in \partial \Omega_{\ub}$ \citep{nguyen_hyper_DEM_2020}. Alternatively, $A_i(\xb)$ and $B_i(\xb)$ can be pre-trained NNs that satisfy the BCs \citep{rao_physics_2021}.
However, this approach has two limitations: (1) it is difficult to construct appropriate $A_i(\xb)$ and $B_i(\xb)$ for domains with irregular geometries, and (2) the choice of $A_i(\xb)$ and $B_i(\xb)$ is not unique, which may affect the final optimization results. To overcome these issues, we employ GPs as priors for the displacement fields in our framework, enabling BCs to be naturally enforced on domains with arbitrary shapes.

\section{Proposed Framework} \label{sec method}
TO determines the spatial material distribution that minimizes the scalar objective functional $L_C\big(\ub(\xb,\rho), \rho(\xb)\big)$ subject to a set of design constraints $C_i(\rho(\xb)) = 0$ for $i = 1,\dots,n_c$. We presume that the constraints imposed by the BCs are described by $R_b^{(i)}\big(\ub(\xb,\rho)\big)$ and that the underlying physics of the system are governed by a set of state equations $R_p^{(i)}\big(\ub(\xb,\rho), \rho(\xb)\big)$ where $\ub(\xb,\rho) = [u_1(\xb,\rho), u_2(\xb,\rho)]^{\text{T}}$ denotes the two displacement components which are functions of the spatial coordinate $\xb = [x,y]^{\text{T}}$ and the design variable $\rho(\xb)$. The general formulation of TO for CM can be written as:
\begin{subequations}\label{eq cm1}
\begin{align}
\ub\big(\xb,\rho\big), \,\rho(\xb)  &= 
\argmin_{\ub,\,\rho} 
L_C\big(\ub(\xb,\rho), \rho(\xb)\big) = 
\argmin_{\ub,\,\rho} 
\int_{\Omega} l_c\big(\ub(\xb,\rho), \rho(\xb)\big)\, dV, \label{eq cm1 a}
\intertext{subject to:}
C_1\big(\rho(\xb)\big) &= \int_{\Omega} \rho(\xb)\, dV - \psi_f V = 0, \label{eq cm1 b}
\\
C_i\big(\rho(\xb)\big) &= 
\int_{\Omega} c_i\big(\rho(\xb)\big)\, dV = 0, 
\quad i = 2,\dots,n_c, \label{eq cm1 c}
\\
R_{p}^{(i)}\big(\ub(\xb,\rho), \rho(\xb)\big) &= 
\int_{\Omega} r_p^{(i)}\big(\ub(\xb,\rho), \rho(\xb)\big)\, dV = 0, 
\quad i = 1,2, \label{eq cm1 d}
\\
R_{b}^{(i)}\big(\ub(\xb,\rho)\big) &= 
\int_{\partial\Omega_{\ub}} r_b^{(i)}\big(\ub(\xb,\rho)\big)\, dA = 0, 
\quad i = 1,\dots,n_{b}, \label{eq cm1 e}
\\
\rho(\xb) &\in [0,1], 
\quad \forall\, \xb \in \Omega, \label{eq cm1 f}
\end{align}
\end{subequations}
where $V$ is the domain volume and $\psi_f$ is the target volume fraction, the residuals $R_p^{(i)}\big(\ub(\xb,\rho), \rho(\xb)\big)$ and $R_b^{(i)}\big(\ub(\xb,\rho)\big)$ correspond to PDEs and BCs/ICs, respectively. \Cref{eq cm1} assumes that the objective function, constraints, and residuals can be represented as integrals of appropriately defined local functions. The use of the integral form relies on the assumption that as the residual integral approaches zero, the residual function at each collocation point also vanishes, provided that the local functions are sufficiently smooth and bounded. However, enforcing BCs or design constraints solely through residual minimization can be inaccurate particularly during the early stages of optimization, which may complicate the optimization process.

We construct our PIGP framework based on the DEM method where the potential energy is minimized instead of the PDE residuals. Accordingly, for CM of linear elastic solids, we replace \Cref{eq cm1 d} with the potential energy functional $L_P(\ub(\xb,\rho), \rho(\xb))$:

\begin{subequations}\label{eq cm2}
\begin{align}
\rho(\xb) &=
\argmin_{\rho} 
L_C\big(\ub(\xb,\rho), \rho(\xb)\big) = 
\argmin_{\rho}
\int_{\Omega} l_c\big(\ub(\xb,\rho), \rho(\xb)\big)\, dV, \label{eq cm2 a}
\intertext{subject to:}
\ub(\xb,\rho) &= \argmin_{\ub} L_P\big(\ub(\xb,\rho), \rho(\xb)\big) \notag\\
&= \argmin_{\ub}\left( \frac{1}{2}
\int_{\Omega} l_p\big(\ub(\xb,\rho), \rho(\xb)\big)\, dV - \int_{\partial \Omega_{F}} \fb(\xb)^{\text{T}} \ub(\xb,\rho) \, dA\right), \label{eq cm2 b}
\\
C_1\big(\rho(\xb)\big) &= \int_{\Omega} \rho(\xb)\, dV - \psi_f V = 0, \label{eq cm2 c} 
\\
C_i\big(\rho(\xb)\big) &= 
\int_{\Omega} c_i\big(\rho(\xb)\big)\, dV = 0, 
\quad i = 2,\dots,n_c, \label{eq cm2 d}
\\
R_{b}^{(i)}\big(\ub(\xb,\rho)\big) &= 
\int_{\partial\Omega_{\ub}} r_b^{(i)}\big(\ub(\xb,\rho)\big)\, dA = 0, 
\quad i = 1,\dots,n_{b}, \label{eq cm2 e}
\\
\rho(\xb) &\in [0,1], 
\quad \forall\, \xb \in \Omega, \label{eq cm2 f}
\end{align}
\end{subequations}
where $l_p(\ub,\rho)$ is the strain energy density function. $\rho(\xb)$ is the design variable converged at the optimum objective compliance, and the solution variable $\ub(\xb,\rho)$ in \Cref{eq cm2 b} satisfy the force equilibrium constraint.
To solve the constrained optimization problem in \Cref{eq cm2} with a simultaneous and mesh-free approach, we first parameterize all variables via the differentiable multi-output function $\zb(\xb;\zetab)$:
\begin{equation}\label{eq param}
\zb(\xb;\zetab) := [ u_1(\xb,\rho),u_2(\xb,\rho),\rho(\xb)]^{\text{T}},
\end{equation}
where $\zetab$ denotes the parameters. Then, we reformulate the constrained problem in \Cref{eq cm2} as the unconstrained optimization problem using the penalty method \citep{nocedal_numerical_2006}:
\begin{equation}\label{eq penal1}
\begin{aligned}
\zetab = \argmin_{\zetab} L\big(\zb(\xb;\zetab)\big) = \argmin_{\zetab} \bigg[
&L_C\big(\zb(\xb;\zetab)\big) + \alpha_{0} L_P\big(\zb(\xb;\zetab)\big) + \alpha_{1} C_1^2(z_{\rho}(\xb;\zetab))
+ 
\\ &\sum_{i=2}^{n_c} \alpha_i C_i^2\big(z_{\rho}(\xb;\zetab)\big) + \sum_{j=1}^{n_b} \alpha_j R_b^{(j)}\big(\zb_{u}(\xb;\zetab)\big)^2
\bigg],
\end{aligned}
\end{equation}
where $\alpha_i$ for $i = 0, 1, \cdots, n_b + n_c$ are scalar penalty factors, and for notational convenience the shorthands $z_{\rho}(\xb;\zetab) := \rho(\xb)$ and $\zb_u(\xb;\zetab) := [u_1(\xb,\rho),u_2(\xb,\rho)]^{\text{T}}$ are defined.
As detailed below, each term on the right-hand side of \Cref{eq penal1} can be evaluated by numerically approximating the integrals.

When parameterizing the solution space, we represent both $\ub(\xb,\rho)$ and $\rho(\xb)$ as outputs of a single function, i.e., $\zb(\xb;\zetab)$, to capture their natural correlations. However, this unified formulation must 
(1) be sufficiently expressive to capture diverse topologies and their associated state fields, 
(2) be able to capture sharp interfaces and stress localizations and 
(3) accommodate efficient gradient computation with respect to both $\zetab$ and $\xb$, particularly for estimating the gradient of the objective function in \Cref{eq penal1} with respect to $\rho(\xb)$ at each CP using the adjoint Lagrangian method. In the following subsections, we introduce a specific ML-based parametrization that not only has the above features, but also simplifies the optimization process.  

\subsection{Parameterization of State and Design Variables} \label{sec method param}
While a gradient-based optimizer can solve the minimization problem in \Cref{eq penal1}, care must be taken to ensure all the terms on the right-hand side are effectively minimized within a reasonable number of optimization iterations. This is especially important because the scale of the various terms in \Cref{eq penal1} can dramatically change during the optimization. Moreover, we have observed that if the BCs are not strictly enforced during the entire optimization process, the estimated solution fields (i.e., the displacements) and hence the designed topology substantially suffer. To address all of these issues, we introduce a particular form for $\zb(\xb;\zetab)$.

Specifically, we place GP priors on all state and design variables to leverage their reproducing property which, as shown below, allows the exclusion of all data-based constraints (e.g., BC residuals or specific design constraints) from \Cref{eq penal1}. These GP priors are equipped with distinct kernels but share a common multi-output mean function. We denote this mean function, parameterized by $\thetab$, as:
\begin{equation}\label{eq mean m}
\mb(\xb;\thetab) = [m_{u_1}(\xb;\thetab),m_{u_2}(\xb;\thetab),m_{\rho}(\xb;\thetab)]^{\text{T}},
\end{equation}
where $m_{u_1}(\xb;\thetab)$, $m_{u_2}(\xb;\thetab)$, and $m_{\rho}(\xb;\thetab)$ denote the mean functions for the two displacement components and the density, respectively. The parametric form of this shared mean function will be discussed in detail in \Cref{subsec shared mean}. To illustrate the reproducing property of GPs for the displacement variables, we express the posterior distribution of a GP prior conditioned on the prescribed displacements as:
\begin{subequations} \label{eq GP u}
    \begin{align}
    u_i(\xb^*; \thetab, \phib_{u_i}) &:= \mathbb{E}[u_i^* | \tilde{\ub}_i, \Xb_i] = m_{u_i}(\xb^*; \thetab) + \Vb_i^T \rb_i, \label{eq GP u 1} \\
    \Vb_i & = c_i^{-1}(\Xb_i, \Xb_i; \phib_{u_i}) \, c_i(\Xb_i, \xb^*; \phib_{u_i}), \label{eq GP u 2} \\
    \rb_i & = \tilde{\ub}_i - m_{u_i}(\Xb_i; \thetab), \label{eq GP u 3}
    \end{align}
\end{subequations}
where $i = 1,2$ corresponds to the two spatial dimensions, $\xb^* = [x^*, y^*]^{\text{T}}$ is an arbitrary query point in the design domain, and $c_i(\xb, \xb^{\prime}; \phib_{u_i})$ denotes the kernel with hyper-parameters $\phib_{u_i}$. The vector $\tilde{\ub}_i$ represents the prescribed displacements at points $\Xb_i$ and so the \textit{residual} vector $\rb_i$ quantifies the error of $m_{u_i}(\Xb; \thetab)$ in matching $\tilde{\ub}_i$. 
Using $\Xb_i$ in place of $\xb^*$ it can be easily verified that the left-hand side of \Cref{eq GP u 1} equals $\tilde{\ub}_i$ regardless of the choice of the mean function and the kernel. Hence, we can sample from the functions that specify the constraints on the state variables and then use those samples in our parameterization to automatically ensure that it always satisfies the BC constraints (as long as sufficient samples are taken from the constraint functions). 

Similar to \Cref{eq GP u} we parametrize the intermediate density field as:
\begin{subequations} \label{eq GP rho}
    \begin{align}
    \rho(\xb^*; \thetab, \phib_{\rho}) &:= \mathbb{E}[\rho^* | \tilde{\rhob}, \Xb_{\rho}] = m_{\rho}(\xb^*; \thetab) + \Vb_{\rho}^T \rb_{\rho}, \label{eq GP rho 1} \\
    \Vb_{\rho} & = c_{\rho}^{-1}(\Xb_{\rho}, \Xb_{\rho}; \phib_{\rho}) \, c_{\rho}(\Xb_{\rho}, \xb^*; \phib_{\rho}), \label{eq GP rho 2} \\
    \rb_{\rho} & = \tilde{\rhob} - m_{\rho}(\Xb_{\rho}; \thetab), \label{eq GP rho 3}
    \end{align}
\end{subequations}
where the vector $\tilde{\rhob}$ is the prescribed design variables at points $\Xb_{\rho}$. 
Now, to constrain $\rho(\xb; \thetab, \phib_{\rho})$ to the admissible range $[0,1]$ and promote binary separation for the final density function $\rho(\xb; \thetab, \phib_{\rho})$, the projection function $P(\cdot)$ \citep{wang_projection_2011} is applied to the parametrization in \Cref{eq GP rho} as:
\begin{equation}\label{eq proj rho}
    P\big(\rho(\xb; \thetab, \phib_{\rho})\big) =  \frac{\tanh(\beta \rho_{t}) + \tanh\big(\beta (\rho(\xb; \thetab, \phib_{\rho}) - \rho_{t})\big)}{\tanh(\beta \rho_{t}) + \tanh\big(\beta (1 - \rho_{t})\big)},
\end{equation}
where $\rho_{t} = 0.5$ is the threshold density and $\beta = 8$ controls the sharpness of the transition between the two phases. Using the parameterizations in \Cref{eq GP u} and \Cref{eq proj rho}, we redefine the parameterization in \Cref{eq param} as:
\begin{equation}\label{eq param2}
\zb^{\prime}(\xb;\thetab,\phib) := [ u_1(\xb;\thetab,\phib_{u_1}),u_2(\xb;\thetab,\phib_{u_2}),\rho(\xb;\thetab,\phib_{\rho})]^{\text{T}},
\end{equation}
where $\zb^{\prime}_u(\xb;\thetab,\phib_u) := \ub(\xb;\thetab,\phib_u) = [u_1(\xb;\thetab,\phib_{u_1}), u_2(\xb;\thetab,\phib_{u_2})]^{\text{T}}$, $z^{\prime}_{\rho}(\xb;\thetab,\phib_{\rho}) := \rho(\xb;\thetab,\phib_{\rho})$, $\phib = [\phib_{u_1}, \phib_{u_2}, \phib_{\rho}]$, and $\phib_u = [\phib_{u_1}, \phib_{u_2}]$. 

The performance of our parameterization depends on the kernels chosen for each variable. For simplicity and computational efficiency, we follow the recommendations of \citet{mora_gaussian_2024} and use the Gaussian kernel for all of our variables:
\begin{equation} \label{eq kernel}
    c(\xb, \xb^{\prime}; \phib) = s^2\exp \left\{ - (\xb - \xb')^{\text{T}} \, \mathrm{diag}(\phib) \, (\xb - \xb') \right\} + \mathbbm{1}\{\xb == \xb'\} \, \delta,
\end{equation}
where $\mathbbm{1}$ is a binary indicator function that returns 1 when $\xb == \xb'$ and 0 otherwise, $s^2$ denotes the process variance, and $\delta = 10^{-5}$ is the nugget (a small positive number) that ensures the invertibility of the covariance matrix $c(\Xb, \Xb; \phib)$. 
As recommended by \citet{mora_gaussian_2024}, we assign independent kernels with \textit{fixed} hyperparameters to each mean function in \Cref{eq mean m} to make sure the state and design variables strictly satisfy the applied constraints. The rationale for using separate kernels is as follows:  
(1) it facilitates the handling of heterogeneous data, particularly when BCs for state variables and prescribed design features are defined at different locations or along complex geometries;  
(2) it reduces computational and memory costs by allowing independent covariance matrices to be constructed for each output variable, rather than a single large joint matrix. Additionally, fixing $\phib$ allows us to pre-compute, invert, and cache all covariance matrices; dramatically decreasing the training costs   since repeated inversion of large matrices at each iteration would be computationally expensive.  It also improves training stability by keeping $\phib$ fixed. Compared with our previous work \citep{yousefpour_simultaneous_2025}, we heuristically select\footnote{Further details on the heuristic selection of $\phib$ are provided in \Cref{appendix phi}.} a relatively small length-scale parameter $\phib = 0.5$ for all kernels and keep it fixed throughout training. 

Using \Cref{eq param2} with fixed $\hat{\phib}$, we now rewrite \Cref{eq penal1} as:
\begin{align}\label{eq penal2}
\hat{\thetab} = \argmin_{\thetab} L\big(\zb^{\prime}(\xb;\thetab,\hat{\phib})\big) 
&= \argmin_{\thetab} \bigg[
L_C\big(\zb^{\prime}(\xb;\thetab,\hat{\phib})\big) 
+ \alpha_{0} L_P\big(\zb^{\prime}(\xb;\thetab,\hat{\phib})\big) \notag\\
&\quad + \alpha_{1} C_1^2\big(z^{\prime}_{\rho}(\xb;\thetab,\hat{\phib}_{\rho})\big)
\bigg]
\end{align}
where all data-based design constraints (except the one on volume fraction) are dropped due to the reproducing property of our parametrization with GPs. 

It is important to highlight that \Cref{eq param2} is fundamentally different from \Cref{eq param} since our parameterizations in \Cref{eq GP u,eq GP rho} do \textit{not} explicitly capture the underlying dependency of displacements on the density. As detailed in \Cref{subsec method control}, we implicitly build this dependency into our framework by reformulating the gradients that are used by the Adam optimizer. 

\subsubsection{Shared Mean Function} \label{subsec shared mean}
As motivated earlier, we use a specific NN architecture as the shared mean function of all the three GPs in \Cref{eq GP u,eq GP rho}. Specifically, we use PGCAN which we have recently developed \citep{shishehbor_parametric_2024} to address the spectral bias of MLPs: The idea behind PGCAN is to represent the spatiotemporal domain with a grid-based parametric encoder that has learnable features on its vertices. The encoder is followed by a decoder (an MLP with attention layers) to convert the features to the outputs (displacements and density in our case). For any query point in the domain, its feature vector is obtained by interpolating the feature vectors of the vertices of its enclosing cell\footnote{Note that the encoder partitions the input space into cells but this does \textit{not} translate into any discretization due to the fact that we can obtain features for any point in the domain and then pass these features into the decoder to obtain the responses.}. That is, the model's predictions at $\xb$ do \textit{not} depend on all the features of the encoder; rather they only depend on the features of the cell that encloses $\xb$. This property enables PGCAN to mitigate spectral bias and more accurately estimate high-gradient solutions. 

\begin{figure*}[!h]
    \centering
    \includegraphics[width = 0.9\textwidth]{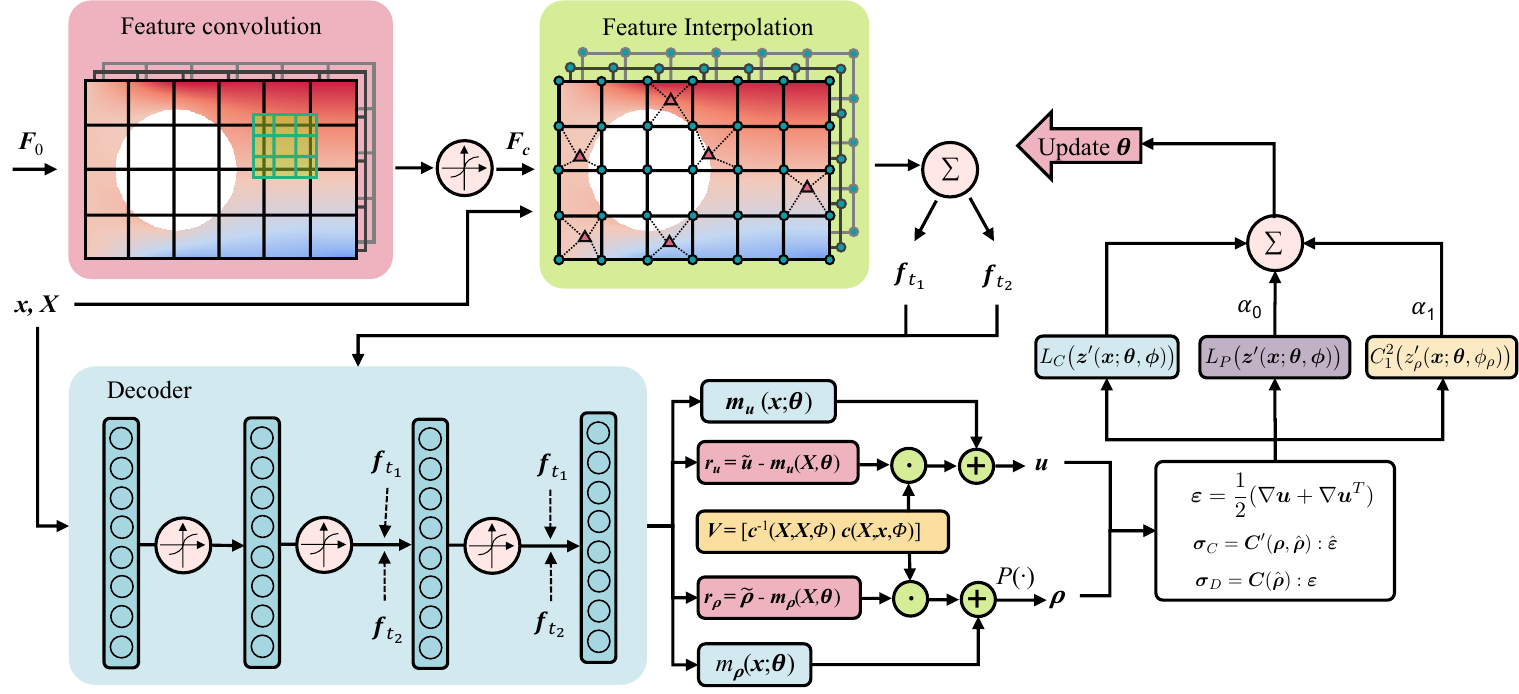}
\caption{\textbf{PIGP framework for CM:} the mean functions are parameterized with the multi-outputs from the PGCAN having three modules: convolutional neural network-based feature encoding, feature interpolation, and decoding via a shallow MLP. GPs are employed on those mean functions to impose displacement BCs and density constraints.}
    \label{fig pigp}
\end{figure*}

\cite{shishehbor_parametric_2024} demonstrate that a grid-based encoder is susceptible to overfitting and to mitigate this issue, we (1) convolve the features with a small kernel, and (2) have multiple grids which are diagonally shifted with respect to the original one so that a query point's coordinates are slightly perturbed. So, as schematically illustrated in \Cref{fig pigp} for a 2D spatial domain, our PGCAN parametrizes the domain via the trainable feature tensor $\Fb_0 \in \mathbb{R}^{N_{rep} \times N_f \times N_x^e \times N_y^e}$ where $N_{rep}$ denotes the number of grid repetitions, $N_f = 128$ represents the number of features per grid vertex, and $N_x^e$ and $N_y^e$ specify the number of vertices along the $x$ and $y$ directions, respectively. 
To promote information flow across neighboring cells, we apply a $3 \times 3$ convolution with $N_f$ input and output channels to $\Fb_0$ followed by a $\tanh$ activation function. The resulting feature map $\Fb_c \in \mathbb{R}^{N_{rep} \times N_f \times N_x^e \times N_y^e}$ is placed over the spatial domain with $N_{rep}$ small diagonal offsets ($N_{rep} = 3$ in \Cref{fig pigp}). 

To compute the feature vector at the arbitrary query point $\xb$, we first obtain the local (i.e., within cell) coordinates $\bar{\xb} \in [0,1]$ and then apply cosine transformation to them to enable differentiability with respect to the inputs:
\begin{equation}\label{eq cosine}
    \xb^* = \frac{1}{2}\big(1 - \cos(\pi\bar{ \xb})\big).
\end{equation}
Then, the feature vector at $\xb^* = [x^*,y^*]^{\text{T}}$ corresponding to each grid $m \in \{1, \cdots,N_{rep}\}$ with perturbation is obtained by bilinearly interpolating the features at the vertices of the cell that encloses $\xb^*$:
\begin{equation}\label{eq interpolation}
\fb_{\tb}^m(\xb^*) = (1 - x^*)(1 - y^*) \fb^m_{\tb(0,0)} + (1 - x^*) y^* \fb^m_{\tb(0,1)} + x^*(1 - y^*) \fb^m_{\tb(1,0)} + x^* y^* \fb^m_{\tb(1,1)},
\end{equation}
\noindent where $\fb^m_{\tb(i,j)} \in \mathbb{R}^{N_f}$ with $i,j \in \{0,1\}$ are the extracted features from $\Fb_c$. The final feature vector at the query point is obtained by summing the feature vectors across the grid repetitions, i.e., $\fb_{\tb}(\xb) = \sum_{m}\fb_{\tb}^m(\xb^*)$. 

Once $\fb_{\tb}(\xb)$ is obtained, it is split into two vectors of equal length $\fb_{\tb_1}(\xb)$ and $\fb_{\tb_2}(\xb)$ and passed to a shallow decoder network with 3 hidden layers of 64 neurons. As described in \cite{shishehbor_parametric_2024}, these features are used to sequentially modulate the hidden states of the network, thereby enhancing gradient propagation during training.

\subsection{Physics-based Gradients} \label{subsec method control}
While $\partial L(\cdot) / \partial \thetab$ can be readily obtained based on \Cref{eq penal2} using AD, it cannot be directly used to optimize $\thetab$ because of the following reasons. First, the penalty term corresponding to the total potential energy can become negative during the optimization. So, we must augment $L_P(\cdot)$ with a term that ensures $L_P(\cdot)$ remains positive during the optimization (otherwise it cannot serve as a proper regularizer). 
Second, our parameterization does not explicitly make $\ub(\cdot)$ dependent on $\rho(\cdot)$ which is a physical requirement. Introducing this dependence relies on deriving the gradients of $L_C(\cdot)$ and $L_P(\cdot)$ based on the physics of the problem such that they can be added and then used for updating $\thetab$ at each optimization iteration. 

Referring to \Cref{sec dem} we observe that the displacement field is obtained by minimizing the total potential of a fixed system, i.e., the minimization must be done for a fixed $\rho(\cdot)$. So, there must be no contributions from the density to the gradients, i.e., $\partial L_P(\cdot)/\partial \thetab = \frac{\partial L_P}{\partial \ub } \frac{\partial \ub}{\partial \thetab} + 0$ where $\frac{\partial L_P}{\partial \rhob} \frac{\partial \rhob}{\partial \thetab}$ is replaced by $0$. 
Conversely, for the compliance, the gradients are taken with respect to the density field while the displacement (or strain\footnote{While fixing the strains or displacements provide the same results, we choose the former due to its computational efficiency.}) field is fixed \textit{and} assumed to satisfy the force equilibrium constraints. As detailed below, we satisfy these conditions by obtaining $\partial L_C(\cdot) / \partial \rho(\cdot)$ based on the adjoint method. 

To accommodate the above requirements we remove or detach the gradients with respect to certain variables in our computational graph. To facilitate the descriptions, we denote \textit{detached} variables with a hat symbol such as $\hat{\epsilonb}(\xb;\thetab,\phib_u)$ and $\hat{\rho}(\xb;\thetab,\phib)$ which do not have gradients and share the same numerical values as their with-gradient counterparts. 
Using these detached variables, the individual terms in \Cref{eq penal2} can be approximated as:
\begin{subequations}\label{eq loss}
    \begin{align}
    L_C\big(\zb^{\prime}(\xb;\thetab,\phib)\big) &= \frac{V}{n_{cp}} \sum_{i=1}^{n_{cp}}\sigmab_C(\xb_i;\thetab,\phib) : \hat{\epsilonb}(\xb_i;\thetab,\phib_u), \label{eq loss 1} \\
    L_P\big(\zb^{\prime}(\xb;\thetab,\phib)\big) &= \frac{V}{2 n_{cp}} \sum_{i=1}^{n_{cp}}\sigmab_P(\xb_i;\thetab,\phib) : \epsilonb(\xb_i;\thetab,\phib_u) \notag \\
    &\quad - \sum_{i=1}^{n_{f}} \fb(\xb_i)^{\text{T}} \ub(\xb_i;\thetab,\phib_u)s_i + \tau\big(\zb^{\prime}(\xb;\thetab,\phib)\big), \label{eq loss 2} \\
    C_1^2\big(z^{\prime}_{\rho}(\xb;\thetab,\phib_{\rho})\big) &= \left( \frac{1}{n_{cp}} \sum_{i=1}^{n_{cp}} \rho(\xb_i;\thetab,\phib_{\rho}) / \psi_f - 1 \right)^2,\label{eq loss 3}
    \end{align}
\end{subequations}
where $n_{cp}$ and $n_{f}$ denote the numbers of CPs in the entire domain and along the traction BCs, respectively. The offset term $\tau(\cdot)$ as well as the two stress tensors $\sigmab_C(\cdot)$ and $\sigmab_P(\cdot)$ are calculated as detailed below. 

At force equilibrium, $L_P(\cdot)$ is negative since the strain energy equals one half of the external work. Even before equilibrium during training, a specific set of parameters might cause $L_P(\cdot)$ to become negative. 
Since a negative $L_P(\cdot)$ cannot be a regularizer, we add $\tau(\cdot)$ to it:
\begin{equation}\label{eq tau}
    \tau\big(\zb^{\prime}(\xb;\thetab,\phib)\big) := \frac{\Delta}{2}\sum_{i=1}^{n_{f}} \fb(\xb_i)^{\text{T}} \hat{\ub}(\xb_i;\thetab,\phib_u)s_i,
\end{equation}
where $\Delta = 1.01$ is a constant that is slightly greater than 1 to ensure that $\tau(\cdot)$ is always greater than one half of the external work (which has a negative sign in $L_P(\cdot)$, see \Cref{eq pi}). Note that $\tau(\cdot)$ is based on $\hat{\ub}(\cdot)$ so it does not have any gradients with respect to $\ub(\cdot)$.

The two stress tensors in \Cref{eq loss} are calculated via the constitutive law of a linear elastic material:
\begin{subequations}\label{eq sigma}
    \begin{align}
        \sigmab_C(\xb;\thetab,\phib) &:= \Cb_C(\xb;\thetab,\phib_{\rho}) : \hat{\epsilonb}(\xb;\thetab,\phib_u), \label{eq sigma 1} \\
        \sigmab_P(\xb;\thetab,\phib) &:= \Cb_P(\xb;\thetab,\phib_{\rho}) : \epsilonb(\xb;\thetab,\phib_u), \label{eq sigma 2}
    \end{align}
\end{subequations}
where both $\Cb_C(\cdot)$ and $\Cb_P(\cdot)$ are stiffness tensors that are numerically the same but have different gradients. We achieve this feature by appropriately parameterizing the Young's modulus which, in turn, affects the gradients of stiffness tensor with respect to $\rho(\cdot)$. Following the SIMP methodology, we build $\Cb_P(\cdot)$ based on the following Young's modulus:
\begin{equation}\label{eq E hat}
    E_P\big(\xb;\thetab,\phib_{\rho}\big) = E_{min} + \hat{\rho}^p(\xb;\thetab,\phib_{\rho})(E_{max} - E_{min}),
\end{equation}
where $E_{\max} = 1$ denotes the Young’s modulus of the solid phase, $E_{\min} = 10^{-3}$ is a small positive constant introduced to further ensure numerical stability, and $p = 3$ is a penalty factor that promotes binary designs. The Poisson’s ratio $\nu$ is fixed at 0.3 to construct stiffness tensors.

In \Cref{subsec method sensitivity} we use the adjoint variable method to demonstrate that $\Cb_C(\cdot)$ can be constructed via the following Young's modulus:
\begin{equation}\label{eq E prime}
    E_C(\xb;\thetab,\phib_{\rho}\big) = 
    E_{\min} + \frac{\hat{\rho}^{2p}(\xb;\thetab,\phib_{\rho})}{\rho^p(\xb;\thetab,\phib_{\rho}\big) + \epsilon} \bigg(E_{\max} - E_{\min}\bigg),
\end{equation}
which has the same numerical value as in \Cref{eq E hat} but has different gradients, and $\epsilon = 10^{-5}$ is a small constant introduced to ensure numerical stability. It is straightforward to verify that, using \Cref{eq E prime}, the gradient of $L_C\big(\zb^{\prime}(\xb;\thetab,\phib)\big)$ with respect to $\rho(\xb;\thetab,\phib_{\rho})$ reproduces the same gradient as given in \Cref{eq sens dL3} from the adjoint variable method.

\subsubsection{Sensitivity Analysis by Adjoint Variable Method} \label{subsec method sensitivity}  
We use the adjoint method to obtain the gradient of $L_C(\cdot)$ with respect to $\rho(\cdot)$ while (1) detaching the gradients with respect to $\ub(\cdot)$ as motivated earlier, and (2) consider the dependence of $\ub(\cdot)$ on $\rho(\cdot)$ based on the physics of the problem.
We start by writing the Lagrangian:
\begin{align}\label{eq sens L1}
    L_C\big(\ub(\xb,\rho),\rho(\xb), \lambdab(\xb)\big) 
    &= \int_{\Omega} l_c\big(\ub(\xb,\rho),\rho(\xb)\big)\,dV + \int_{\Omega} \lambdab^T(\xb) \big(\nabla \cdot \sigmab(\xb)\big) \, dV\notag \\
    &= \int_{\Omega} \epsilonb(\xb) : \Cb(\xb) : \epsilonb(\xb) \, dV 
    + \int_{\Omega} \lambdab^T(\xb) \big(\nabla \cdot \sigmab(\xb)\big) \, dV,
\end{align}
where $\lambdab(\xb) = [\lambda_1(\xb), \lambda_2(\xb)]^{\text{T}}$ are the Lagrange multipliers. 
Using the divergence theorem, the last term on the right-hand side of \Cref{eq sens L1} can be expanded as:
\begin{equation}\label{eq sens divergence}
    \int_{\Omega} \lambdab^T(\xb) \big(\nabla \cdot \sigmab(\xb)\big) \, dV = - \int_{\Omega} \nabla \lambdab(\xb) : \sigmab(\xb) \, dV + \int_{\partial\Omega} \lambdab^{\text{T}}(\xb) \big(\sigmab(\xb) \nb(\xb) \big) \, dA,
\end{equation}
where $\nb(\xb)$ denotes the outward unit normal vector on the domain boundary. Additionally, $\sigmab(\xb) \nb(\xb) = \fb(\xb)$ and since the Cauchy stress tensor $\sigmab(\xb)$ is symmetric, we have $\nabla \lambdab(\xb) : \sigmab(\xb) = \text{sym}\big(\nabla \lambdab(\xb)\big) : \sigmab(\xb) = \epsilonb_{\lambdab}(\xb) : \sigmab(\xb)$ where $\epsilonb_{\lambdab}(\xb)$ denotes the strain tensor associated with the augmented field $\lambdab(\xb)$. So, we can rewrite \Cref{eq sens L1} as:
\begin{align}\label{eq sens L2}
    L_C(\cdot) 
    &= \int_{\Omega} \epsilonb(\xb) : \Cb(\xb) : \epsilonb(\xb) \, dV - \int_{\Omega} \epsilonb_{\lambdab}(\xb) : \Cb(\xb) : \epsilonb(\xb) \, dV 
    + \int_{\partial\Omega} \lambdab(\xb)^{\text{T}} \fb(\xb) \, dA.
\end{align}
Assuming that the external force $\fb(\xb)$ is independent of $\rho(\xb)$, the total derivative of $L_C(\cdot)$ with respect to $ \rho(\cdot)$ is given as:
\begin{align}\label{eq sens dL1}
    \frac{dL_C(\cdot)}{d\rho} &= 2 \int_{\Omega}\bigg( \frac{\partial \epsilonb(\xb)}{\partial \ub} : \frac{\partial \ub}{\partial \rho}\bigg) : \Cb(\xb) : \epsilonb(\xb) \, dV
    + \int_{\Omega} \epsilonb(\xb) : \frac{\partial \Cb(\xb)}{\partial \rho} : \epsilonb(\xb) \, dV \notag \\
    &\quad - \int_{\Omega} \epsilonb_{\lambdab}(\xb) : \frac{\partial \Cb(\xb)}{\partial \rho} : \epsilonb(\xb) \, dV
    - \int_{\Omega} \epsilonb_{\lambdab}(\xb) : \Cb(\xb) : \bigg(\frac{\partial \epsilonb(\xb)}{\partial \ub} : \frac{\partial \ub}{\partial \rho}\bigg) \, dV,
\end{align}
where we have used the fact that the stiffness tensor $ \Cb(\xb) $ is symmetric, that is:
\begin{equation}\label{eq sens sym}
    \int_{\Omega} \frac{\partial \epsilonb(\xb)}{\partial \ub} : \frac{\partial \ub}{\partial \rho} : \Cb(\xb) : \epsilonb_{\lambdab}(\xb) \, dV 
    = \int_{\Omega} \epsilonb_{\lambdab}(\xb) : \Cb(\xb) : \frac{\partial \epsilonb(\xb)}{\partial \ub} : \frac{\partial \ub}{\partial \rho} \, dV.
\end{equation}

Introducing $ \Ab(\cdot) = \frac{\partial \epsilonb(\xb)}{\partial \ub} : \frac{\partial \ub}{\partial \rho} $, we can write \Cref{eq sens dL1} as:
\begin{align}\label{eq sens dL2}
    \frac{dL_C(\cdot)}{d\rho} &= \int_{\Omega} \Ab(\xb) : \Cb(\xb) : \big(2 \epsilonb(\xb) - \epsilonb_{\lambdab}(\xb)\big) \, dV 
    + \int_{\Omega} \big(\epsilonb(\xb) - \epsilonb_{\lambdab}(\xb)\big): \frac{\partial \Cb(\xb)}{\partial \rho} : \epsilonb(\xb) \, dV.
\end{align}
Since the multiplier $\lambdab(\xb)$ is arbitrary, we set it such that $\epsilonb_{\lambdab}(\xb) =2\epsilonb(\xb)$ which simplifies \Cref{eq sens dL2} to:
\begin{equation}\label{eq sens dL3}
    \frac{dL_C(\cdot)}{d\rho} = -\int_{\Omega} \epsilonb(\xb) : \frac{\partial \Cb(\xb)}{\partial \rho} : \epsilonb(\xb) \, dV,
\end{equation}
where a negative sign appears in the gradient. So, in our approach, we define the Young's modulus per \Cref{eq E prime} so that when AD is used to obtain the gradients, a negative sign appears. 

\subsection{Computational Accelerations and Stability} \label{subsec comp accele}

As mentioned earlier, in our implementation we assign a GP prior with fixed hyper-parameters to each variable. This choice dramatically accelerates the optimization process as covariance matrices and their inverses (see \Cref{eq GP u,eq GP rho}) can be pre-computed and stored for repeated uses while optimizing $\thetab$ per \Cref{eq penal2}. 
To further increase the computational efficiency and robustness of our approach, we develop the following two strategies. 

\subsubsection{Numerical Approximations}\label{subsubsec adaptive}
We can improve computational performance by (1) numerically approximating some of the partial derivatives in \Cref{eq loss}, and (2) arranging the CPs in particular configurations that accelerate convergence rate while reducing overfitting issues and memory requirements. 

To simultaneously address these issues, we follow \cite{yousefpour_simultaneous_2025} who approximate the spatial derivatives such as $\partial \ub(\cdot)/\partial \xb$ via finite difference (FD). To this end, we place the CPs on a regular grid and pad the domain via two layers of points in each direction to increase gradient approximation accuracy close to the boundaries. This arrangement of CPs is in sharp contrast to most works where PDEs are solved via PIML because in these works the CPs are distributed in the domain either randomly or via a space-filling algorithm such as Sobol sequence. 
Details on our FD approximation and implementation are provided in \Cref{appendix ND}.

In addition to accelerating the estimation of spatial derivatives, arranging the CPs on a grid that does not change during optimization allows us to pre-compute and store the covariance matrices (and their inverses) that appear in \Cref{eq GP u,eq GP rho}. This approach dramatically decreases the computational costs while increasing the robustness as repeated matrix inversion are eliminated. 

The disadvantage of placing the CPs on a fixed regular grid is that, when combined with PGCAN, it can result into overfitting which is not observed in \cite{yousefpour_simultaneous_2025} as their work is based on MLPs. To address this issue, we define a series of grids with various resolutions and randomly select one for each optimization iteration. 
More specifically, we first define a fine and a coarse grid where the former is twice denser than the latter. Then, by linearly interpolating between these two grids, we generate a set of grids denoted by $\boldsymbol{\Sigma} = [\Xb_1, \dots, \Xb_{n_g}]$ where $\Xb_1$ and $\Xb_{n_g}$ represent the coarse and fine grids, respectively. For each grid in $\boldsymbol{\Sigma}$, we pre-compute and cache relevant quantities such as CP coordinates and the corresponding external load vectors. During training, one grid is randomly sampled from $\boldsymbol{\Sigma}$ at each step and used to compute spatial gradients via FD and evaluate the loss function. Further details on the overfitting issue associated with a fixed grid and the impact of the proposed adaptive grid scheme are provided in \Cref{appendix overfitting}.

\subsubsection{Curriculum Training} \label{sec method training}

Our framework for CM aims to simultaneously minimize three loss terms. In our studies we have observed that when the target volume fraction is very low the training process can become unstable. This issue is manifested by the large oscillations in the loss histories and high roughness in the designed topologies.

Motivated by curriculum training, in which a model is progressively trained by adapting its learning objectives \citep{soviany_curriculum_2022,Yousefpour2025LocalizedPG}, we address this issue by introducing a volume fraction scheduling strategy. In this strategy, we first initialize the target volume fraction $\psi_f$ with a value of $\psi_0$ that is close to the average initial density predicted by the mean function. As training progresses, we linearly reduce this target volume fraction to the desired value $\psi_{l}$ over a specified portion $\gamma$ (e.g., $30\%$ or $50\%$) of the total training epochs. This strategy can be formulated as:
\begin{equation}\label{eq vf schedule}
\psi_f =
\begin{cases}
\displaystyle \frac{\psi_l - \psi_0}{\gamma n_{tol}}\, n + \psi_0, & n \leq \gamma n_{tol}, \\
\psi_l, &  n > \gamma n_{tol},
\end{cases}
\end{equation}
where $n_{tol}$ denotes the total number of epochs and $\gamma$ controls the scheduling. The gradual reduction of volume fraction imposes a smooth topological evolution into the optimization; increasing the stability of the training.
In all of our experiments we use $\psi_0 = 0.6$ and $\gamma = 50\%$.

\section{Results and Discussion} \label{sec results}
\begin{figure*}[!t]
    \centering
    \begin{subfigure}[t]{0.195\textwidth}
        \captionsetup{justification=raggedright, singlelinecheck=false, skip=0pt, position=top}
        \caption{\textbf{MBB beam}}
        \includegraphics[width=\linewidth]{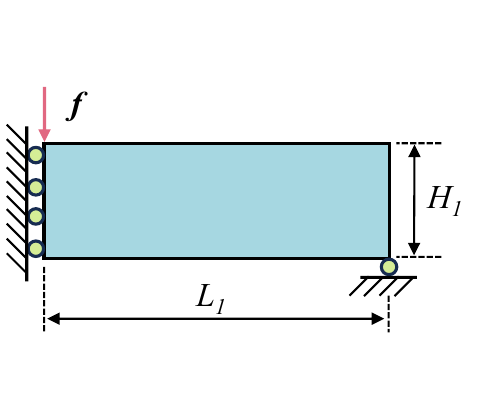}
        \label{fig example 1}
    \end{subfigure}
    \begin{subfigure}[t]{0.195\textwidth}
        \captionsetup{justification=raggedright, singlelinecheck=false, skip=0pt, position=top}
        \caption{\textbf{Cantilever beam}}
        \includegraphics[width=\linewidth]{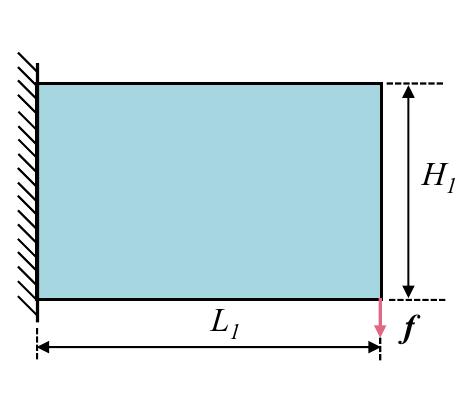}
        \label{fig example 2}
    \end{subfigure}
    \begin{subfigure}[t]{0.195\textwidth}
        \captionsetup{justification=raggedright, singlelinecheck=false, skip=0pt, position=top}
        \caption{\textbf{Uniformly loaded beam}}
        \includegraphics[width=\linewidth]{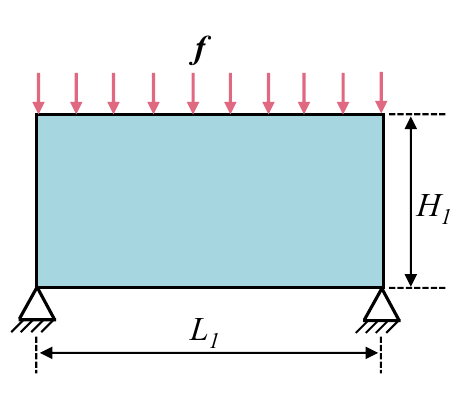}
        \label{fig example 3}
    \end{subfigure}
    \begin{subfigure}[t]{0.195\textwidth}
        \captionsetup{justification=raggedright, singlelinecheck=false, skip=0pt, position=top}
        \caption{\textbf{L-shape beam}}
        \includegraphics[width=\linewidth]{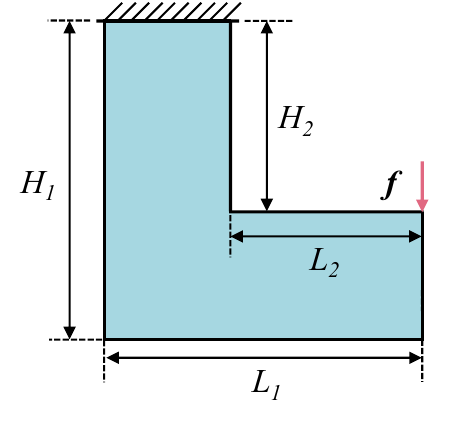}
        \label{fig example 4}
    \end{subfigure}
    \begin{subfigure}[t]{0.195\textwidth}
        \captionsetup{justification=raggedright, singlelinecheck=false, skip=0pt, position=top}
        \caption{\textbf{Hollow beam}}
        \includegraphics[width=\linewidth]{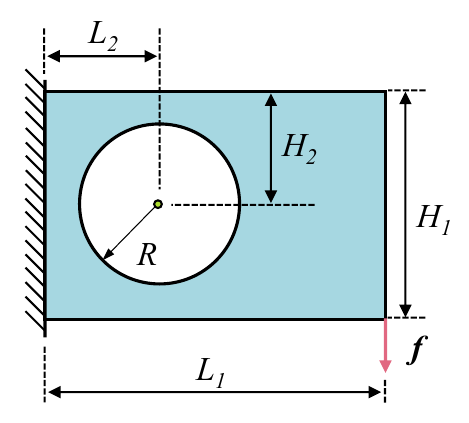}
        \label{fig example 5}
    \end{subfigure}
    \vspace{-2em}
    \caption{\textbf{Benchmark examples:} We consider five cases for compliance minimization including (a) Messerschmitt-Bölkow-Blohm (MBB) beam, (b) Cantilever beam, (c) Uniformly loaded beam, (d) L-shape beam, and (e) Hollow beam with a circular hole.}
    \label{fig example}
\end{figure*}

We compare our approach against SIMP and other ML-based methods on a set of benchmarks. Specifically, we consider the five canonical examples illustrated in \Cref{fig example} with the corresponding dimensions, desired volume fractions, and applied external forces detailed in \Cref{tab dimension}.
The size of the design domains and magnitude of the loads are selected to satisfy the small deformation assumption of linear elasticity. 

\begin{table*}[!b]
    \centering
    \renewcommand{\arraystretch}{1.5}
    \small
    \setlength\tabcolsep{8pt}
    \resizebox{\textwidth}{!}{%
    \begin{tabular}{l|c|c|c|c|c|c|c} 
    \hline
    \textbf{Example} & $H_1$ ($\mathrm{mm}$) & $L_1$ ($\mathrm{mm}$) & $H_2$ ($\mathrm{mm}$) & $L_2$ ($\mathrm{mm}$) & $R$ ($\mathrm{mm}$) & $f$ ($\mathrm{N}$ or $\mathrm{N/mm}$) & \textbf{$\psi_f$} \\ 
    \hline
    MBB beam           & 50   & 150 & --   & --   & --    & $0.1$       & 0.5 \\
    Cantilever beam    & 100  & 160 & --   & --   & --    & $0.1$       & 0.3 \\
    Uniformly Loaded beam& 100  & 200 & --   & --   & --    & $10^{-3}$   & 0.3 \\
    L-shape beam       & 100   & 100  & 60  & 60  & --    & $0.1$       & 0.5 \\
    Hollow beam        & 100   & 150  & 50  & 50  & 33.33 & $0.1$       & 0.5 \\
    \hline
    \end{tabular}
    }
    \caption{\textbf{Parameter values:} Geometric parameters, external force, and target volume fractions for the five examples are listed. All dimensions are in $\mathrm{mm}$ and $f$ denotes the external force magnitude for either a point ($\mathrm{N}$) or distributed ($\mathrm{N/mm}$) load.}
    \label{tab dimension}
\end{table*}

We implement SIMP in MATLAB   from \citet{andreassen_efficient_2011} as our baseline method and execute it on a Windows desktop with an Intel\textsuperscript{\textregistered} Core\textsuperscript{TM} i7-11700K CPU. The Hevaviside projection function in \Cref{eq proj rho} is added to the original SIMP method in order to promote binary design. More details are provided in \Cref{appendix proj}. Our PIGP framework is implemented in Python and executed on an NVIDIA A100 GPU. Due to the non-convex nature of TO, we run both approaches $10$ times with different random initializations to assess variability. Three evaluation metrics are used for comparing the optimization effectiveness across the two approaches: final compliance, gray area fraction, and interface fraction.  For consistency, the density at the center of each FE element is predicted via our PIGP to match SIMP (which directly optimizes the density at the element centers) and the final compliance is then evaluated with FEM on both optimized topologies prior to any post-processing. It is noted that our PIGP is trained on nodal (not element center) data but we do not use nodal values in the comparisons since interpolating SIMP results in excessive gray regions (this comparison procedure favors SIMP slightly). Also, the gray area fraction is defined as the fraction of element centers with densities between 0.05 and 0.95 \citep{chandrasekhar_tounn_2021}. The interface fraction refers to the proportion of  element centers located on solid-void interfaces. More details about interface fraction are provided in \Cref{appendix interface}.

As detailed in \Cref{subsubsec adaptive}, we use the set of regular grids $\boldsymbol{\Sigma} = [\Xb_1, \dots, \Xb_{n_g}]$ for estimating the loss components in \Cref{eq penal2}. Since the domain sizes in \Cref{fig example} are different, we ensure consistency by (1) selecting the coarse grid $\Xb_1$ such that the spacing between adjacent grid vertices is 1 $\mathrm{mm}$ along both $H_1$ and $L_1$, and (2) density of grid cells in the fine grid $\Xb_{n_g}$ is twice that of $\Xb_1$. The CP resolutions of the coarse and fine grids, along with the  number of pre-defined grids $n_g$ are summarized in \Cref{tab grid} for each example. Our experiments indicate that the final topologies produced by our framework are insensitive to $n_g$, as long as it is set to a reasonably large constant (i.e., $100$). For the baseline SIMP method, we also report the numbers of finite element nodes along the two edges of the design domain in \Cref{tab grid}.

\begin{table*}[!t]
    \centering
    \renewcommand{\arraystretch}{1.5}
    \small
    \setlength\tabcolsep{6pt}
    \caption{\textbf{Summary of finite element nodes and the grid densities:} The finite element node resolutions $(N_x, N_y)$ along $L_1$ and $H_1$ are reported for the SIMP baseline method. For the PIGP framework, the numbers of CPs are listed for the coarse $(N_x^{(1)}, N_y^{(1)})$ and fine $(N_x^{(n_g)}, N_y^{(n_g)})$ grids in each example, together with the total number of grids $n_g$. The intermediate grids are constructed as described in \Cref{subsubsec adaptive}.}
    \begin{tabular}{l|c|ccc}
        \hline
        \multirow{2}{*}{\textbf{Example}}
        & \multicolumn{1}{c|}{ \textbf{SIMP}} 
        & \multicolumn{3}{c}{ \textbf{PIGP}} \\
        \cline{2-5}
        & \textbf{ Node Resolution $(N_x, N_y)$} 
        & \textbf{Coarse Grid $(N_x^{(1)}, N_y^{(1)})$} 
        & \textbf{Fine Grid $(N_x^{(n_g)}, N_y^{(n_g)})$} 
        & $n_g$ \\
        \hline
        MBB               &  (151, 51)   & (151,  51)  & (301, 101) & 150 \\
        Cantilever        &  (161, 101)  & (161, 101)  & (321, 201) & 160 \\
        Uniformly loaded  &  (201, 101)   & (201, 101)  & (401, 201) & 200 \\
        L-shape           &  (101, 101)   & (101, 101)  & (201, 201) & 100 \\
        Hollow            &  (151, 101)  & (151, 101)  & (301, 201) & 150 \\
        \hline
    \end{tabular}
    \label{tab grid}
\end{table*}

We use PGCAN as the backbone network architecture across all benchmarks and include an MLP model for comparison. Both architectures use the $\swish$ activation function within the hidden layers. In PGCAN, we define $N_x^e$ and $N_y^e$ as the number of grid vertices along the $x$ and $y$ directions for the feature tensor $\Fb_0 \in \mathbb{R}^{N_{\mathrm{rep}} \times N_f \times N_x^e \times N_y^e}$. $N_x^e$ and $N_y^e$ are determined using the resolution parameter $\Res$, which specifies the number of grid vertices per $100\mathrm{mm}$ edge length. Specifically,
\begin{subequations}\label{eq res}
\begin{align}
N_x^e &= \Res \cdot \frac{L_1}{100}, \label{eq:res 1}\\
N_y^e &= \Res \cdot \frac{H_1}{100}. \label{eq:res 2}
\end{align}
\end{subequations}
In MLP, we use 6 fully-connected hidden layers with 64 neurons for each layer. We employ the Adam optimizer to train all models up to $20,000$ epochs, with an initial learning rate of $10^{-3}$ that decays by a factor of $0.75$ at four evenly spaced intervals. The loss weight factors  in \Cref{eq penal2} are fixed as $\alpha_0 = 10^2$ and $\alpha_1 = 10^3$ across all examples.

\subsection{Comparison Studies and Topology Evolution} \label{sec results comp}

\begin{figure*}[!t]
    \centering
    \includegraphics[width = 1.0\textwidth]{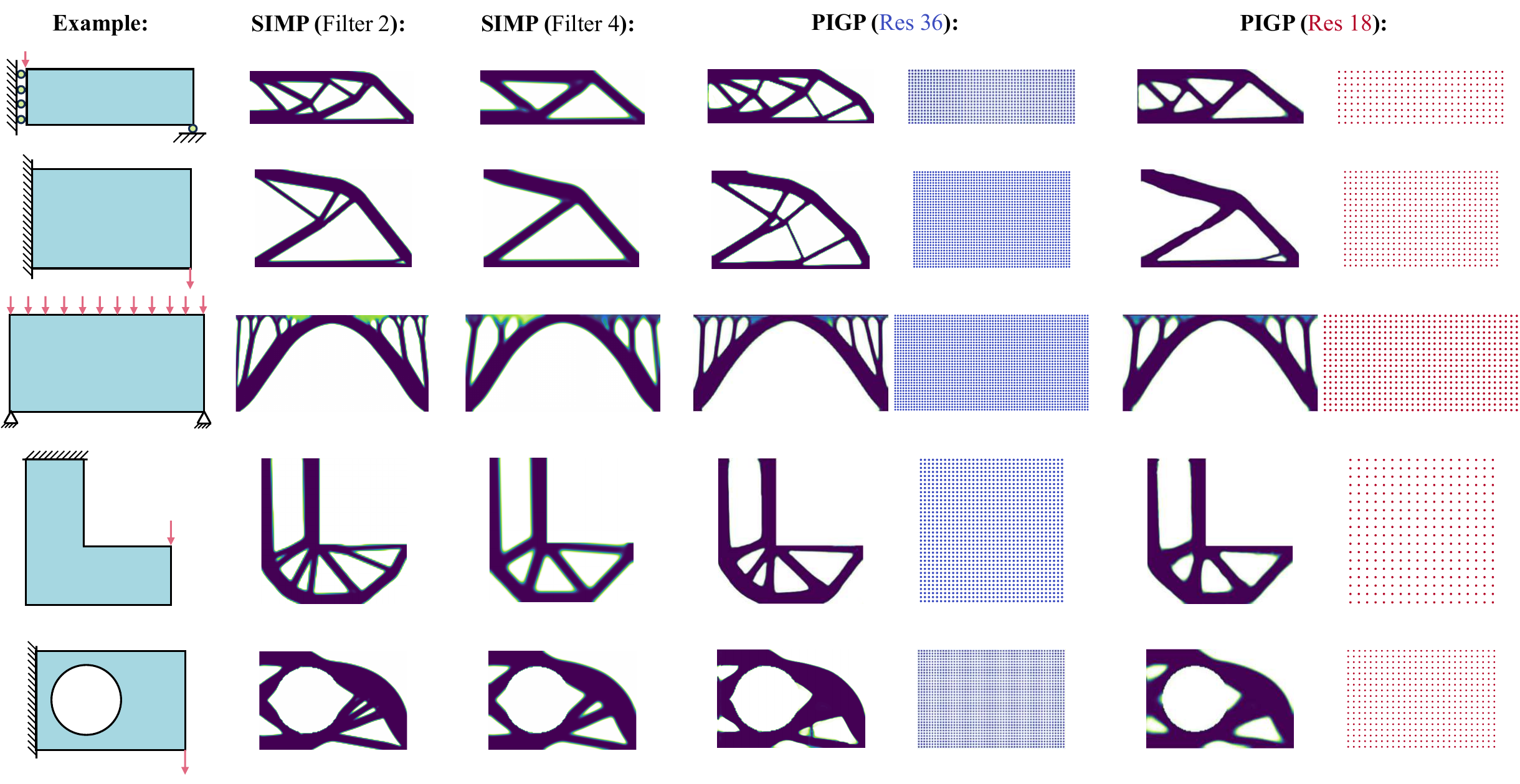} \caption{\textbf{Comparison of final designs:} For both approaches we visualize the topologies corresponding to the median compliance in each example. The cell vertices of PGCAN are also shown to demonstrate the effect of $Res$ on partitioning the design domain via PGCAN's encoder.}
    \label{fig comparison}
\end{figure*}

The comparison of the topological structures from our PIGP framework against those from SIMP is shown in \Cref{fig comparison}. We employ the sensitivity filters \citep{sigmund_99_2001} with radius 2 and 4 to address mesh-dependency while controlling the complexity of structures in SIMP. In our approach, we can control the structural complexity via $\Res$ and so consider two values ($36$ and $18$) for it. The corresponding feature grids for all examples are displayed in \Cref{fig comparison} next to the obtained topologies by our method.

As shown in \Cref{fig comparison}, the designed topologies by both approaches are quite similar in all cases. However, we observe that SIMP solutions tend to have higher gray fraction, especially when the filter size is set to 4. 
We also observe that both approaches can effectively control the complexity of the solutions by interpretable parameters (filter size in SIMP and number of cells in our approach's PGCAN) where smaller filter radius and larger $\Res$ values result in more intricate designs. 
We will further examine the effects of $\Res$ in \Cref{sec results sens}. 

\begin{table*}[!b]
    \centering
    \renewcommand{\arraystretch}{1.2}
    \smaller
    \setlength\tabcolsep{3pt}
    \caption{\textbf{Summary of performance:} Median values of compliance ($\mathrm{mJ}$), gray area fraction (\%), and computational time ($\mathrm{sec}$) are listed for comparison between SIMP and PIGP. The reported computational time for PIGP corresponds to the total runtime over 20,000 training epochs.}
    \begin{tabular}{l|ccc|ccc|ccc|ccc}
        \hline
        \multirow{2}{*}{\textbf{Example}}
        & \multicolumn{3}{c|}{\textbf{SIMP} ($\Filter$ $2$)} 
        & \multicolumn{3}{c|}{{ \textbf{SIMP} ($\Filter$ $4$)}} 
        & \multicolumn{3}{c|}{\textbf{PIGP} ($\Res$ $36$)} 
        & \multicolumn{3}{c}{{ \textbf{PIGP} ($\Res$ $18$)}} \\
        \cline{2-13}
        & Compliance & Gray & Time 
        & { Compliance} & { Gray} & { Time} 
        & Compliance & Gray & Time 
        & { Compliance} & { Gray} & { Time} \\
        \hline
        MBB              & { 1.891} & { 10.1} & { 16.9 } & { 1.994} & { 13.8} & { 13.6 } & { 1.924} & { 8.1} & { 549} & { 1.949} & { 9.4} & { 552} \\
        Cantilever       & { 0.712} & { 5.2 } & { 29.8} & { 0.739} & { 8.3 } & { 13.6} & { 0.709} & { 3.5} & { 765} & { 0.736} & { 3.4} & { 756} \\
        Uniformly loaded & { 0.284} & { 8.9} & { 39.7} & { 0.448} & { 15.4} & { 28.9} & { 0.243} & { 7.9} & { 798} & { 0.247} & { 9.7} & { 809} \\
        L-shape          & { 1.757} & { 5.9 } & { 14.3 } & { 1.874} & { 10.7 } & { 10.2} & { 1.759} & { 2.9} & { 609} & { 1.784} & { 4.5} & { 671} \\
        Hollow           & { 0.528} & { 4.2 } & { 15.2 } & { 0.538} & { 7.9 } & { 12.9 } & { 0.530} & { 5.8} & { 921} & { 0.535} & { 7.2} & { 924} \\
        \hline
    \end{tabular}
    \label{tab summary perform}
\end{table*}

We report the median values of compliance ($\mathrm{mJ}$), gray area fraction ($\%$), and computation time ($\mathrm{sec}$) in Table {\color{blue}3} across the 10 repetitions for each example using SIMP ($\Filter$ $2$ and $4$) and PIGP ($\Res$ $36$ {and $18$) as comparison. Compliance values for both methods are computed using the FEM solver from \citep{andreassen_efficient_2011} prior to binarization. PIGP ($\Res$ $36$) achieves lower compliance than SIMP ($\Filter$ $2$) in the cantilever and uniformly loaded beams, but slightly higher values in the other three examples. PIGP ($\Res$ $18$) consistently outperforms SIMP ($\Filter$ $4$) with lower compliance. Overall, compliance values from PIGP are closely aligned with those from the SIMP method with the exception of SIMP ($\Filter$ $4$) in the Uniformly loaded beam. For more details, compliance values before and after dynamic binarization are also reported in \Cref{appendix proj}.

As a density-based method, our PIGP framework also exhibits some gray regions but it typically achieves lower gray area fractions than SIMP. For instance, the median value for the cantilever beam is $8.1\%$ in PIGP ($\Res$ $36$) versus $10.1\%$ using SIMP ($\Filter$ $2$). The only exception is the hollow beam, where SIMP attains $4.2\%$ compared to $5.8\%$ from our approach. Moreover, SIMP ($\Filter$ $4$) consistently yields higher gray area fractions than PIGP at both feature resolutions ($\Res$ $36$ and $18$). Overall, our PIGP framework provides a more effective treatment of gray areas than SIMP, without requiring additional modifications such as sharpness continuation scheme or iterative thresholding procedure. However, our approach is slower than SIMP which leverages the self-adjoint nature of the problem to achieve significant speedups. Reducing computational costs is part of our future works.

Statistical details of the compliance values for all five examples are summarized in \Cref{tab stats comp}. We observe that our approach yields quite comparable median and mean compliance values  with those from SIMP, with the most notable difference corresponding to the uniformly loaded beam example. The standard deviations in our results are slightly higher across most cases, indicating a greater design variability. Nevertheless, the relatively small standard deviations compared to the mean compliance values suggests that the randomness introduced by parameter initialization and our adaptive grid scheme is quite small.

\begin{table*}[!h]
    \centering
    \renewcommand{\arraystretch}{1.5}
    \small
    \setlength\tabcolsep{5pt}
    \caption{\textbf{Statistical comparison of compliance:} Median, mean, standard deviation, maximum, and minimum values for compliance ($\mathrm{mJ}$) are reported for SIMP ($\Filter$ $2$) and PIGP ($\Res$ 36).}
    \begin{tabular}{l|ccccc|ccccc}
        \hline
        \multirow{2}{*}{\textbf{Example}} 
        & \multicolumn{5}{c|}{\textbf{SIMP} ($\Filter$ $2$)} 
        & \multicolumn{5}{c}{\textbf{PIGP} ($\Res$ $36$)} \\
        \cline{2-11}
        & median & mean & std & max & min 
        & median & mean & std & max & min \\
        \hline
        MBB        &  1.891 &  1.892 &  $8.3\times10^{-3}$ &  1.906 &  1.878 &  1.924 &  1.929 &  $2.7\times10^{-2}$ &  1.989 &  1.898 \\
        Cantilever     &  0.712  &  0.712  &  $5.6\times10^{-3}$ &  0.719  &  0.700 &  0.709  &  0.717  &  $1.7\times10^{-2}$ &  0.759  &  0.704 \\
        Uniformly loaded     &  0.284 &  0.284 &  $5.2\times10^{-3}$ &  0.294 &  0.277 &  0.243 &  0.243 &  $2.1\times10^{-3}$ &  0.246 &  0.239\\
        L-shape  &  1.757 &  1.757 &  $2.3\times10^{-3}$ &  1.761 &  1.755 &  1.759 &  1.762 &  $7.5\times10^{-3}$ &  1.773 &  1.752 \\
        Hollow  &  0.528 &  0.529 &  $2.8\times10^{-3}$ &  0.535 &  0.528 &  0.530 &  0.531 &  $1.8\times10^{-3}$ &  0.534 & 0.528 \\
        \hline
    \end{tabular}
    \label{tab stats comp}
\end{table*}

To gain more insights into the performance and training dynamics of our approach, we visualize the loss histories and topology evolutions in \Cref{fig loss 1,fig evolution}, respectively (as loss curves are quite similar across the five cases, we only provide them for the MBB beam example). 
The curves for $L_P(\cdot)$ and $C_1^2(\cdot)$ with more fluctuations represent raw loss values recorded at each epoch, while the smoother dark-colored curves are the corresponding moving averaged versions. All three loss terms follow a similar trend: they rapidly increase early in the training, peaking around the 1000\textsuperscript{th} epoch, followed by a gradual decrease toward their respective steady-state values. The potential energy loss $L_P(\cdot)$ remains strictly positive which is due to the regularization term $\tau(\cdot)$. The volume fraction constraint is effectively enforced as $C_1^2(\cdot)$ is consistently on the order of $10^{-5}$ after $5,000$ of training epochs.
\begin{figure*}[!t]
    \centering
    \begin{subfigure}[t]{0.32\textwidth}
        \includegraphics[width=\linewidth]{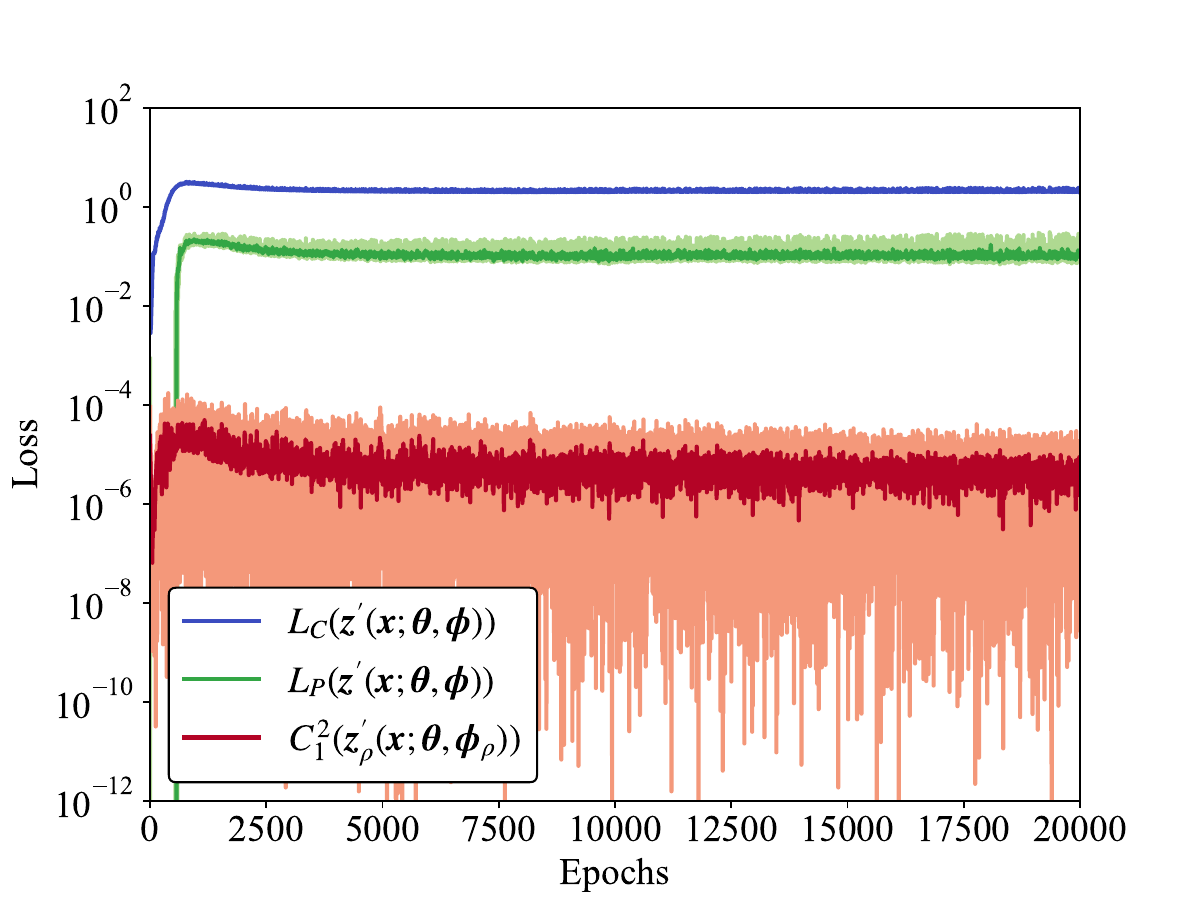}
        \vspace{-11em}
        \captionsetup{justification=raggedright, singlelinecheck=false, skip=-3.5pt, position=top}
        \caption[]{}
        \label{fig loss 1}
    \end{subfigure}
    \begin{subfigure}[t]{0.32\textwidth}
        \includegraphics[width=\linewidth]{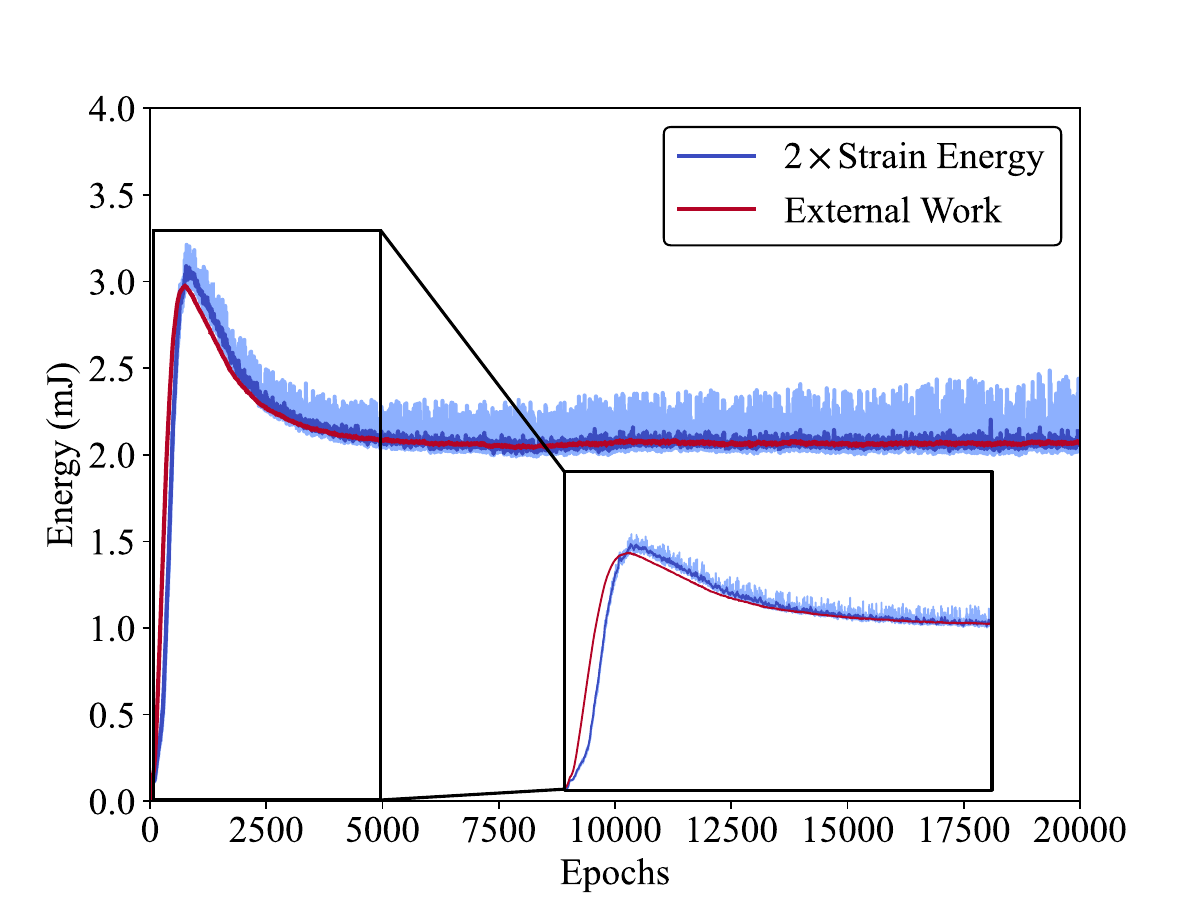}
        \vspace{-11em}
        \captionsetup{justification=raggedright, singlelinecheck=false, skip=-3.5pt, position=top}
        \caption[]{}
        \label{fig loss 2}
    \end{subfigure}
    \begin{subfigure}[t]{0.32\textwidth}
        \includegraphics[width=\linewidth]{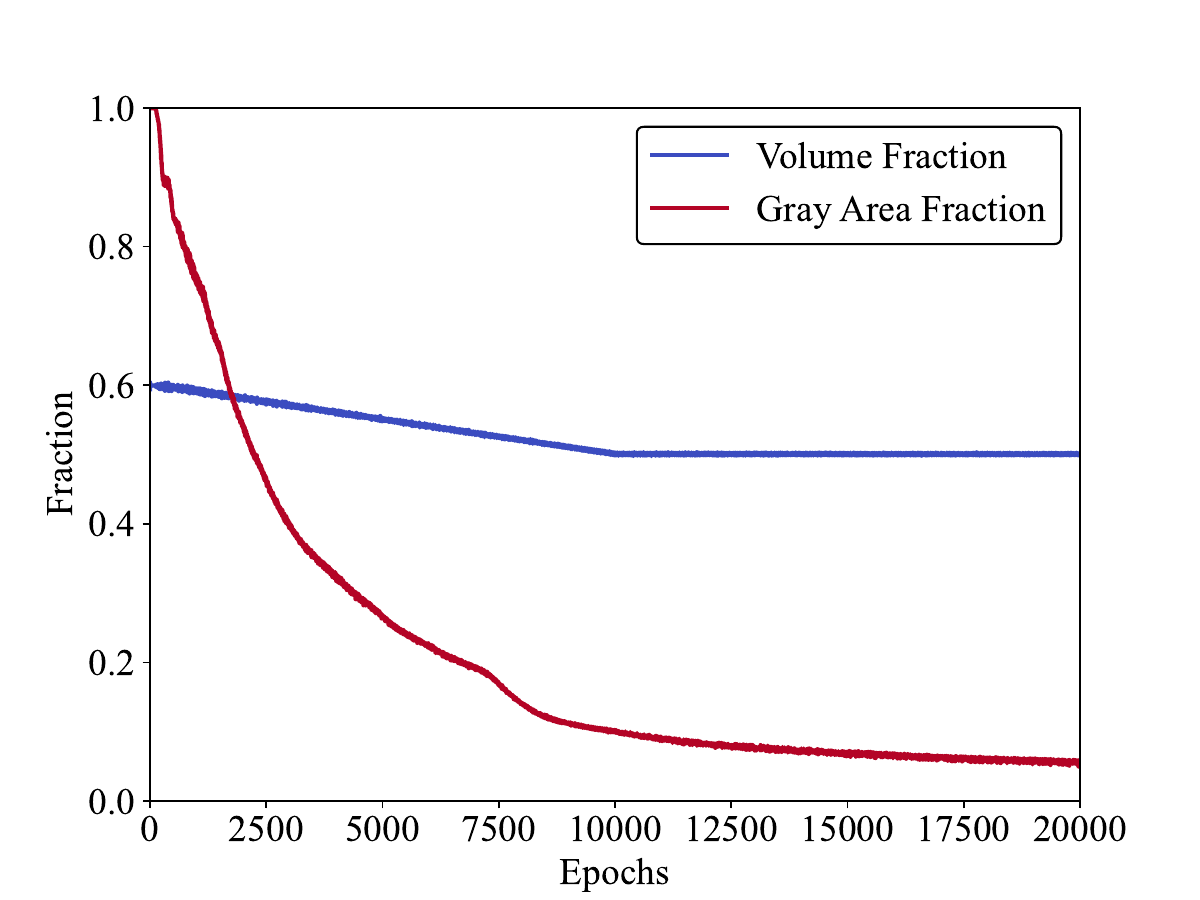}
        \vspace{-11em}
        \captionsetup{justification=raggedright, singlelinecheck=false, skip=-3.5pt, position=top}
        \caption[]{}
        \label{fig loss 3}
    \end{subfigure}
    \caption{\textbf{Training dynamics in the MBB example:} (a) evolution of the three loss terms, (b) convergence of the equilibrium condition shown by the agreement between $2\times$strain energy and the external work, (c) evolution of volume fraction and the gray area fraction. Similar trajectories are observed for other four benchmark examples.}
    \label{fig loss}
\end{figure*}

\begin{figure*}[!t]
    \centering
    \begin{subfigure}[t]{0.9\textwidth}
        \includegraphics[width=\linewidth]{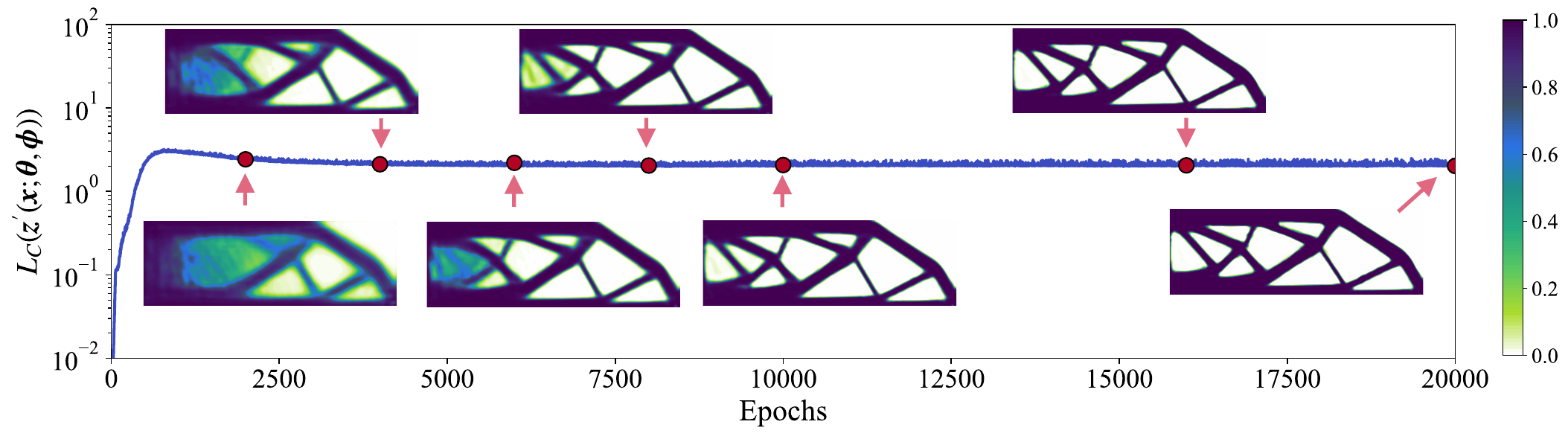}
        \vspace{-12em}
        \captionsetup{justification=raggedright, singlelinecheck=false, skip=-3.5pt, position=top}
        \caption[]{}
        \label{fig dynamics 1}
    \end{subfigure}
    \begin{subfigure}[t]{0.9\textwidth}
        \includegraphics[width=\linewidth]{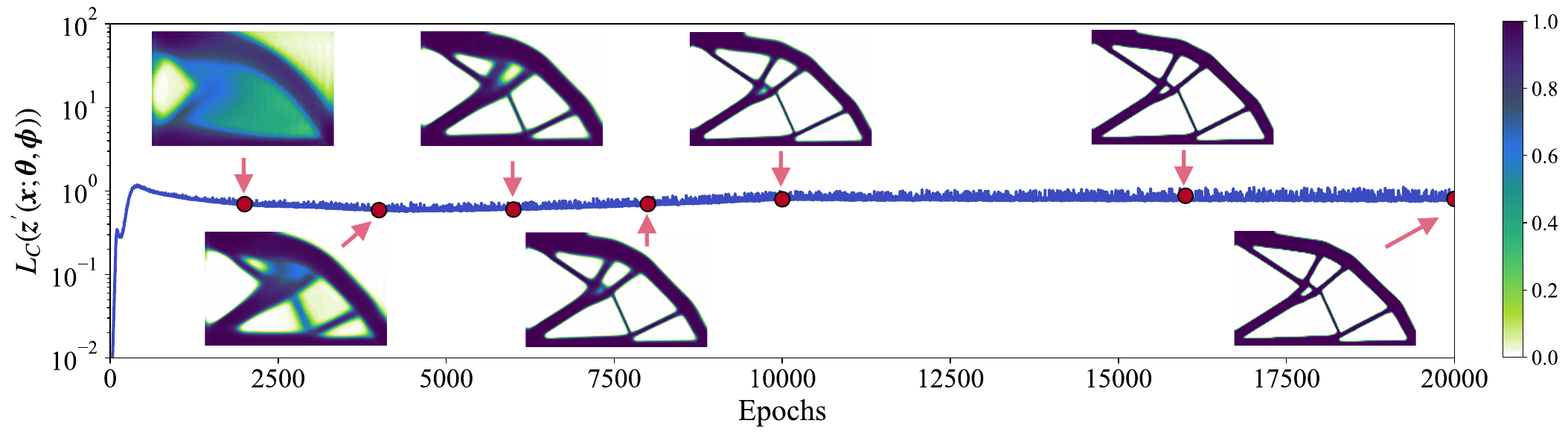}
        \vspace{-12em}
        \captionsetup{justification=raggedright, singlelinecheck=false, skip=-3.5pt, position=top}
        \caption[]{}
        \label{fig dynamics 2}
    \end{subfigure}
    \begin{subfigure}[t]{0.9\textwidth}
        \includegraphics[width=\linewidth]{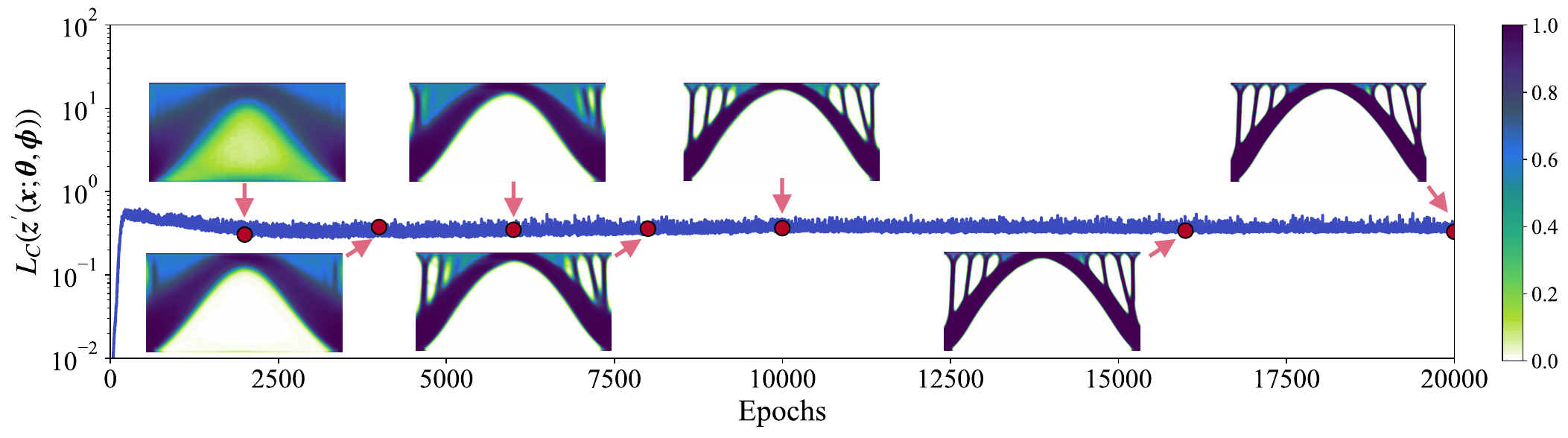}
        \vspace{-12em}
        \captionsetup{justification=raggedright, singlelinecheck=false, skip=-3.5pt, position=top}
        \caption[]{}
        \label{fig dynamics 3}
    \end{subfigure}
    \begin{subfigure}[t]{0.9\textwidth}
        \includegraphics[width=\linewidth]{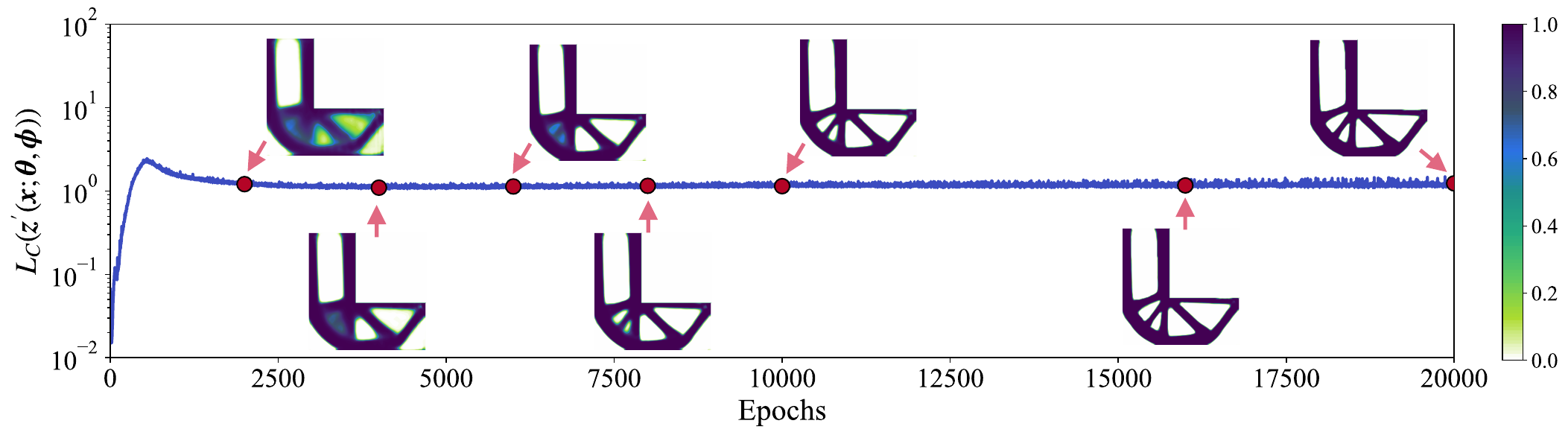}
        \vspace{-12em}
        \captionsetup{justification=raggedright, singlelinecheck=false, skip=-3.5pt, position=top}
        \caption[]{}
        \label{fig dynamics 4}
    \end{subfigure}
    \begin{subfigure}[t]{0.9\textwidth}
        \includegraphics[width=\linewidth]{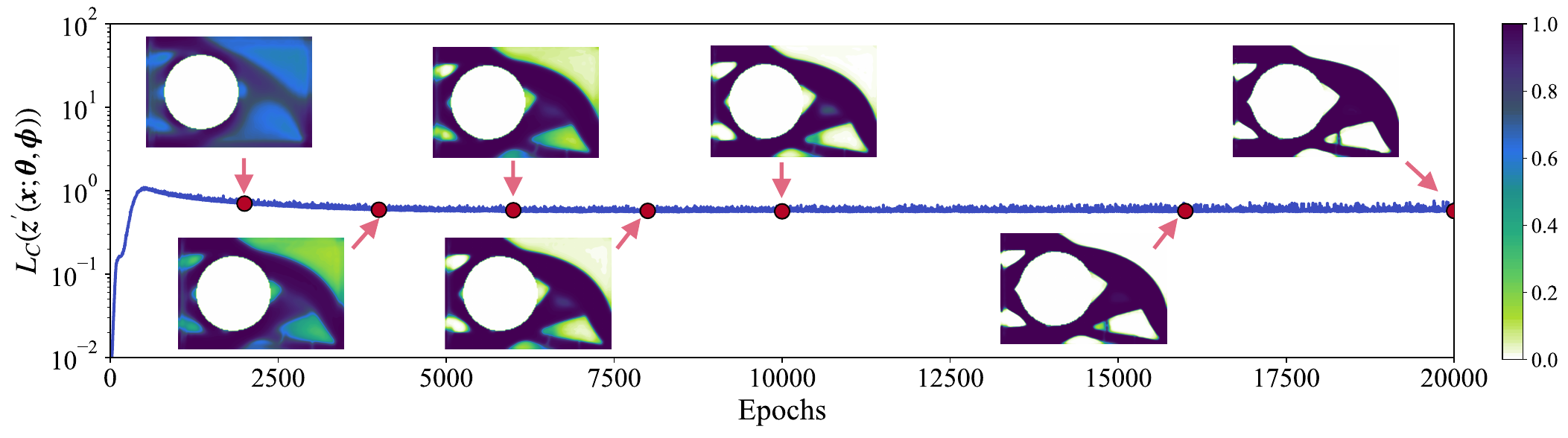}
        \vspace{-12em}
        \captionsetup{justification=raggedright, singlelinecheck=false, skip=-3.5pt, position=top}
        \caption[]{}
        \label{fig dynamics 5}
    \end{subfigure}
    \caption{\textbf{Topology evolution throughout training:} The median topologies at selected training epochs are visualized with respect to the objective $L_C(\cdot)$ for (a) MBB beam, (b) cantilever beam, (c) uniformly loaded beam, (d) L-shaped beam, and (e) hollow beam.}
    \label{fig evolution}
\end{figure*}

We can assess the force balance condition by examining the energy history curves shown in \Cref{fig loss 2}.
The strain energy and external work both initially rise from zero to their respective peaks, a trend that is similar to the loss trajectories. Small energy values indicate that the initial displacement fields are very small but these values rapidly increase as the optimization traverses through nontrivial displacement solutions. 
We observe a small mismatch between the two energy curves up to 2500 epochs, indicating that the force-balance condition has not yet been satisfied. In the following epochs, the curves gradually decrease and both converge to $\sim2 ,\mathrm{mJ}$ within $10{,}000$ epochs. The achieved force-balance constraint is an essential condition for obtaining accurate displacement fields and topologies with smooth structural features, which are consistently established once equilibrium is reached and maintained throughout the remainder of training. We emphasize that the displacement solutions from PIGP are accurate for the optimized topologies, as demonstrated by the comparison with FEM results in \Cref{appendix displacement}.

As shown in \Cref{fig loss 3}, the volume fraction history closely follows the curriculum training strategy of \Cref{sec method training} in the first 50\% of epochs, and remains constant thereafter. Concurrently, the gray area fraction drops rapidly during the first half of training and stays below $10\%$ thereafter. This efficient reduction in gray regions further highlights the effectiveness of our PIGP framework in generating binary, well-defined topologies with minimal intermediate regions. We note that CM problems can be highly non-convex due to added features such as density projection continuation in SIMP or curriculum training in our simultaneous PIGP framework which remains stable as evidenced by the smooth and gradual evolution of loss histories and topology structures.

We present the evolution of the topologies for each example at selected training epochs in \Cref{fig evolution} (the topologies correspond to the median runs). We observe that in the early stage of training (around the 2000\textsuperscript{th} epoch), the density field exhibits a large gray area with some discernible design features. A distinct topology emerges around the 4000\textsuperscript{th} epoch in all examples, marking the onset of meaningful material redistribution. As training progresses, the overall structural layout stabilizes and the intermediate gray regions are gradually eliminated. By the end of training, the density field approaches a nearly binary distribution. Notably, high-quality topologies are often obtained with 10,000 epochs, indicating that the total computational cost can be reduced by half by adopting appropriate early stopping criteria.

\begin{figure*}[!t]
    \centering
    \begin{subfigure}[t]{0.9\textwidth}
        \captionsetup{justification=raggedright, singlelinecheck=false, skip=-1pt, position=top}
        \caption{Solutions from FEM:}
        \includegraphics[width=\linewidth]{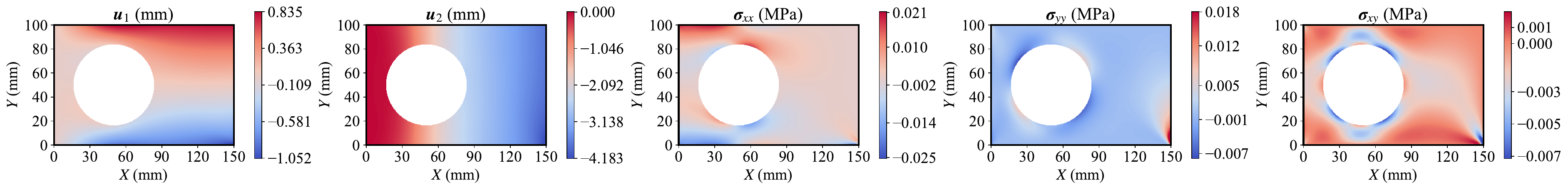}
        \label{fig dem contour 1}
    \end{subfigure}
    \vspace{-1.2em}

    \begin{subfigure}[t]{0.9\textwidth}
        \captionsetup{justification=raggedright, singlelinecheck=false, skip=-1pt, position=top}
        \caption{DEM solutions and error distributions using PGCAN:}
        \includegraphics[width=\linewidth]{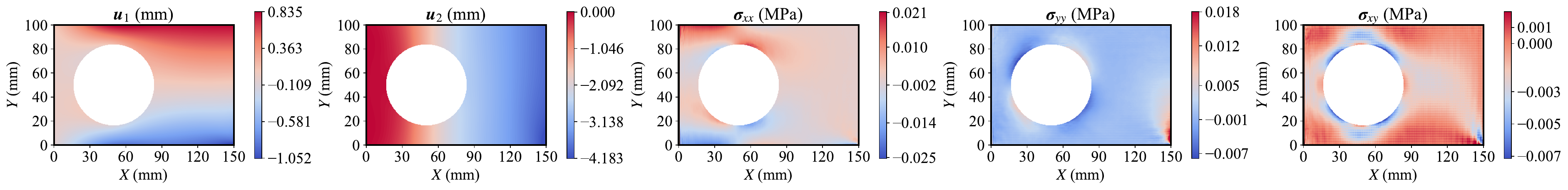}
        \includegraphics[width=\linewidth]{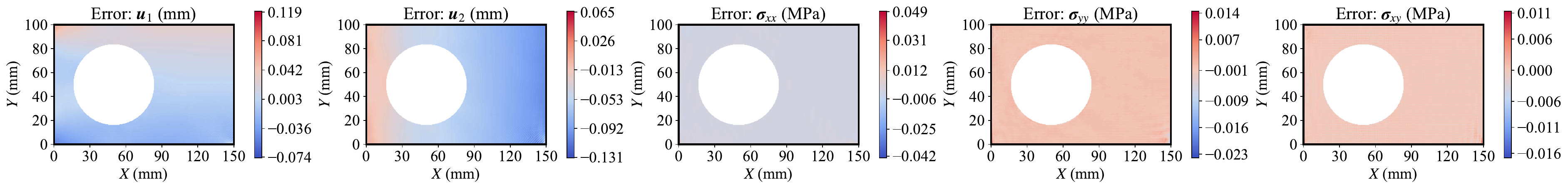}
        \label{fig dem contour 2}
    \end{subfigure}
    \vspace{-1.2em}

    \begin{subfigure}[t]{0.9\textwidth}
        \captionsetup{justification=raggedright, singlelinecheck=false, skip=-1pt, position=top}
        \caption{DEM solutions and error distributions using MLP:}
        \includegraphics[width=\linewidth]{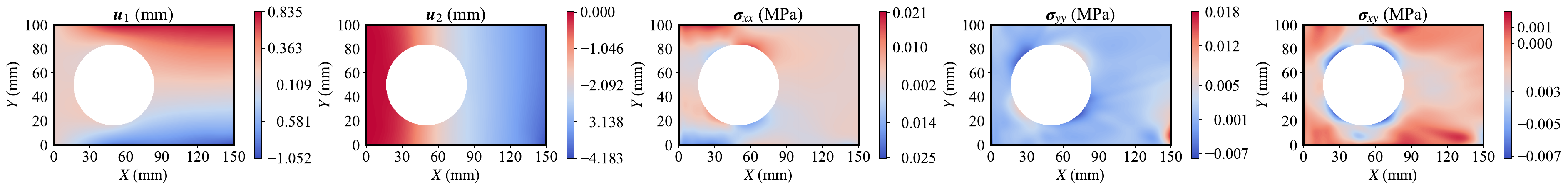}
        \includegraphics[width=\linewidth]{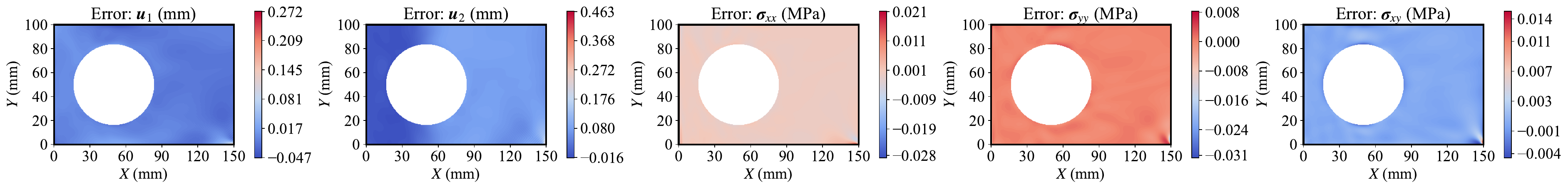}
        \label{fig dem contour 3}
    \end{subfigure}
    \vspace{-1.2em}

    \caption{\textbf{Displacement and stress fields:} (a) FEM reference solutions, (b) DEM solutions using PGCAN, and (c) DEM solutions using MLP. The corresponding error distributions relative to FEM are also presented.}
    \label{fig dem contour}
\end{figure*}

\subsection{Effects of PGCAN} \label{sec results dem}
To illustrate the effect of PGCAN compared to MLP, we first obtain the displacement and stress fields via the DEM (i.e., topology is fixed and not optimized). The potential energy $L_P(\cdot)$ is the only loss term, and the density field $\rho(\cdot)$ is uniformly prescribed to $1$ in the entire domain except in the hole. FEM reference solutions of displacements and stress components, along with the DEM solutions using PGCAN and MLP are shown in \Cref{fig dem contour}. Overall, both ML-based models produce displacement and stress fields that are comparable to the FEM reference but the displacement error distributions obtained from PGCAN are consistently lower than those from MLP. Due to the localized mechanism enforced by the feature encoder, PGCAN captures sharper and more realistic stress localizations than MLP near the external loading point or along the hole boundary.

The stress error plots in \Cref{fig dem contour} may indicate that an MLP may outperform PGCAN in TO. To reject this statement, we visualize the energy histories in \Cref{fig dem history} where PGCAN and MLP stabalize and achieve equilibrium at epochs 200 and 1000, respectively. This large difference is due to PGCAN's ability in capturing stress concentrations much faster than MLPs which suffer from spectral bias. 

\begin{figure*}[!t]
    \centering
    \includegraphics[width = 0.4\textwidth]{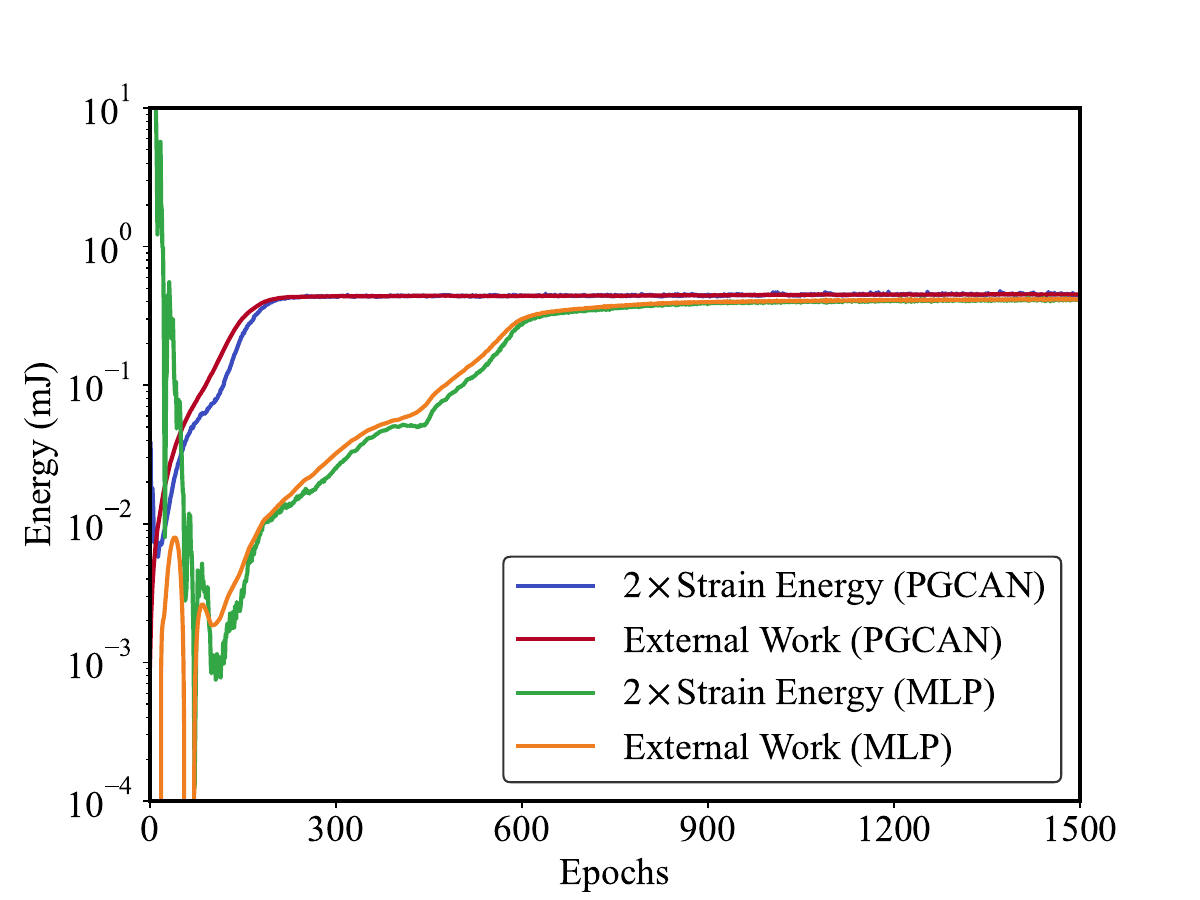}
    \caption{\textbf{Comparison of convergence rate to force balance condition:} Evolution curves of $2\times$strain energy and external work in the DEM using PGCAN are compared with those obtained using the MLP.}
    \label{fig dem history}
\end{figure*}

To further demonstrate the superiority of PGCAN over MLPs, we visualize the CM results for the uniformly loaded beam in \Cref{fig uniform beam}. PGCAN effectively suppresses gray regions and yields a well-defined structure within 8000 epochs, with the gray area further minimized after 20,000 epochs. PGCAN captures localized design features with higher complexity and significantly reduced gray regions beneath the applied load. In contrast, the MLP model fails to produce a discernible topology until the end of training and even then provides an overly simplified structure with a very high gray area fraction. This observation is consistent with the findings of \citet{chandrasekhar_tounn_2021}, where MLPs were unable to solve this example due to the persistence of a large gray area fraction. Although ML-based TO can achieve some mesh independence without explicit filtering, owing to the tendency of MLPs to favor low-frequency patterns, additional control mechanisms such as the $\Res$ parameter or length-scale filters \citep{chandrasekhar_approx_2022} are needed to capture fine-scale features.

\begin{figure*}[!b]
    \centering
    \begin{subfigure}[t]{0.4\textwidth}
        \includegraphics[width=\linewidth]{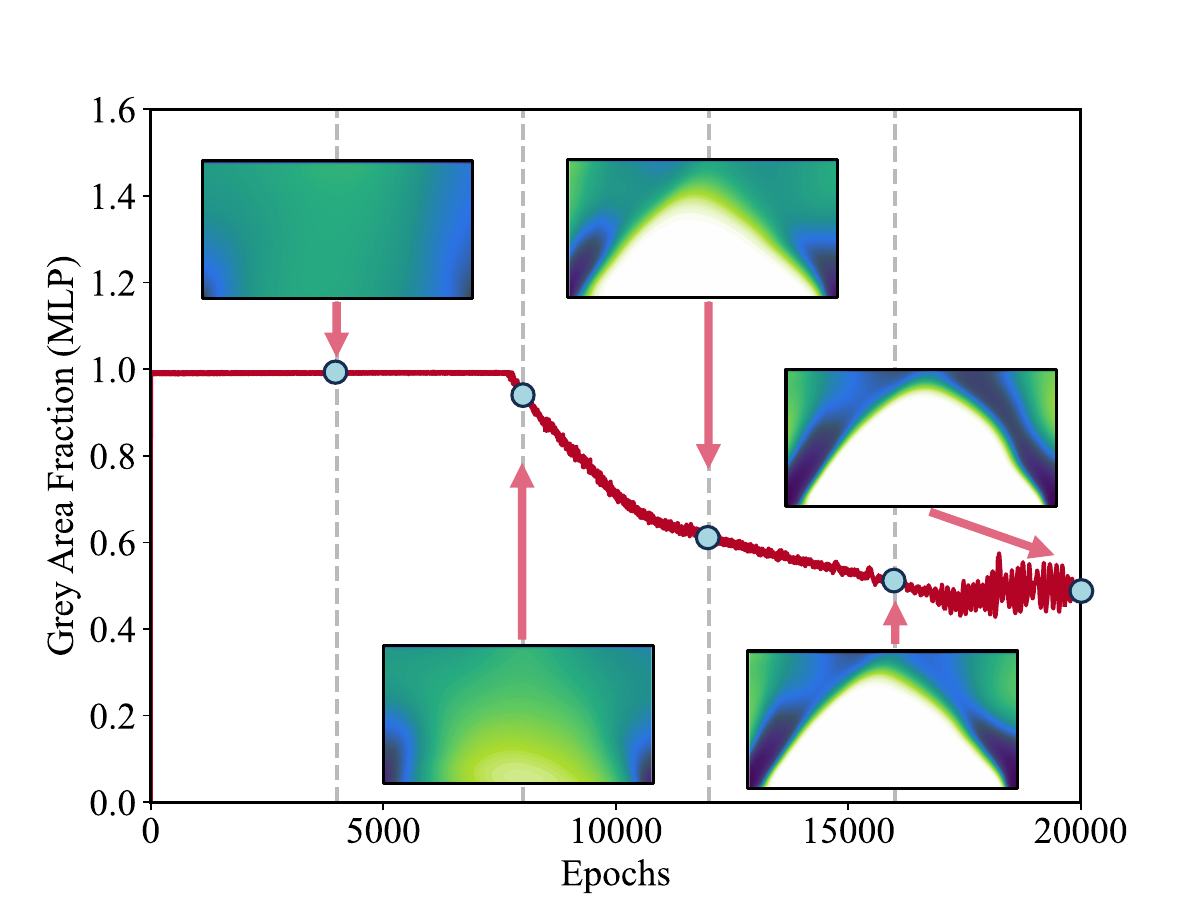}
        \vspace{-14em}
        \captionsetup{justification=raggedright, singlelinecheck=false, skip=-3.5pt, position=top}
        \caption[]{}
        \label{fig uniform beam 1}
    \end{subfigure}
    \begin{subfigure}[t]{0.4\textwidth}
        \includegraphics[width=\linewidth]{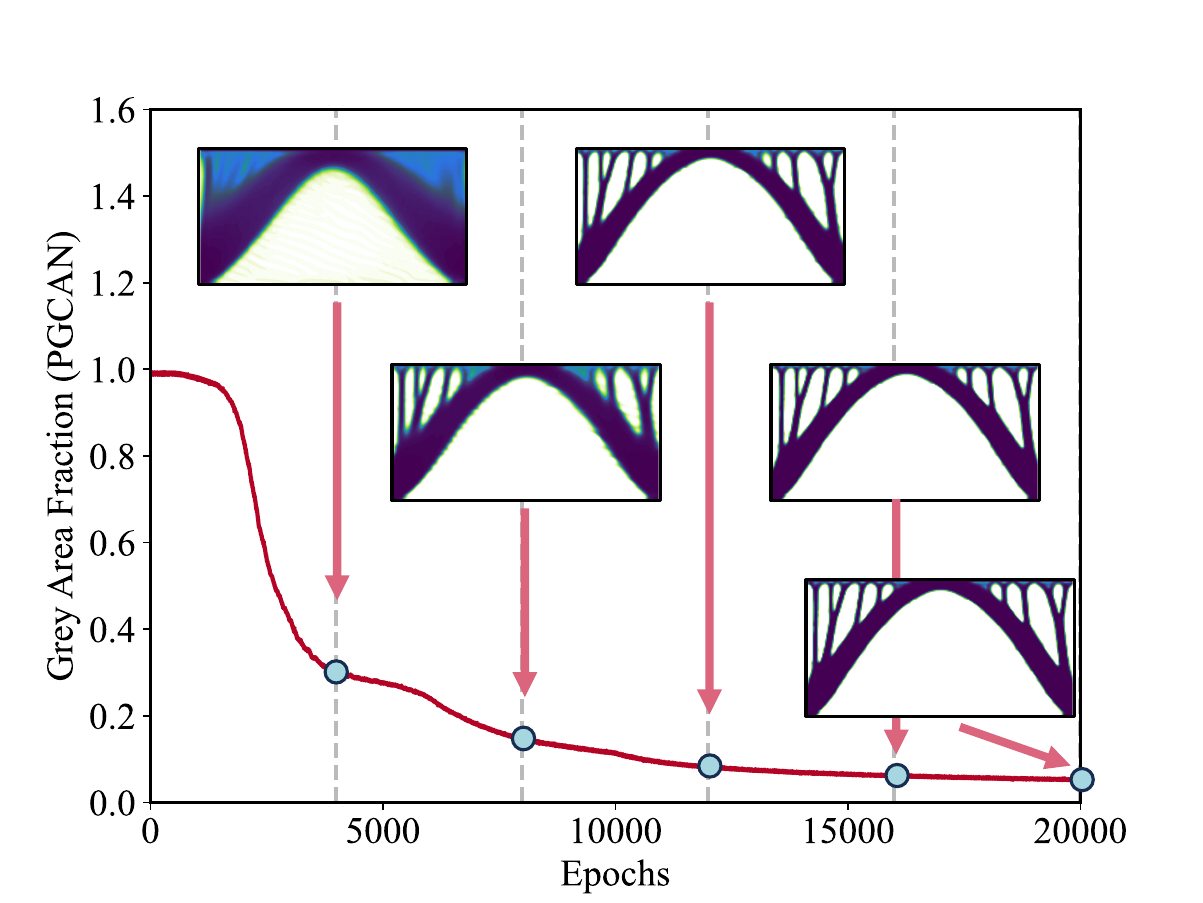}
        \vspace{-14em}
        \captionsetup{justification=raggedright, singlelinecheck=false, skip=-3.5pt, position=top}
        \caption[]{}
        \label{fig uniform beam 2}
    \end{subfigure}
    \caption{\textbf{Topology evolution for the uniformly loaded beam:} Results from (a) MLP and (b) PGCAN with inset figures showing density distributions at selected epochs. PGCAN outperforms MLP in resolving fine design features.}
    \label{fig uniform beam}
\end{figure*}

\subsection{Sensitivity Studies} \label{sec results sens}
To evaluate the robustness of our framework, we examine its sensitivity to three factors: (1) the resolution parameter $\Res$ in PGCAN, (2) the adaptive grid scheme, and (3) the curriculum training strategy introduced in \Cref{sec method training}. 
To assess the effect of $\Res$ on the resulting topologies, we consider the MBB beam example in \Cref{fig res}, where $\Res$ is varied from 18 to 54. For comparison, we also include SIMP ($\Filter$ $2$, $3$ and $4$) results initialized with random density fields. As shown in \Cref{fig res 1}, increasing $\Res$ leads to more intricate topological features. This behavior is attributed to the denser parameterization of the initial feature field $\Fb_0$ at higher $\Res$, which enhances the model’s ability to resolve finer spatial localizations.

\begin{figure*}[!b]
    \centering
    \begin{subfigure}[t]{0.7\textwidth}
        \includegraphics[width=\linewidth]{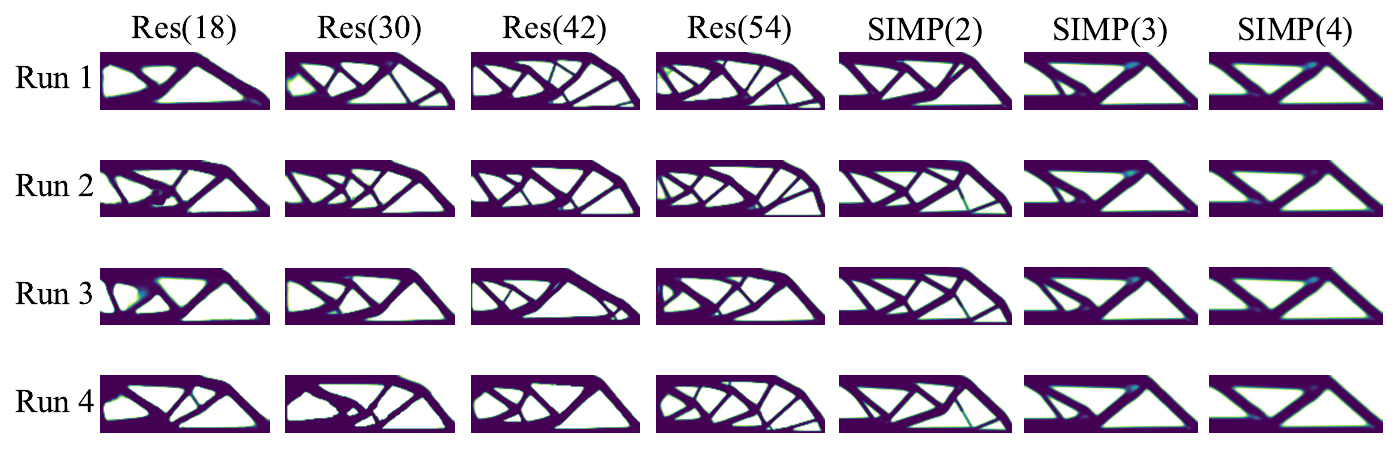}
        \vspace{-11.2em}
        \captionsetup{justification=raggedright, singlelinecheck=false, skip=-3.5pt, position=top}
        \caption[]{}
        \label{fig res 1}
    \end{subfigure}
    \begin{subfigure}[t]{0.7\textwidth}
        \includegraphics[width=\linewidth]{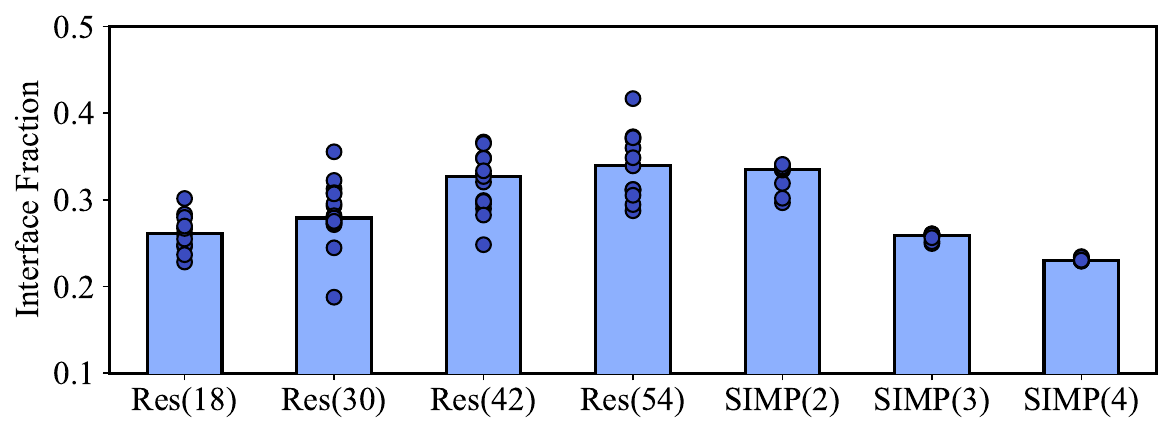}
        \vspace{-12.5em}
        \captionsetup{justification=raggedright, singlelinecheck=false, skip=-3.5pt, position=top}
        \caption[]{}
        \label{fig res 2}
    \end{subfigure}
    \begin{subfigure}[t]{0.7\textwidth}
        \includegraphics[width=\linewidth]{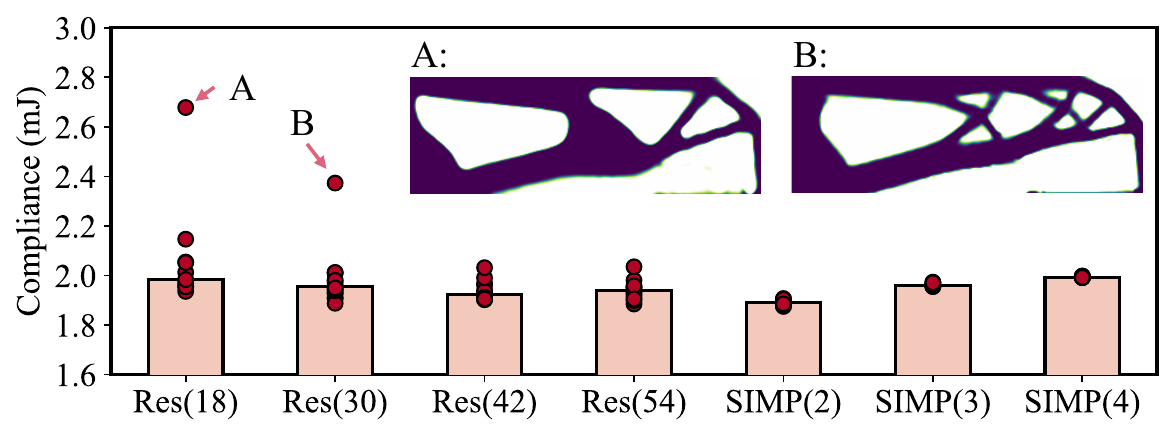}
        \vspace{-12.5em}
        \captionsetup{justification=raggedright, singlelinecheck=false, skip=-3.5pt, position=top}
        \caption[]{}
        \label{fig res 3}
    \end{subfigure}
    \caption{\textbf{Effect of $\Res$ on the MBB beam example:} (a) Final topologies from PGCAN with $\Res$ ranging from 18 to 54, compared to SIMP results with varying filter radii; (b) comparison of interface fractions; and (c) comparison of compliance values. Outlier cases with elevated compliance at $\Res = 18$ and $\Res = 30$ are highlighted in the inset figures. The bar plots are made based on the median values across all repetitions.}
    \label{fig res}
\end{figure*}

To quantify the contribution of $\Res$ to structural complexity, we use  median interface fractions as the metric. As shown in \Cref{fig res 2}, the median interface fraction is directly correlated to $\Res$. The discrete points on this plot show that our PIGP framework produces structures spanning a wider range of interface fractions for each $\Res$, including both simple and intricate designs. This variability is expected due to the inherent randomness in our ML-based approach from the initialization of $\thetab$ and the adaptive grid strategy. In contrast, the interface fractions from SIMP locate within a narrower range for each filter size due to its deterministic formulation, where the only source of randomness is the initial density field.

We next compare the compliance values in \Cref{fig res 3} which shows that most topologies generated by our approach achieve median compliance values around $2.0$, comparable to those from SIMP. However, two outlier cases exhibit significantly higher compliance, corresponding to suboptimal designs shown in the inset figures. These outliers reflect the broader design space explored by PIGP, which can yield less favorable solutions.

\begin{figure*}[!b]
    \centering
    \begin{subfigure}[t]{0.7\textwidth}
        \includegraphics[width=\linewidth]{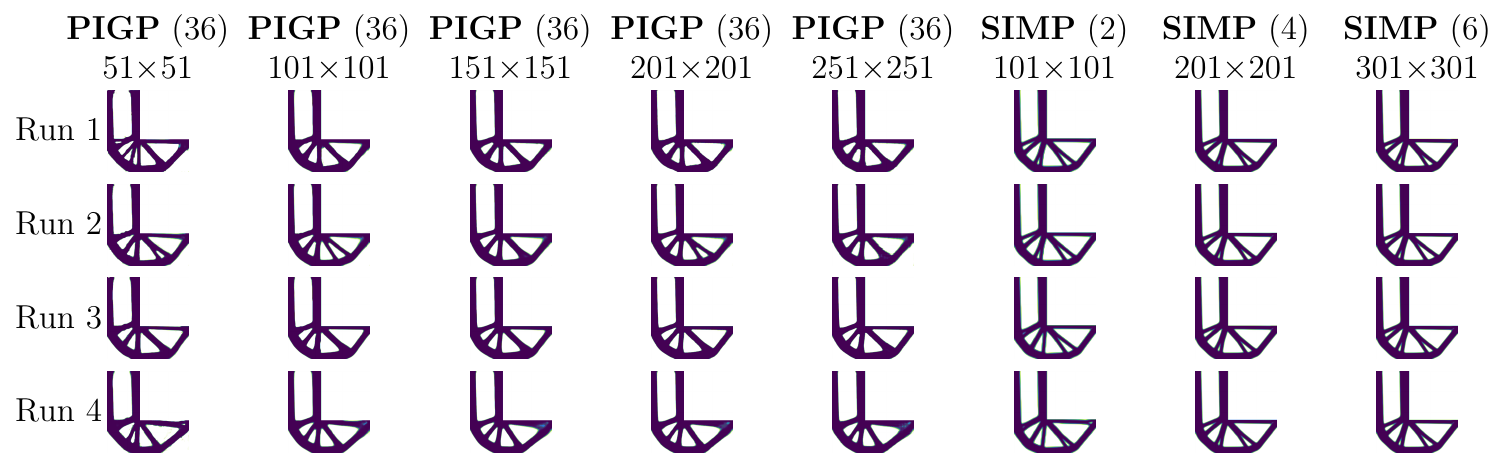}
        \vspace{-11.0em}
        \captionsetup{justification=raggedright, singlelinecheck=false, skip=-3.5pt, position=top}
        \caption[]{}
        \label{fig grid 1}
    \end{subfigure}
    \begin{subfigure}[t]{0.7\textwidth}
        \includegraphics[width=\linewidth]{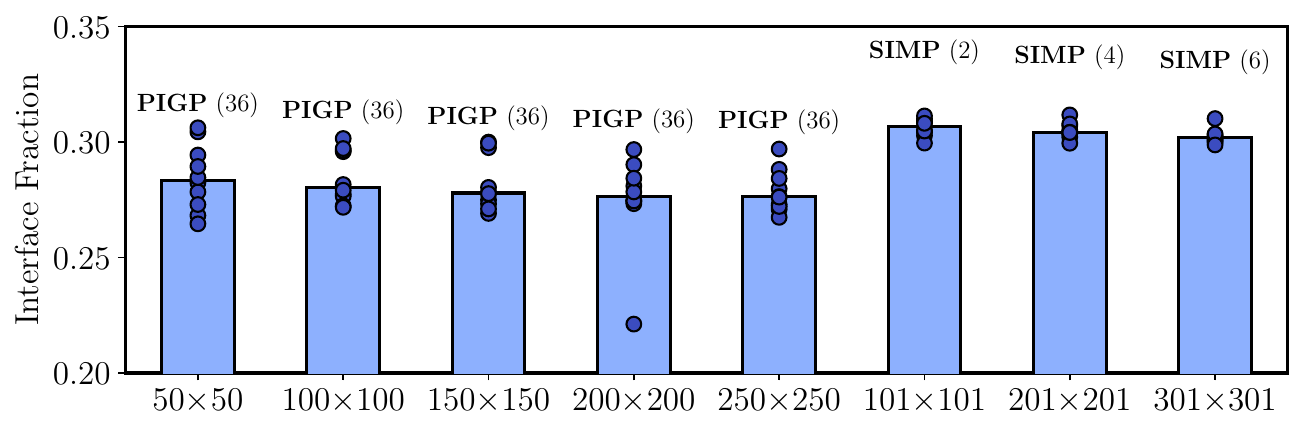}
        \vspace{-11.5em}
        \captionsetup{justification=raggedright, singlelinecheck=false, skip=-3.5pt, position=top}
        \caption[]{}
        \label{fig grid 2}
    \end{subfigure}
    \begin{subfigure}[t]{0.7\textwidth}
        \includegraphics[width=\linewidth]{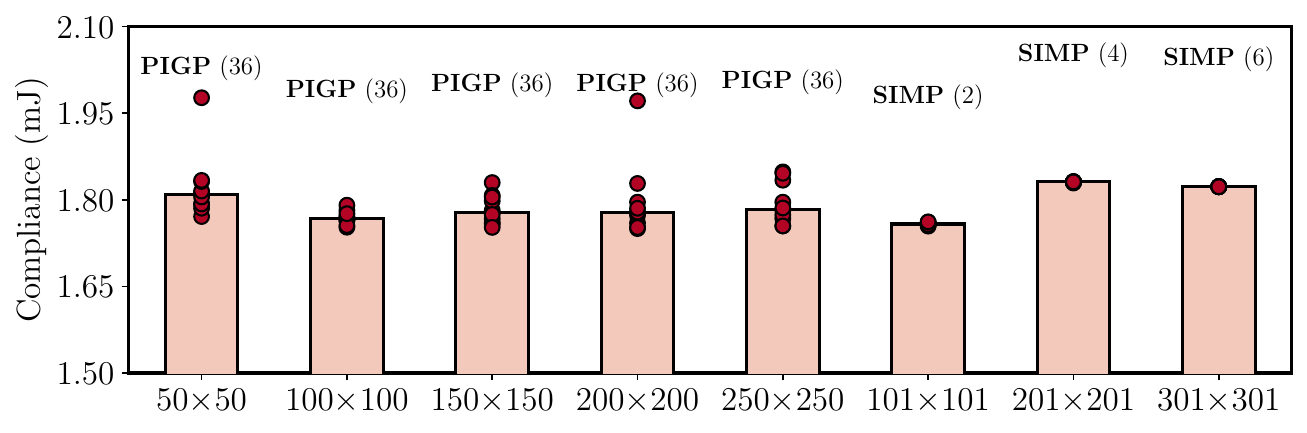}
        \vspace{-11.5em}
        \captionsetup{justification=raggedright, singlelinecheck=false, skip=-3.5pt, position=top}
        \caption[]{}
        \label{fig grid 3}
    \end{subfigure}
    \caption{\textbf{Effect of grid resolutions for the L-shape beam:} (a) Topologies using adaptive grid scheme with corresponding SIMP results; (b) comparison of interface fractions; and (c) compliance values across grid configurations. The bar plots are based on the median values.}
    \label{fig grid}
\end{figure*}

To examine the sensitivity to the adaptive grid scheme, we use the L-shape beam example with different sets of dynamic grids $\Sigmab$. The results are summarized in \Cref{fig grid 1} where SIMP baselines are also included for comparison. For each case, the coarse grid dimensions $(N_x^{(1)}, N_y^{(1)})$ are indicated in the subfigure titles, and the corresponding fine grid $(N_x^{(n_g)}, N_y^{(n_g)})$ doubles the cell resolution in both the $x$- and $y$-directions. For the SIMP cases, we consider three finite element grids $(N_x, N_y)$: $(101,101)$, $(201,201)$, and $(301,301)$, with scaled filter radius of $2$, $4$, and $6$, respectively.

\Cref{fig grid 1} shows that the topologies produced by our PIGP framework are insensitive to adaptive grid resolutions. This grid-independence is further supported by the median interface fraction results in \Cref{fig grid 2}, which remain nearly constant across all PIGP cases. SIMP can also achieve similar mesh-independence by scaling the filter radius with the grid resolution. The median compliance values shown in \Cref{fig grid 3}, indicate that our PIGP framework yields {\color{blue}quite comparable} and more broadly distributed compliance (most between $1.9$ and $2.0$) compared to SIMP which provides values that are concentrated around $1.9$. To further assess grid resolution effects, we compare the optimized topologies and computational time of our framework with those from SIMP using the cantilever beam example in \Cref{appendix grid sensitivity}.

\begin{figure*}[!b]
    \centering
    \begin{subfigure}[t]{0.7\textwidth}
        \includegraphics[width=\linewidth]{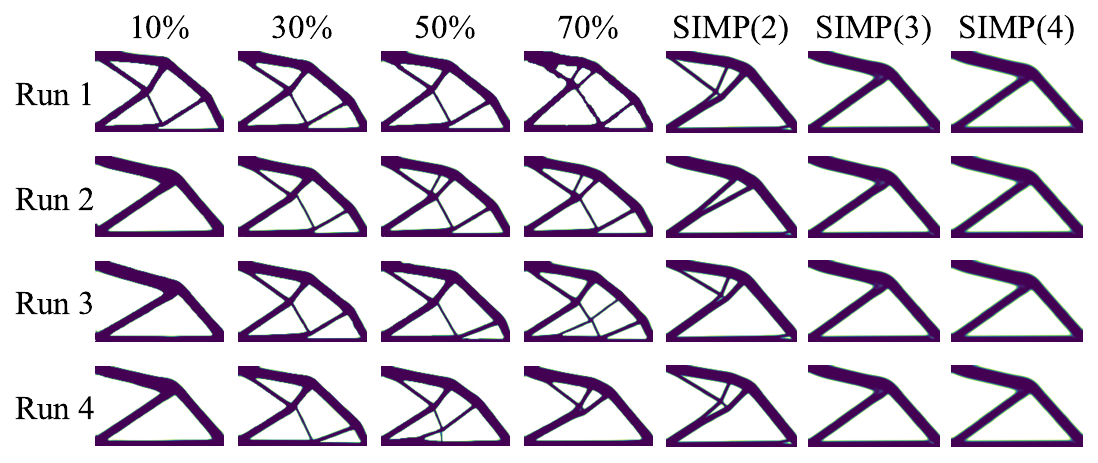}
        \vspace{-14.0em}
        \captionsetup{justification=raggedright, singlelinecheck=false, skip=-3.5pt, position=top}
        \caption[]{}
        \label{fig vf 1}
    \end{subfigure}
    \begin{subfigure}[t]{0.7\textwidth}
        \includegraphics[width=\linewidth]{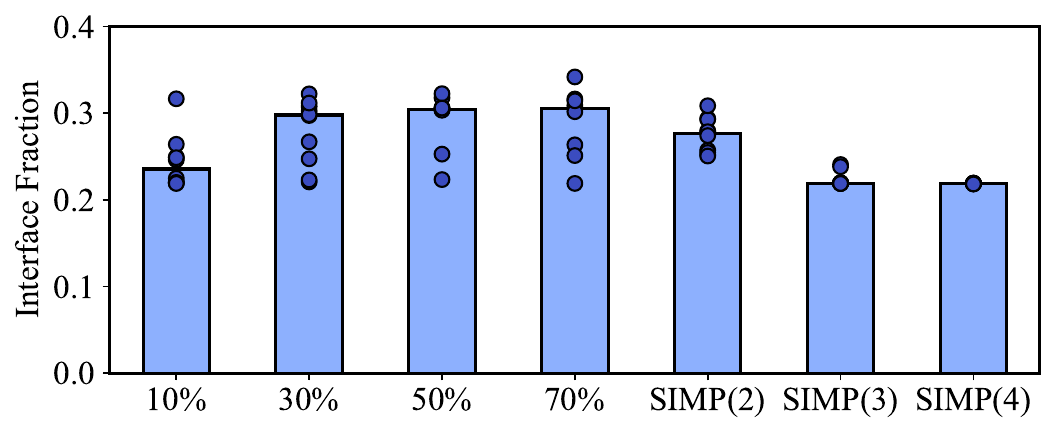}
        \vspace{-13.8em}
        \captionsetup{justification=raggedright, singlelinecheck=false, skip=-3.5pt, position=top}
        \caption[]{}
        \label{fig vf 2}
    \end{subfigure}
    \begin{subfigure}[t]{0.7\textwidth}
        \includegraphics[width=\linewidth]{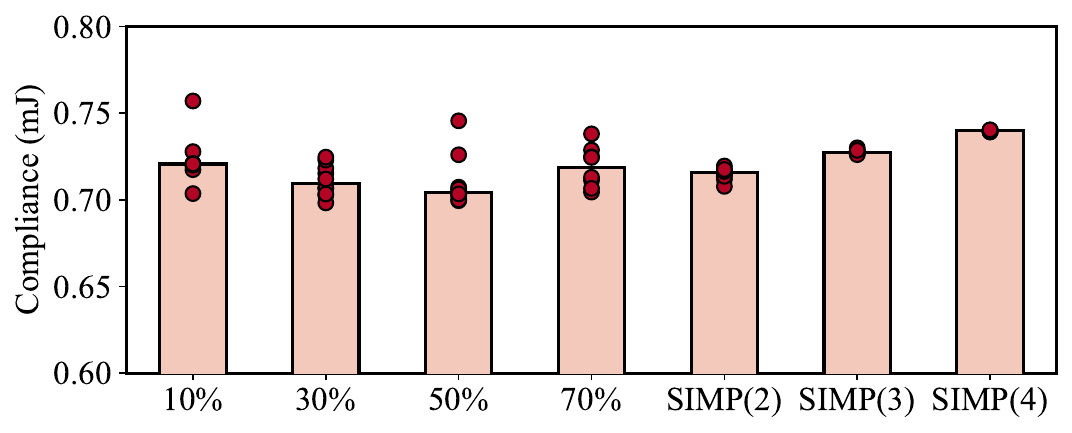}
        \vspace{-13.6em}
        \captionsetup{justification=raggedright, singlelinecheck=false, skip=-3.5pt, position=top}
        \caption[]{}
        \label{fig vf 3}
    \end{subfigure}
    \caption{\textbf{Effect of curriculum training with scheduled volume fraction:} (a) Final topologies for $\gamma$ values ranging from 10\% to 70\%, shown alongside SIMP results; (b) comparison of interface fractions; and (c) corresponding compliance values. The bar plots use the median values.}
    \label{fig vf}
\end{figure*}

Another factor that may influence the performance of our PIGP framework is the curriculum training schedule applied to the volume fraction constraint. We define the portion of the epochs where volume fraction is scheduled to decrease using $\gamma$ in \Cref{eq vf schedule}. To assess the impact of $\gamma$, we evaluate the cantilever beam example with values ranging from $10\%$ to $70\%$ as shown in \Cref{fig vf 1} along with the corresponding SIMP results.
We observe that rapid volume fraction reduction (e.g., $\gamma = 10\%$) tends to produce topologies with fewer fine-scale features, whereas more gradual reductions (e.g., $\gamma \geq 30\%$) better preserve structural complexity. This behavior is quantitatively captured in the interface fractions shown in \Cref{fig vf 2}. Additionally, increasing $\gamma$ from $30\%$ to $70\%$ does not noticeably affect the median interface fraction, indicating that our approach remains robust to the choice of $\gamma$ within this range.

We attribute the loss of fine features at small $\gamma$ to the competition between $C_1^2(\cdot)$ and $L_P(\cdot)$. During the early training epochs, force equilibrium is not accurately satisfied which compromises the accuracy of stress and strain predictions in the localized regions associated with fine structural features. Our curriculum training strategy mitigates this issue by gradually applying the volume fraction penalty; enabling the optimization to first focusing on converging to an accurate displacement solution and then satisfying the design constraint. As shown in \Cref{fig vf 3}, the median compliance values for $\gamma$ of $30\%$ and $50\%$ from our approach are slightly lower than those obtained using SIMP, consistent with the results reported in \Cref{tab stats comp}.

\subsection{Enforcing Design Features and Super-resolution} \label{sec results density}
As discussed in \Cref{sec method param}, the GP formulation in our framework enables the direct specification of the design feature $\tilde{\rhob}$ within designated regions of the domain. 
Additionally, our approach is mesh-free and provides a continuous representation of the state and design variables, i.e., we can visualize the fields with arbitrary resolution. 
To illustrate these capability, we present two examples by modifying the uniformly loaded beam and the hollow beam as shown in \Cref{fig density}. For the uniformly loaded beam, we require a solid thin beam along the top edge to support the applied distributed load while in the hollow beam a solid circular ring is concentrically embedded around the hole. 

\begin{figure*}[!b]
    \centering
    \includegraphics[width = 1\textwidth]{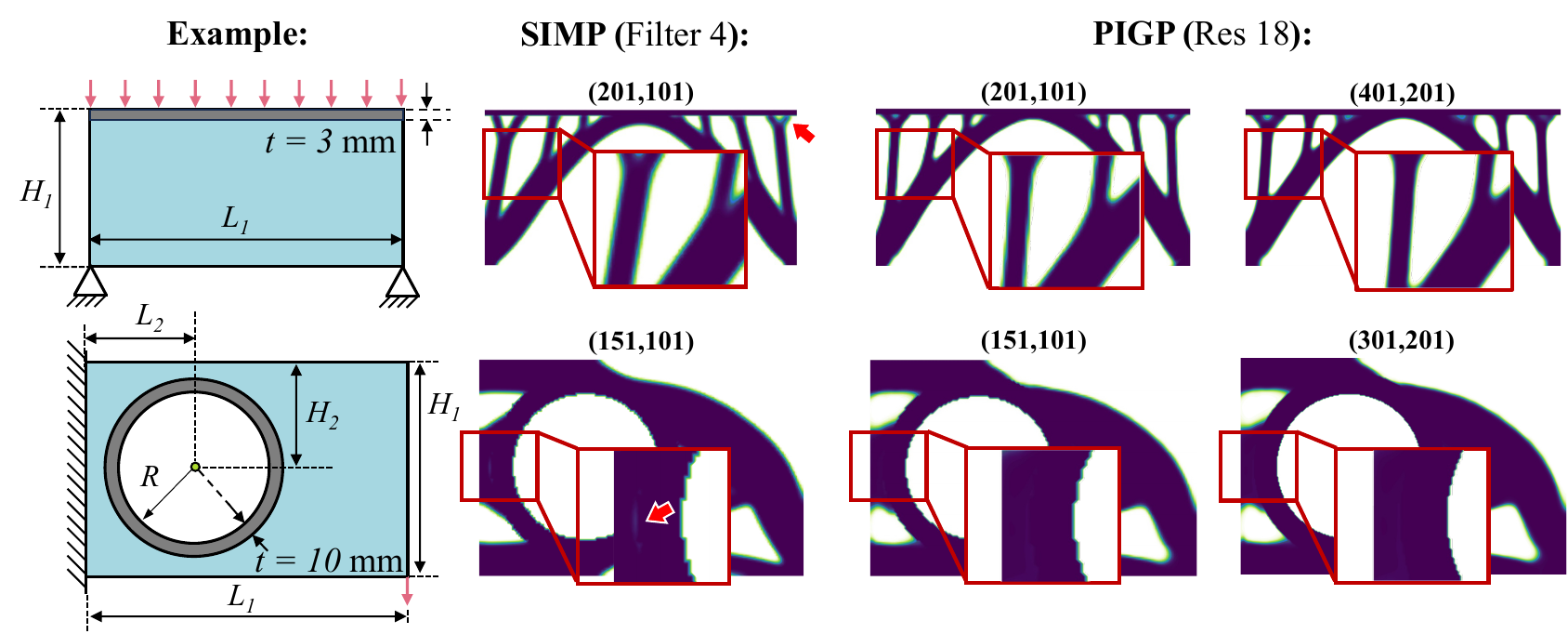}
    \caption{\textbf{Enforcing design features and super-resolution:} The topology results obtained from SIMP ($\Filter$ $4$) and PIGP ($\Res$ $18$) are shown for the modified uniformly loaded and hollow beams. The uniformly loaded beam is required to have a solid thin beam underneath the distributed load while the hollow beam has a solid ring around its hole. The dimensions for the added design features are included and the rest (along with $\psi_f$) can be found in \Cref{tab dimension}. Each figure includes a subtitle indicating the number of CPs ($N_x$, $N_y$) along the two spatial dimensions, which are positioned on the finite element mesh for SIMP, or on the structured grid for prediction in our PIGP framework. For our approach, we visualize the results (after training) at two grid resolutions to demonstrate its super-resolution capability (the pixels are artifacts of visualization).}
    \label{fig density}
\end{figure*}

We compare the topologies obtained from SIMP ($\Filter = 4$) with those generated using our PIGP framework with $\Res = 18$. Note that in SIMP, design features are enforced by assigning active or passive elements, whereas our PIGP framework achieves design constraints by sampling with GPs. The subtitles of the grid resolutions ($N_x$, $N_y$) in \Cref{fig density} indicate the number of CPs placed on the finite element mesh for SIMP, or on the structured grid used for \textit{prediction} in our approach which used the adaptive grid resolution as listed in \Cref{tab grid} during training.

For the uniformly loaded beam, SIMP produces large gray regions between the solid and void phases. In contrast, the structures generated by our approach have clean feature boundaries on both the ($201,101$) and ($401,201$) grids, with smoother surface for ($401,201$). A similar trend is observed in the hollow beam example, where the void edge becomes smoother when using the denser ($301,201$) grid to predict the density.
While we learn a continuous representation for $\rho(\xb; \thetab, \phib)$, SIMP directly updates a discrete density field defined on a fixed finite element mesh so it lacks the ability to naturally improve surface smoothness through resolution refinement.  
Across both examples we also observe that, unlike our approach, SIMP struggles to maintain strong connectivity across the regions where design constraints are enforced. In the uniformly loaded beam, there is weak connectivity between the top beam and the thin supporting structure. In the hollow beam example, SIMP solution shows a visible discontinuity at the interface between the solid ring and the surrounding material (see the areas pointed with the red arrows in \Cref{fig density}). 

\subsection{Comparison of Computational Cost}  \label{sec results cost}
To demonstrate the computational efficiency of our PIGP framework, we compare its runtime with several representative methods using the cantilever beam as the benchmark example. Specifically, we compare against: (1) SIMP implemented in MATLAB \citep{andreassen_efficient_2011}, (2) Topology Optimization using Neural Networks (TOuNN) \citep{chandrasekhar_tounn_2021}, (3) the Deep Energy Method for Topology Optimization (DEMTOP) \citep{he_deep_TO_2023}, and (4) the Complete Physics-Informed Neural Network for Topology Optimization (CPINNTO) \citep{jeong_complete_2023}. Notably, all the four reference methods adopt a similar \textit{nested} optimization strategy. It is worth noting that SIMP and TOuNN rely on FEM to compute the displacement fields, whereas DEMTOP and CPINNTO employ the DEM.

\begin{figure*}[!b]
    \centering
    \begin{subfigure}[t]{0.45\textwidth}
        \includegraphics[width=\linewidth]{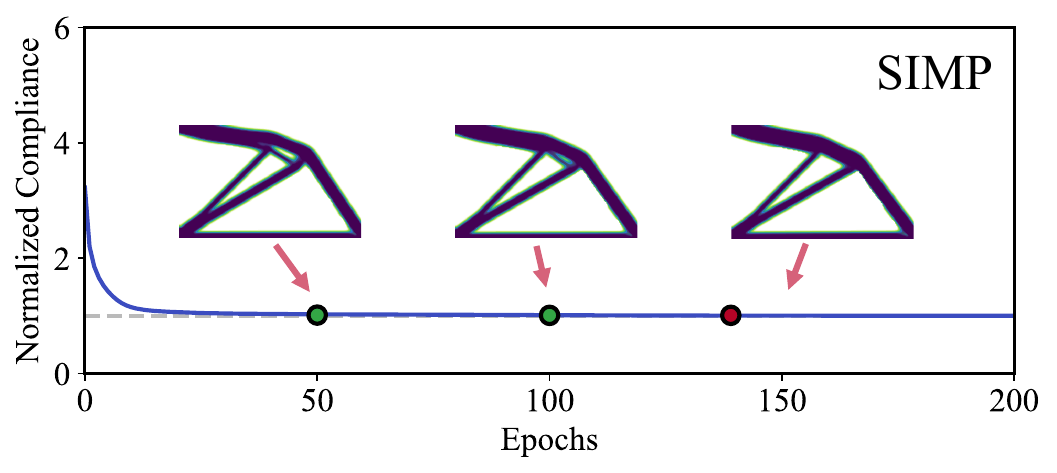}
        \vspace{-11.0em}
        \captionsetup{justification=raggedright, singlelinecheck=false, skip=-3.5pt, position=top}
        \caption[]{}
        \label{fig cost 1}
    \end{subfigure}
    \begin{subfigure}[t]{0.45\textwidth}
        \includegraphics[width=\linewidth]{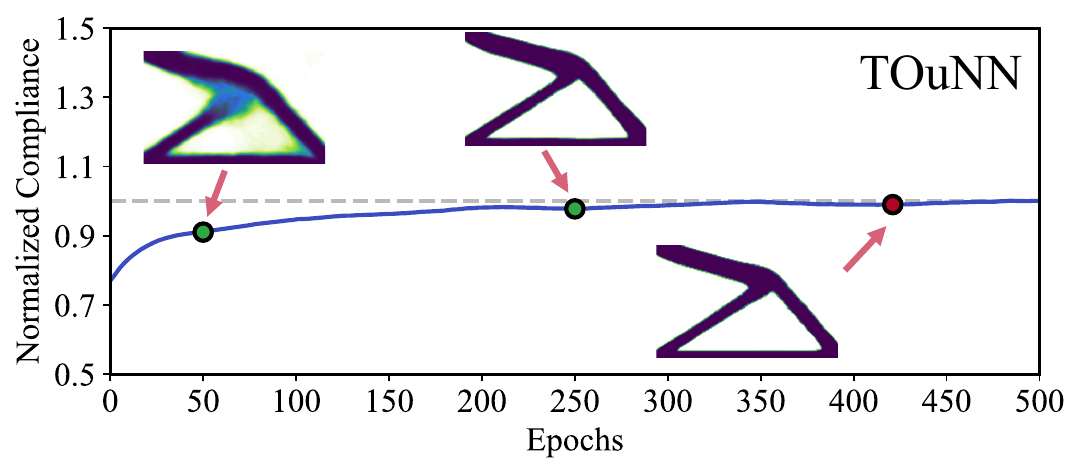}
        \vspace{-11.0em}
        \captionsetup{justification=raggedright, singlelinecheck=false, skip=-3.5pt, position=top}
        \caption[]{}
        \label{fig cost 2}
    \end{subfigure}
    \begin{subfigure}[t]{0.45\textwidth}
        \includegraphics[width=\linewidth]{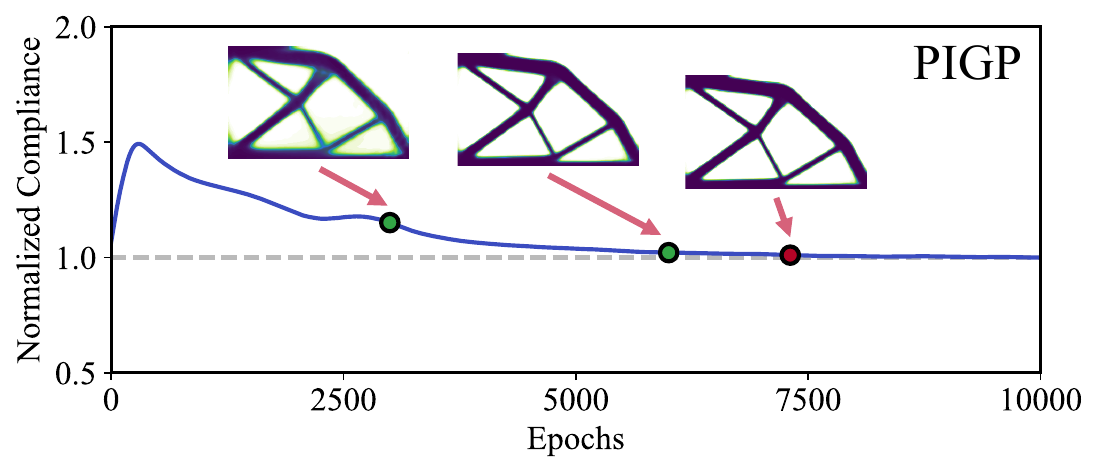}
        \vspace{-11.0em}
        \captionsetup{justification=raggedright, singlelinecheck=false, skip=-3.5pt, position=top}
        \caption[]{}
        \label{fig cost 3}
    \end{subfigure}
    \begin{subfigure}[t]{0.45\textwidth}
        \includegraphics[width=\linewidth]{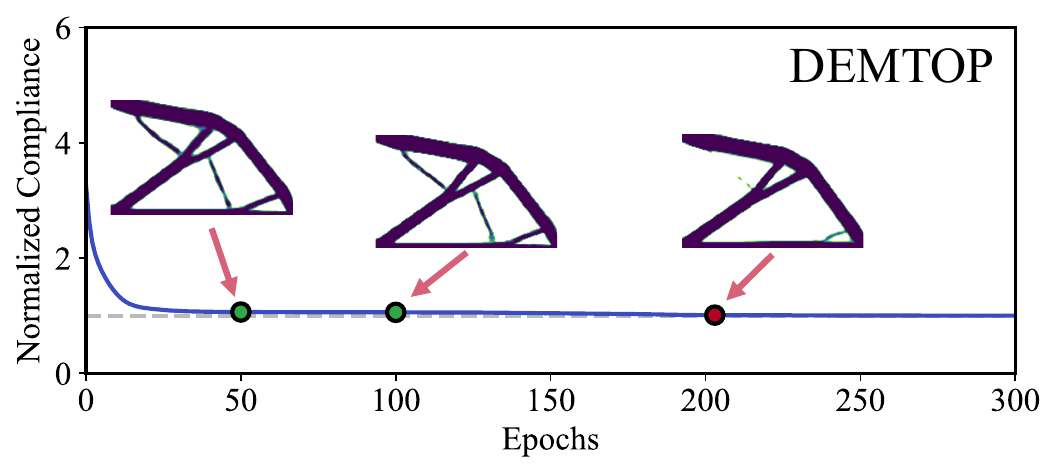}
        \vspace{-11.0em}
        \captionsetup{justification=raggedright, singlelinecheck=false, skip=-3.5pt, position=top}
        \caption[]{}
        \label{fig cost 4}
    \end{subfigure}
    \begin{subfigure}[t]{0.45\textwidth}
        \includegraphics[width=\linewidth]{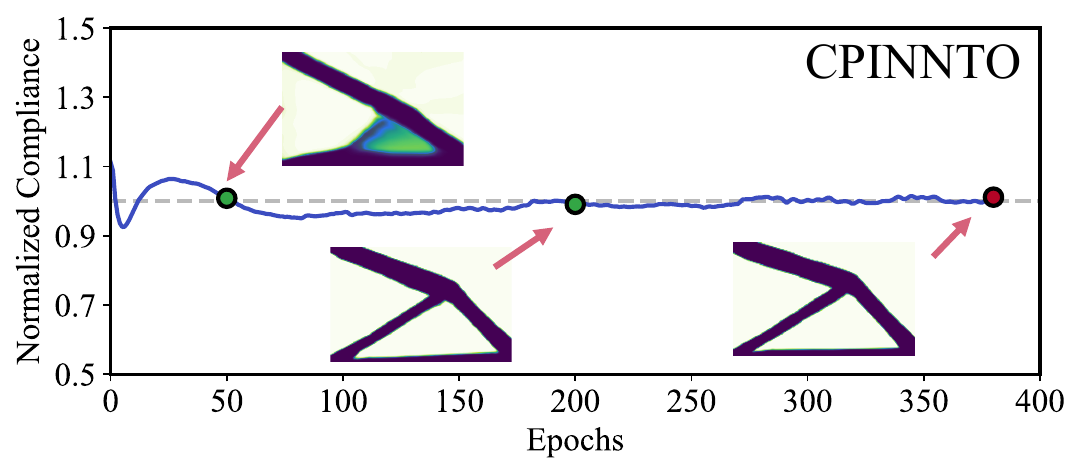}
        \vspace{-11.0em}
        \captionsetup{justification=raggedright, singlelinecheck=false, skip=-3.5pt, position=top}
        \caption[]{}
        \label{fig cost 5}
    \end{subfigure}
    \begin{subfigure}[t]{0.45\textwidth}
        \hspace{-0.6em}
        \includegraphics[width=\linewidth]{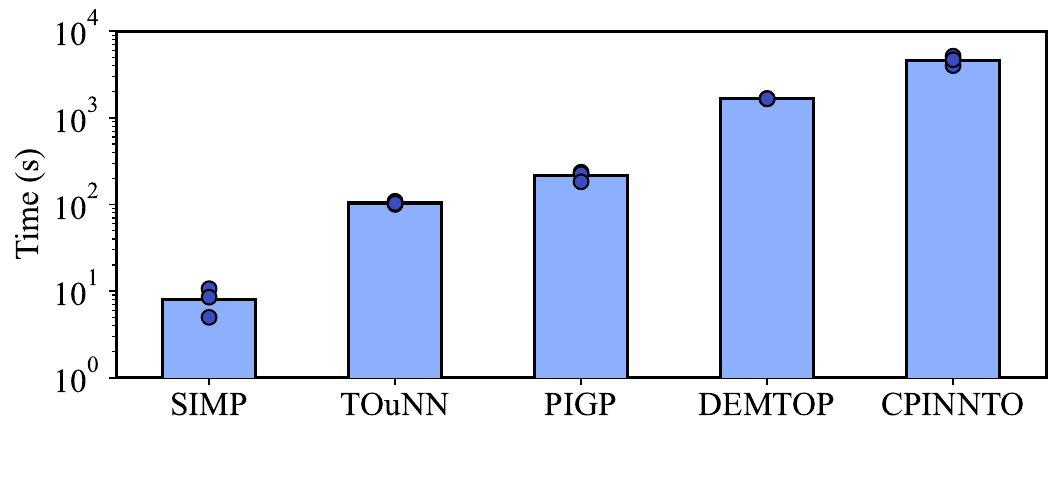}
        \vspace{-9.7em}
        \captionsetup{justification=raggedright, singlelinecheck=false, skip=-3.5pt, position=top}
        \caption[]{}
        \label{fig cost 6}
    \end{subfigure}
    \caption{\textbf{Comparison of computational cost for the cantilever beam example:} Normalized compliance histories for (a) SIMP, (b) TOuNN, (c) PIGP, (d) DEMTOP, and (e) CPINNTO, with intermediate and final topologies shown at selected epochs. (f) presents the total computational time at the point (red marker) where the final design criterion is met.}
    \label{fig cost}
\end{figure*}

To ensure a fair comparison, each reference method is implemented using a comparable mesh size and CP density. We use an adaptive grid scheme with the median intermediate grid matching that used in the other methods. Our PIGP framework and all other ML-based approaches are trained on the same GPU device. The SIMP baseline is implemented on CPU and GPU acceleration is unnecessary.

Due to differences in training objectives and optimization strategies, it is difficult to determine a consistent point at which the final optimal design is achieved across all methods. For instance, SIMP terminates when the maximum change in density between iterations falls below a specified threshold. However, this method is not directly applicable to ML-based approaches where local density variations can be large unless very small learning rates are used.

To ensure a consistent basis for comparison of computational time, we define a unified final design criterion based on normalized compliance histories. Each method is run for a sufficiently large number of epochs and the resulting compliance curve is smoothed using a moving average filter. The final smoothed compliance value is used to normalize the entire history, further reducing fluctuations. The final design and effective runtime are determined at the first epoch where the normalized compliance reaches $1.01 $ or $0.99$, and stay within this range thereafter.

The normalized compliance curves for all five methods, each evaluated with three repetitions, are shown in \Cref{fig cost 1}–\Cref{fig cost 5}, along with the intermediate (green markers) and final (red markers) topologies. Among these methods, SIMP and DEMTOP exhibit smoother compliance histories, which can be explained with the use of deterministic solvers such as MMA or OC. It is evident that high-quality designs are often achieved well before the final design criterion is met for all methods.

The total computational times required by each method at the final design point are summarized in Fig. {\color{blue}14f}. As expected, SIMP is the most efficient, completing in approximately $10$ seconds. TOuNN and our PIGP framework exhibit comparable runtimes, despite TOuNN relying on an efficient 2D FEM implementation to solve the displacement fields. Although both DEMTOP and CPINNTO also leverage DEM, their \textit{nested} optimization strategies result in significantly longer runtimes, each exceeding $1000$ seconds. This comparison underscores the computational efficiency of our \textit{simultaneous} and mesh-free framework relative to more complex \textit{nested} ML-based optimization pipelines such as DEMTOP and CPINNTO. We recognize that SIMP remains the most efficient method for linear problems, but expect the advantages of our framework to be more pronounced in nonlinear TO, where FEM-based approaches are substantially more expensive due to the iterative solution of state variables. This will be the focus of our future work.

We also note that our PIGP framework requires longer epochs than other approaches due to the nature of its simultaneous optimization. In contrast, the SIMP baseline and TOuNN employ FEM to directly solve displacements without iterative training, while DEMTOP and CPINNTO adopt \textit{nested} schemes that require hundreds of displacement solves per epoch. These additional sub-iterations with DEM substantially increase the computational cost of the latter two approaches.

\section{Conclusions and Future Works} \label{sec conclusion}
In this work, we propose a mesh-free physics-informed framework for density-based CM of solids. To parameterize the solution and design variables, our method employs independent GPs as priors with a shared multi-output NN as the mean function. We choose the PGCAN architecture as the mean function because of its ability to capture stress localization and fine-scale design features. By leveraging DEM, we reformulate the constrained CM problem with a \textit{nested} structure into a computationally efficient and robust \textit{simultaneous} and \textit{mesh-free} optimization scheme. Our approach minimizes the compliance objective subject to two penalty terms that ensure the force balance and volume fraction constraints. Notably, no explicit penalty terms are required for BCs or design constraints, as these are inherently satisfied by the GP priors.

During model training, we incorporate an adaptive grid scheme along with the FD method for gradient evaluation to mitigate overfitting and enhance computational efficiency. Additionally, we adopt a curriculum training strategy with scheduling to prevent the volume fraction constraint from promoting overly simple topologies.

We tested our framework using canonical 2D examples and compared our results against SIMP. Our findings show that the proposed PIGP approach: (1) achieves compliance comparable to SIMP and tends to yield a smaller gray area fraction; (2) provides improved feature resolution and faster convergence with PGCAN compared to MLP; (3) effectively controls structural complexity through the interpretable parameter $\Res$; (4) maintains grid-size independence without requiring a filtering technique; (5) can improve quality and continuity of topologies by leveraging GP sampling on a finer grid; and (6) offers higher computational efficiency compared to other ML-based approaches that do not rely on FEM solvers.
 
Currently, our framework relies on the FD method for spatial gradient estimation, which offers computational efficiency but suffers from limited accuracy. Additionally, a large number of CPs is required to evaluate volume integrals in the loss functions. These limitations become more pronounced in 3D problems where the number of CPs can increase by an order of magnitude. We have added preliminary 3D CM results in Appendix {\color{blue}G} for reference.} A promising direction for the future work is the incorporation of interpolation-based techniques, such as shape functions from FEM, into the PIGP framework to reduce the number of CPs while maintaining high numerical fidelity and computational efficiency. We will furhter pursue this direction in our future works. 
\begin{figure*}[!t]
    \centering
    \includegraphics[width = 0.4\textwidth]{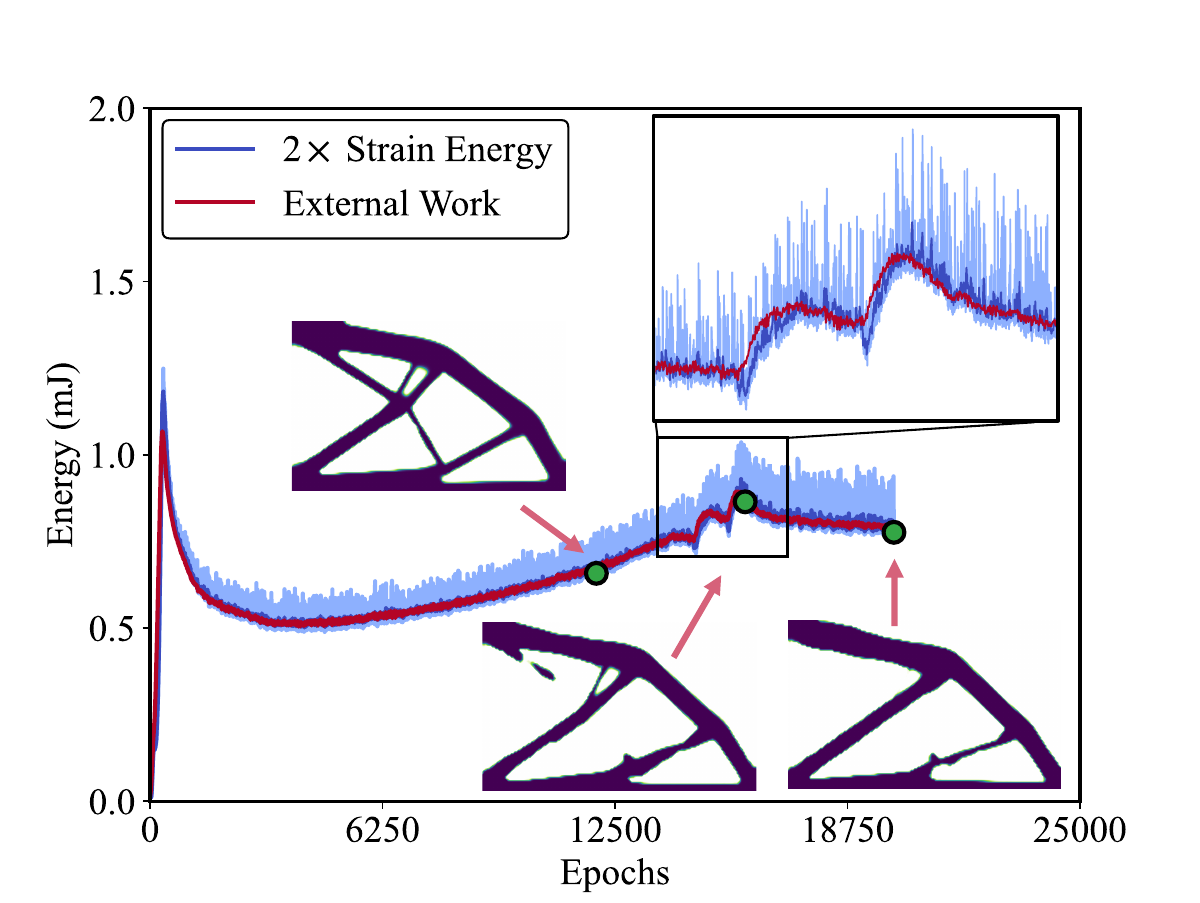}
    \caption{\textbf{Instability in the cantilever beam example:} Evolution histories of $2\times$strain energy and external work, with the inset highlighting their divergence, indicating a violation of the force balance constraint.}
    \label{fig instability}
\end{figure*}

During the implementation of our framework, we occasionally observed training instabilities as illustrated in \Cref{fig instability} using the cantilever beam example. At certain epochs, we observe an abrupt divergence between the curves of $2\times$strain energy and external work (highlighted in the inset figure), indicating a breakdown in force equilibrium. This instability leads to degraded structural solutions, characterized by disconnected regions and irregular surfaces, where finer topological features are lost. While these instabilities can be detected early in training and mitigated via reinitialization, we aim to prevent them altogether in our future works. 

\section*{Statements and Declarations} \label{sec declarations}

\textbf{Funding} We appreciate the support from the Office of the Naval Research (award number N000142312485), NASA’s Space Technology Research Grants Program (award number 80NSSC21K1809), and National Science Foundation (award number 2211908).

\noindent \textbf{Conflict of interest} The authors declare that they have no conflicts of interest.

\noindent \textbf{Author contributions} Xiangyu Sun: conceptualization, methodology, formal analysis and investigation, and writing—original draft; Amin Yousefpour: conceptualization, methodology, and writing—
review and editing; Shirin Hosseinmardi: conceptualization, methodology, and writing—
review and editing; Ramin Bostanabad: supervision, funding acquisition, conceptualization, and writing—
review and editing.

\noindent \textbf{Ethics approval and Consent to participate} Not applicable.

\noindent \textbf{Data Availability} Data will be available on request.

\noindent \textbf{Replication of results} The codes of the model are accessible via \href{https://github.com/Bostanabad-Research-Group}{GitHub} upon publication. 

\section*{Appendix}

\appendix
\renewcommand{\thesection}{\Alph{section}} 

\section{Nomenclature}\label{appendix nomenclature}
Throughout this paper, we use regular, bold lowercase, and bold uppercase letters to denote scalars, vectors, and matrices, respectively. Vectors are assumed to be column. We distinguish between a function and its evaluations by explicitly showing functional dependence. For instance, $f(\xb)$ and $\fb(\xb)$ represent scalar- and vector-valued functions, respectively. 

When dealing with vector functions, we use subscripts for differentiation. As an example, the vector function of displacement $\ub(\cdot)$ has two components denoted as $\ub(\xb) = [u_1(\xb),u_2(\xb)]^{\text{T}}$. All functions are assumed to support batch evaluations; that is, a function $\fb(\cdot)$ with $n$ inputs and $m$ outputs, applied to a matrix $\Xb$ of $n$ inputs (columns), returns a same-length matrix with $m$ outputs (columns), i.e., $\Ub = \fb(\Xb)$. Key symbols and their definitions are provided in \Cref{tab nomenclature} with more symbols specified in each section.
\begin{table*}[!h]
    \centering
    \renewcommand{\arraystretch}{1.5}
    \small
    \setlength\tabcolsep{8pt}
    \begin{tabular}{l|l} 
    \hline
    \textbf{Symbols} & \textbf{Description} \\ 
    \hline
    $\thetab$ & Set of all trainable NN parameters  \\ 
    $\xb$ & Coordinates of a spatial point \\ 
    $\Xb$ & A set of CPs \\ 
    $\rho(\xb)$ & Density function \\ 
    $\ub(\cdot)$, $u_1(\cdot)$, $u_2(\cdot)$ & Displacement functions \\ 
    $\tilde{\ub}_1$, $\tilde{\ub}_2$, $\tilde{\rhob}$& BCs for displacement and density \\ 
    $\sigmab(\xb)$, $\epsilonb(\xb)$ & Stress and strain tensors \\ 
    $\hat{\epsilonb}(\xb)$ & Strain with no gradients \\ 
    $\fb(\xb)$ & External force \\ 
    $\Cb(\xb)$ & Elastic stiffness tensor \\ 
    $E_{\max}$, $E_{\min}$ & Young's modulus for solids and void \\ 
    $\nu$ & Poisson's ratio \\ 
    $p$ & Penalization factor \\ 
    $\Omega$, $\Omega_{\ub}$, $\Omega_F$ & Design domain and boundaries\\ 
     $\alpha_0$, $\alpha_1$ & Weight factors for loss terms\\ 
    \hline
    \end{tabular}
    \caption{\textbf{Summary of key notations:} More symbols and notations are given in each section.}
    \label{tab nomenclature}
\end{table*}

\section{Numerical Differentiation with Finite Difference Method} \label{appendix ND}
We use the FD method on a structured grid to compute spatial derivatives \Citep{nocedal_numerical_2006}. Taking the horizontal displacement $u_1(\xb)$ as an example, the gradients at the query point $[x_i, y_j]$ using the FD method can be written as:
\begin{subequations}\label{eq CD}
    \begin{align}
        \frac{\partial u_1(x_i,y_j)}{\partial x} &\approx \frac{u_1(x_{i+1},y_j) - u_1(x_{i-1},y_j)}{2dx}, \label{eq CD 1} \\
        \frac{\partial^2 u_1(x_i,y_j)}{\partial x^2} &\approx \frac{u_1(x_{i+1},y_j) - 2u_1(x_i,y_j) + u_1(x_{i-1},y_j)}{dx^2}, \label{eq CD 2} \\
        \frac{\partial^2 u_1(x_i,y_j)}{\partial x \partial y} &\approx \frac{ u_1(x_{i+1},y_{j+1}) - u_1(x_{i+1},y_{j-1}) - u_1(x_{i-1},y_{j+1}) + u_1(x_{i-1},y_{j-1}) }{4dx\,dy}, \label{eq CD 3}
    \end{align}
\end{subequations}
where $dx$ and $dy$ are the intervals between grid points in the $x$- and $y$-directions, respectively, and subscripts $i\in[1,N_x], j \in[1,N_y]$ denotes adjacent points in a structured grid. Similar formulations can be derived for the partial derivatives with respect to $y$.

We can estimate the residuals of the governing PDEs using \Cref{eq CD}. To begin with, the constitutive relations in \Cref{eq constitutive} can be explicitly written as:
\begin{subequations}\label{eq constitutive2}
    \begin{align}
        \sigma_{xx}(\xb) &= \frac{E(\xb)}{1 - \nu^2} \big( \varepsilon_{xx}(\xb) + \nu \varepsilon_{yy}(\xb) \big), \label{eq constitutive2 1} \\
        \sigma_{yy}(\xb) &= \frac{E(\xb)}{1 - \nu^2} \big( \nu \varepsilon_{xx}(\xb) + \varepsilon_{yy}(\xb) \big), \label{eq constitutive2 2} \\
        \sigma_{xy}(\xb) &= \frac{E(\xb)}{1 + \nu} \varepsilon_{xy}(\xb), \label{eq constitutive2 3}
    \end{align}
\end{subequations}
where $\sigma_{xx}(\xb)$, $\sigma_{yy}(\xb)$, $\sigma_{xy}(\xb)$, $\varepsilon_{xx}(\xb)$, $\varepsilon_{yy}(\xb)$, and $\varepsilon_{xy}(\xb)$ are the independent components for the stress and strain tensors.

Using \Cref{eq kinematic} and \Cref{eq constitutive2}, we can rewrite the governing equation in \Cref{eq pde} as two PDE residuals:
\begin{subequations}\label{eq residuals}
    \begin{align}
        R_{\text{PDE1}}(\xb) &= \frac{E(\xb)}{1 - \nu^2} \frac{\partial^2 u_1(\xb)}{\partial x^2} 
        + \frac{E(\xb)}{2(1 + \nu)} \frac{\partial^2 u_1(\xb)}{\partial y^2} 
        + \left[ \frac{E(\xb) \nu}{1 - \nu^2} + \frac{E(\xb)}{2(1 + \nu)} \right] \frac{\partial^2 u_2(\xb)}{\partial x \partial y}, \label{eq residuals 1} \\
        R_{\text{PDE2}}(\xb) &= \frac{E(\xb)}{2(1 + \nu)} \frac{\partial^2 u_2(\xb)}{\partial x^2} 
        + \frac{E(\xb)}{1 - \nu^2} \frac{\partial^2 u_2(\xb)}{\partial y^2} 
        + \left[ \frac{E(\xb) \nu}{1 - \nu^2} + \frac{E(\xb)}{2(1 + \nu)} \right] \frac{\partial^2 u_1(\xb)}{\partial x \partial y}, \label{eq residuals 2}
    \end{align}
\end{subequations}
where the displacement gradient terms can be estimated with \Cref{eq CD}. The PDE losses ($L_{\text{PDE1}}$ and $L_{\text{PDE2}}$) are defined as the mean squared errors of the residuals in \Cref{eq residuals 1} and \Cref{eq residuals 2} on all CPs.

\section{Length-scale Parameter in Gaussian Processes} \label{appendix phi}

\begin{figure}[!h]
    \centering
    \begin{subfigure}[t]{0.7\textwidth}
        \captionsetup{justification=raggedright, singlelinecheck=false, skip=-1pt, position=top}
        \caption{$\phib_{\rho} = 3.0$}
        \includegraphics[width=\linewidth]{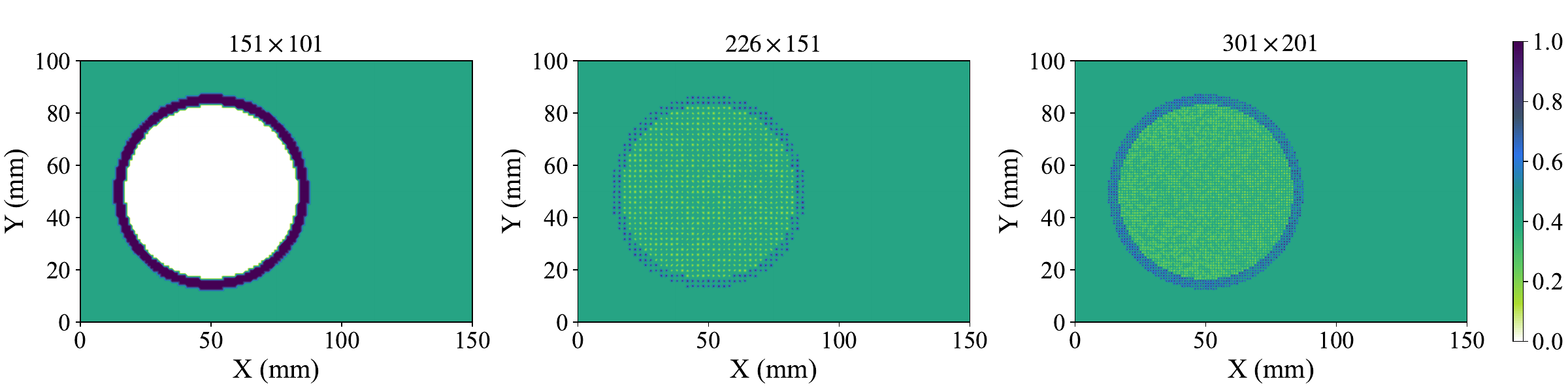}
        \label{fig length_scales 1}
    \end{subfigure}
    \vspace{-1em}

    \begin{subfigure}[t]{0.7\textwidth}
        \captionsetup{justification=raggedright, singlelinecheck=false, skip=-1pt, position=top}
        \caption{$\phib_{\rho} = 2.0$}
        \includegraphics[width=\linewidth]{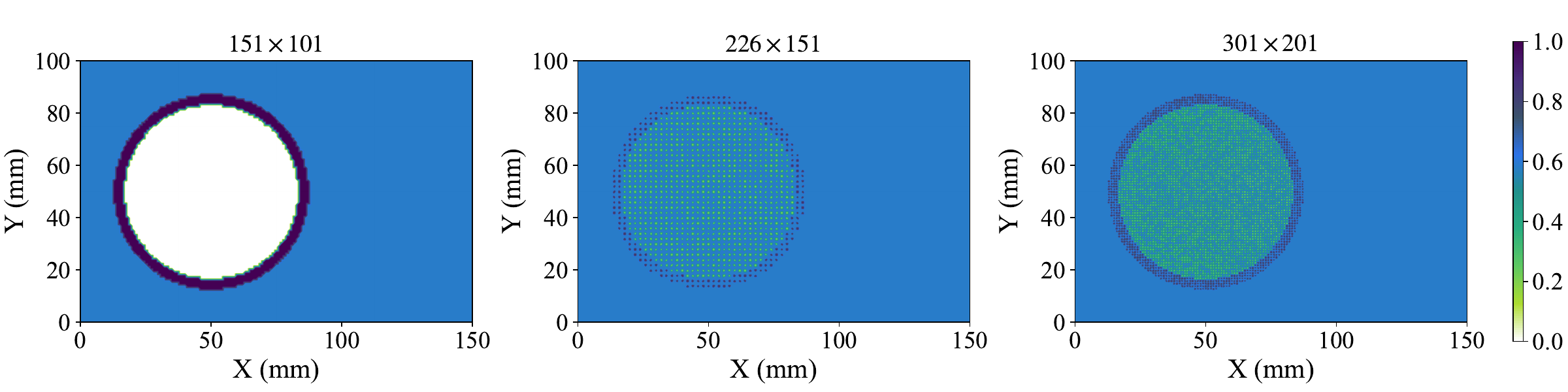}
        \label{fig length_scales 2}
    \end{subfigure}
    \vspace{-1em}

    \begin{subfigure}[t]{0.7\textwidth}
        \captionsetup{justification=raggedright, singlelinecheck=false, skip=-1pt, position=top}
        \caption{$\phib_{\rho} = 1.0$}
        \includegraphics[width=\linewidth]{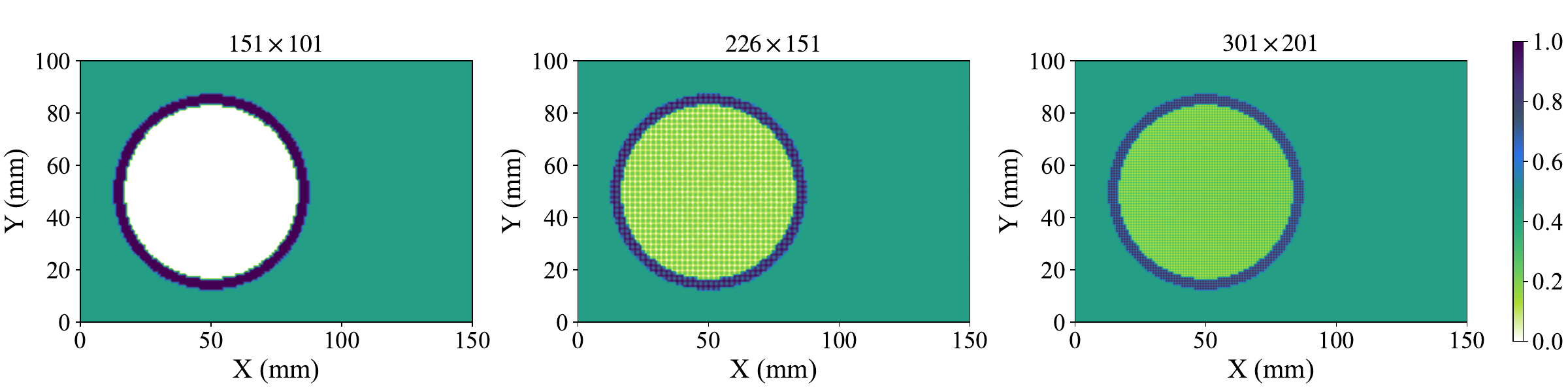}
        \label{fig length_scales 3}
    \end{subfigure}
    \vspace{-1em}

    \begin{subfigure}[t]{0.7\textwidth}
        \captionsetup{justification=raggedright, singlelinecheck=false, skip=-1pt, position=top}
        \caption{$\phib_{\rho} = 0.5$}
        \includegraphics[width=\linewidth]{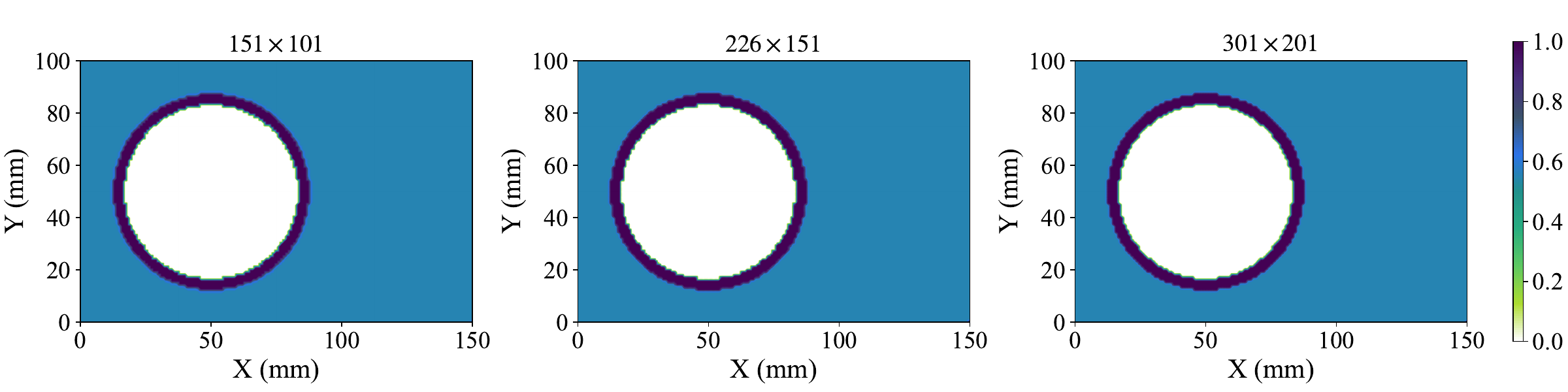}
        \label{fig length_scales 4}
    \end{subfigure}
    \caption{\textbf{Effect of length-scale parameter $\phib_{\rho}$ on density parameterization:} Density fields generated by a PGCAN model before training. The GP interpolations with different values of $\phib_{\rho}$ is applied to prescribed densities along the solid ring and the central void}
    \label{fig length_scales}
\end{figure}

In this study, we adopt a heuristic procedure to select the length-scale parameter $\phib_{\rho}$. As illustrated in \Cref{fig length_scales}, we parameterize the density field for the hollow beam structure consisting of a solid ring enclosing a central hole. The design constraints are enforced using GPs by varying $\phib_{\rho}$ from $3.0$ to $0.5$. The GP kernels are defined over a structured grid with $N_x = 151$ and $N_y = 101$.

We use PGCAN as the network architecture to predict the density fields on three structured grids: $(N_x, N_y) = (151, 101)$, $(226, 151)$, and $(301, 201)$, corresponding to the three columns from left to right in \Cref{fig length_scales}. No training is conducted in this analysis, and the predicted density field is evaluated through a single forward pass. The GP-enforced field is expected to satisfy the prescribed density constraints independently of the network parameters $\thetab$ or the grid resolutions.

As shown in \Cref{fig length_scales}, all four models successfully reconstruct the intended structure when evaluated on the coarse grid with the resolution of $(151, 101)$, which is expected, as the predictions with GP are conditioned on this grid. However, when evaluated on finer grids such as $(226, 151)$ or $(301, 201)$ with $\phib_{\rho}$ greater than $0.5$, regions that are expected to be fully solid ($\rho(\xb) = 1$) or entirely void ($\rho(\xb) = 0$) exhibit large deviations from the prescribed values. These inaccuracies are particularly problematic in the context of adaptive grid scheme, where CPs are dynamically sampled.

We can use small values of $\phib_{\rho}$ to mitigate this issue. From \Cref{fig length_scales 4}, we observe that the parameter $\phib_{\rho} = 0.5$ reliably reconstructs the desired structural features across all three grids. Since we define the grid resolutions consistently in \cref{tab grid}, and all GPs are sampled with the coarse grids, we set the length-scale parameter $\phib = 0.5$, and expect that the displacement BCs and design constraint can be naturally satisfied for all grids in the adaptive grid scheme.

\section{Effects from Adaptive Grids} \label{appendix overfitting}

\begin{figure*}[!h]
    \centering
    \begin{subfigure}[t]{0.35\textwidth}
        \includegraphics[width=\linewidth]{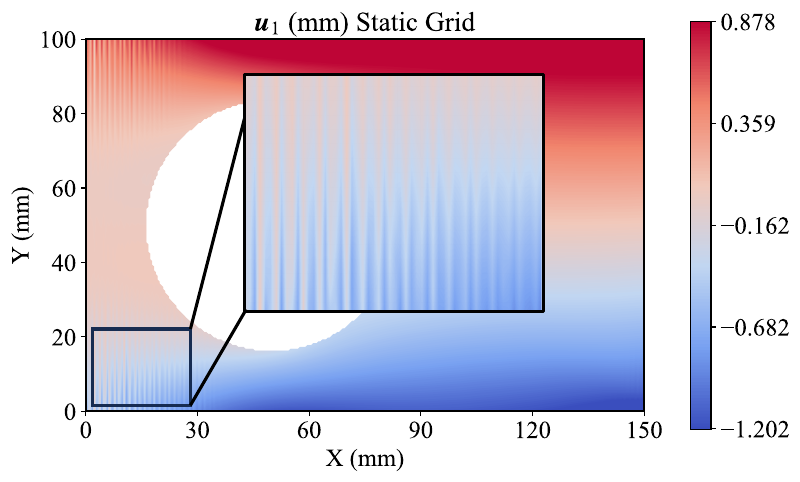}
        \vspace{-11em}
        \captionsetup{justification=raggedright, singlelinecheck=false, skip=-3.5pt, position=top}
        \caption[]{}
        \label{fig overfit 1}
    \end{subfigure}
    \begin{subfigure}[t]{0.35\textwidth}
        \includegraphics[width=\linewidth]{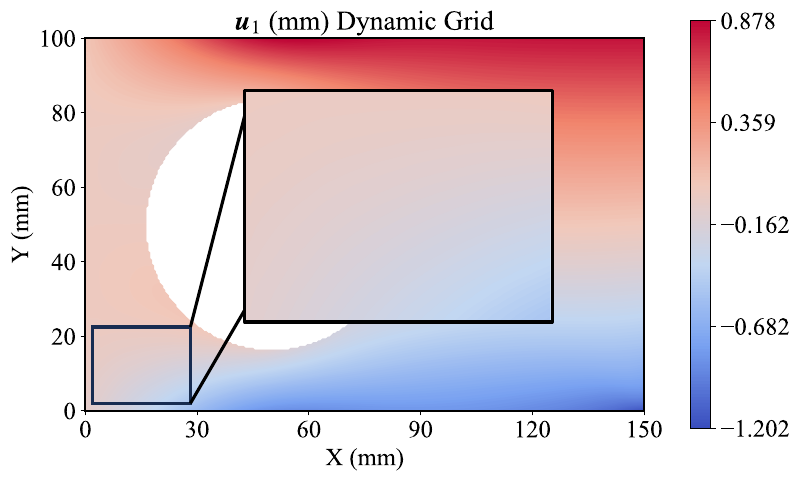}
        \vspace{-11em}
        \captionsetup{justification=raggedright, singlelinecheck=false, skip=-3.5pt, position=top}
        \caption[]{}
        \label{fig overfit 2}
    \end{subfigure}
    \begin{subfigure}[t]{0.35\textwidth}
        \includegraphics[width=\linewidth]{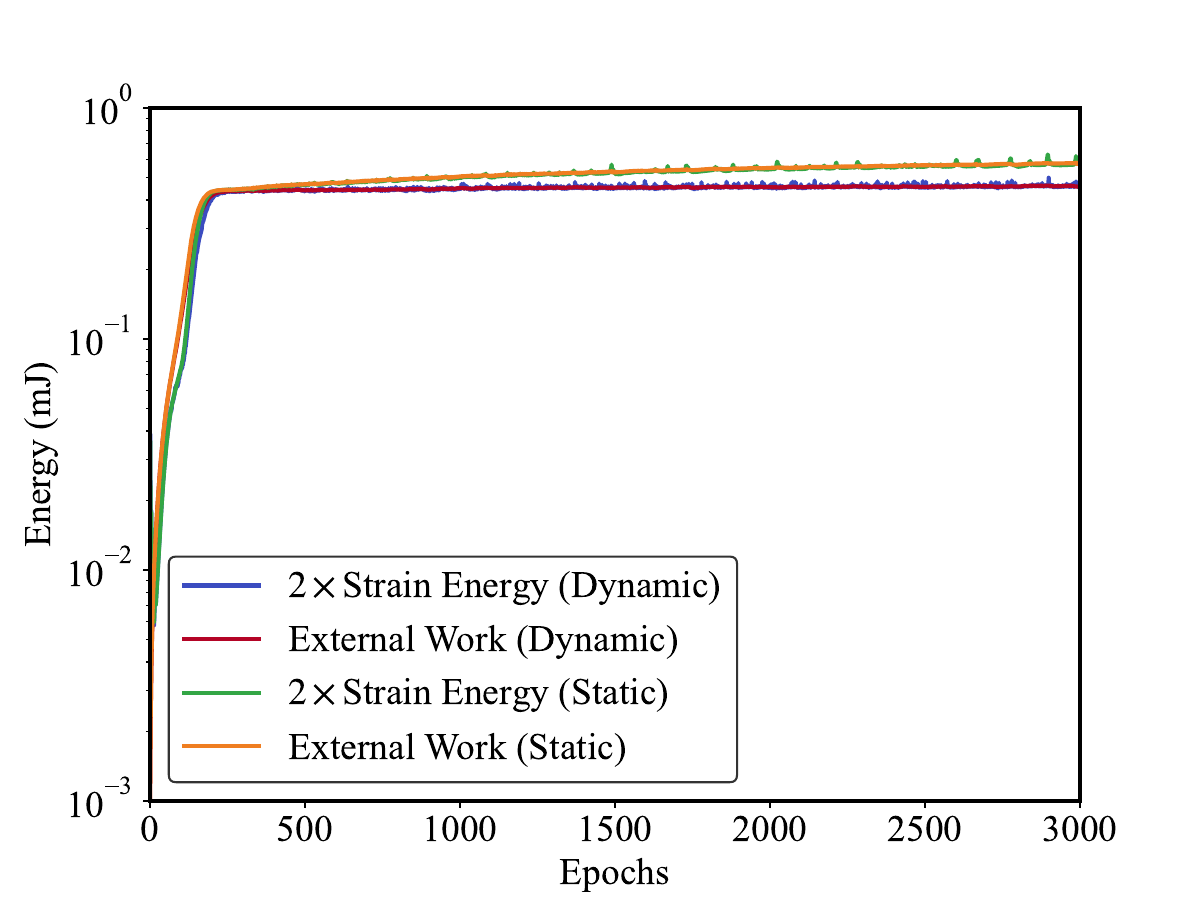}
        \vspace{-12.5em}
        \captionsetup{justification=raggedright, singlelinecheck=false, skip=-3.5pt, position=top}
        \caption[]{}
        \label{fig overfit 3}
    \end{subfigure}
    \begin{subfigure}[t]{0.35\textwidth}
        \includegraphics[width=\linewidth]{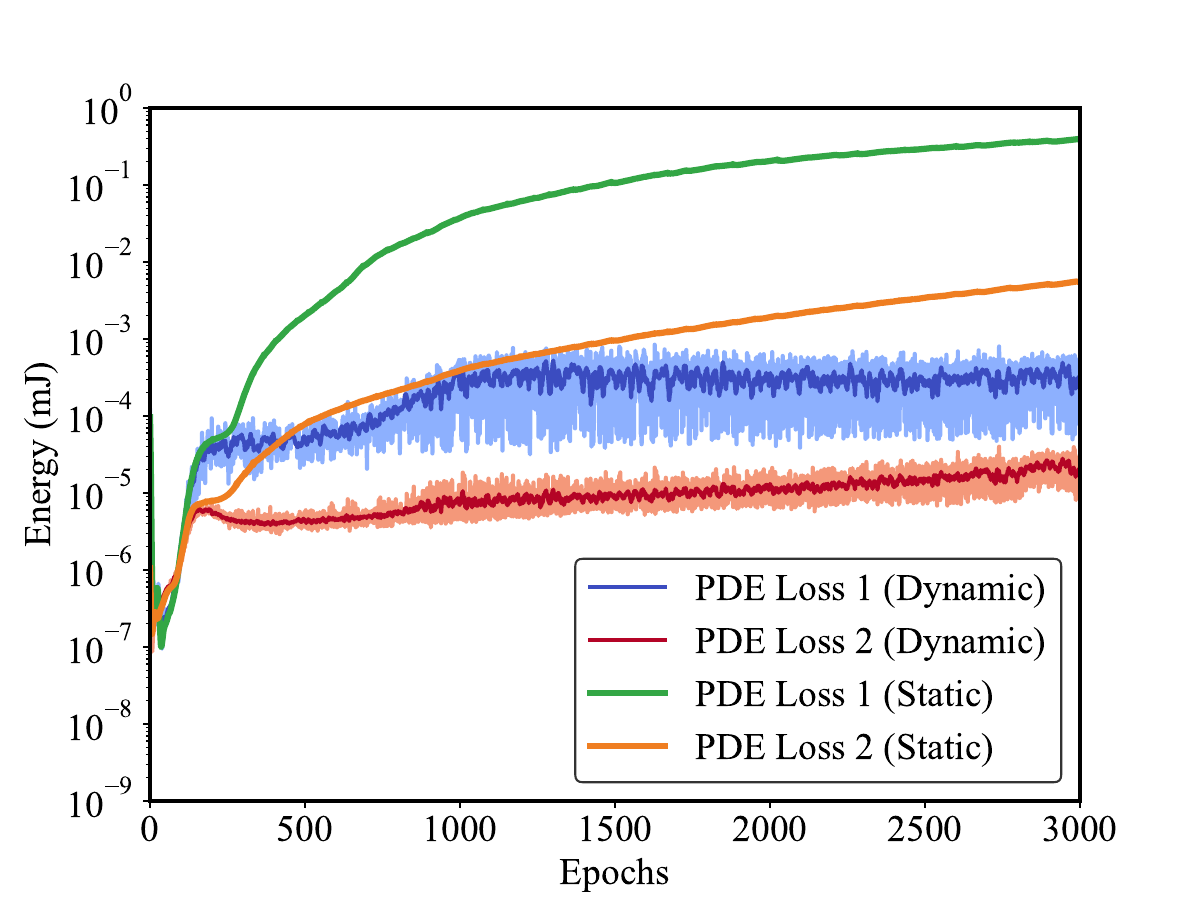}
        \vspace{-12.5em}
        \captionsetup{justification=raggedright, singlelinecheck=false, skip=-3.5pt, position=top}
        \caption[]{}
        \label{fig overfit 4}
    \end{subfigure}
    \caption{\textbf{Comparison of DEM results using PGCAN with static and adaptive grid schemes:} (a) horizontal displacement distribution obtained using the static grid, (b) the corresponding distribution computed with the adaptive grid scheme, (c) evolution histories of strain energy and external work for both schemes, and (d) loss histories of the PDE residuals. A zoomed-in region in (a) highlights the overfitting issue.}
    \label{fig overfit}
\end{figure*}

The model with PGCAN must be trained using an adaptive grid scheme to mitigate overfitting. \Cref{fig overfit 1} shows the horizontal displacement $\ub_1$ obtained from training with CPs placed on a fixed structured grid, where overfitting appears as vertical stripe patterns as highlighted in the inset. These stripe features exhibit high local gradient magnitudes. As the comparison, the horizontal displacement field obtained from the adaptive grid scheme in \Cref{fig overfit 2} is smoother and more physically reasonable.

Our proposed adaptive training strategy also improves robustness in the overall training dynamics. As shown in \Cref{fig overfit 3}, both training schemes reach force equilibrium around the 200\textsuperscript{th} epoch, while the adaptive grid scheme has the two curves (the $2\times$strain energy and external work) closely aligned and stabilized. However, the static grid exhibits a growing trend of these two curves, indicating the progressive overfitting. This divergence behavior is further illustrated by the loss histories of the two PDE residuals\footnote{Additional details on computing PDE residuals are provided in \Cref{appendix ND}.} shown in \Cref{fig overfit 4}, where the adaptive grid remains stable with low residuals on the order of $10^{-4}$, while those in the static grid case steadily increase, leading to significant PDE errors.

\section{Displacement Comparisons} \label{appendix displacement}

\begin{figure}[!h]
    \centering
    \begin{subfigure}[t]{\textwidth}
        \captionsetup{justification=raggedright, singlelinecheck=false, skip=-1pt, position=top}
        \caption{Horizontal displacement}
        \includegraphics[width=\linewidth]{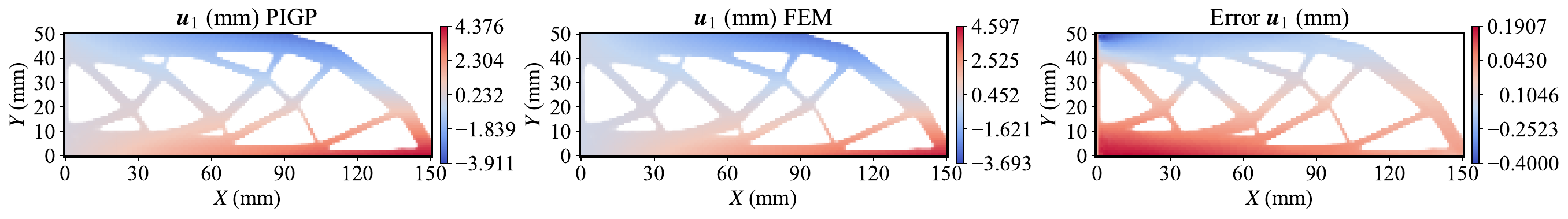}
        \label{fig u 1}
    \end{subfigure}
    \vspace{-2em}

    \begin{subfigure}[t]{\textwidth}
        \captionsetup{justification=raggedright, singlelinecheck=false, skip=-1pt, position=top}
        \caption{Vertical displacement}
        \includegraphics[width=\linewidth]{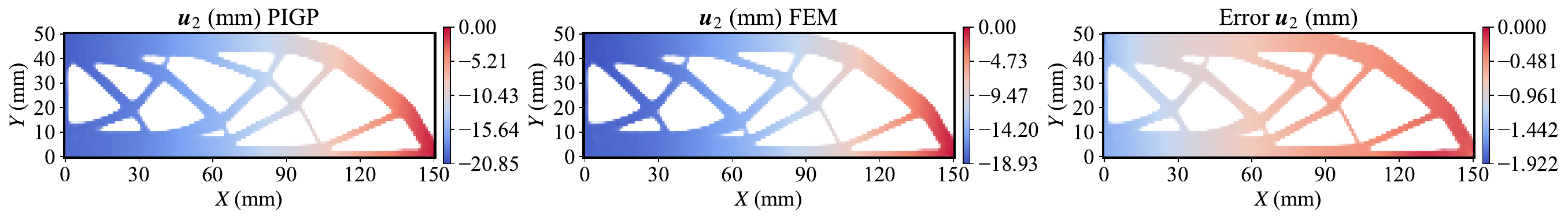}
        \label{fig u 2}
    \end{subfigure}
    \vspace{-1em}
    \caption{\textbf{Comparison of displacement fields obtained from PIGP and FEM for the MBB example:} 
    (a) horizontal displacement $\boldsymbol{u}_1$ and its corresponding error distribution. 
    (b) vertical displacement $\boldsymbol{u}_2$ and its corresponding error distribution.}
    \label{fig u}
\end{figure}

To assess the effectiveness of DEM in resolving displacement fields during TO, we compared the distributions of the two displacement components (horizontal $u_1(\xb)$ and vertical $u_2(\xb)$) obtained from our PIGP framework against FEM results using the same optimized topology as shown in \Cref{fig u}. The relative error was defined as the direct subtraction of each PIGP displacement component from its FEM counterpart. Overall, the displacement fields from PIGP closely match those from FEM, with maximum errors of about $10\%$ of the peak displacement magnitudes. The observed discrepancies mainly arise from two factors: (1) displacement gradients are evaluated using FD in PIGP, whereas FEM employs shape functions, and (2) the loss function integrals are computed via Monte Carlo sampling in PIGP versus Gauss quadrature in FEM. Preliminary tests suggest that incorporating shape function interpolation in our PIGP framework improves gradient accuracy and numerical integration, thereby reducing displacement errors. This enhancement will be further explored in our future work.    

\begin{figure}[!h]
    \centering
    \begin{subfigure}[t]{0.49\textwidth}
        \captionsetup{justification=raggedright, singlelinecheck=false, skip=-1pt, position=top}
        \caption{Horizontal displacement with high resolution:}
        \includegraphics[width=\linewidth]{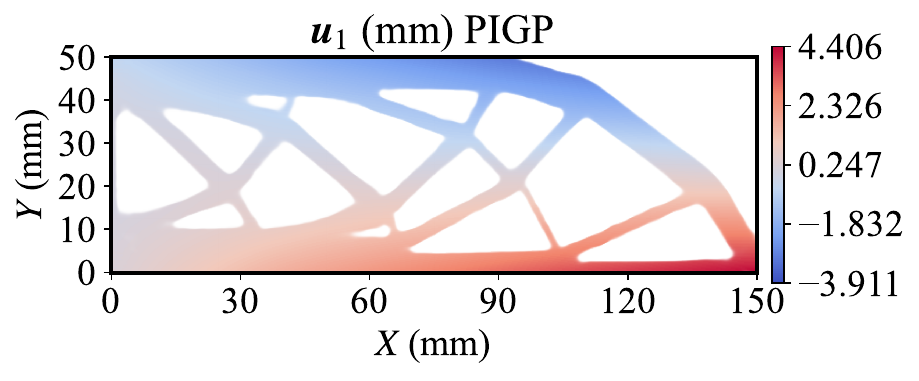}
        \label{fig high res 1}
    \end{subfigure}
    \vspace{-1em}
    \begin{subfigure}[t]{0.49\textwidth}
        \captionsetup{justification=raggedright, singlelinecheck=false, skip=-1pt, position=top}
        \caption{Vertical displacement with high resolution:}
        \includegraphics[width=\linewidth]{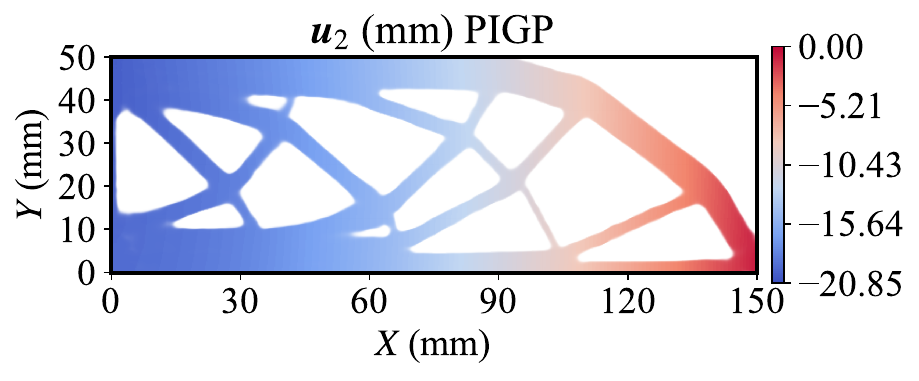}
        \label{fig high res 2}
    \end{subfigure}
     \vspace{-1em}
    \caption{\textbf{Super-resolution from PIGP:} High-resolution displacement fields on the optimized MBB beam topology with (a) horizontal and (b) vertical displacements.}
    \label{fig high res}
\end{figure}

We note that the interfaces in the PIGP topologies shown in \Cref{fig u} appear rough because the predictions are made on a coarse grid to match the density resolution of SIMP. In contrast, our PIGP framework naturally enables super-resolution by predicting the same displacement fields on a much denser grid, as illustrated in \Cref{fig high res}. This results in significantly smoother interfaces, which would be computationally expensive to achieve with SIMP on such fine grids.

\section{Interface Fraction} \label{appendix interface}

The interface fraction is defined as the ratio of the number of interface pixels to the total number of solid pixels. This metric is inherently sensitive to the grid resolution. To be consistent, we interpolate the predicted density fields onto their respective coarse grids and then estimate the interface fractions.

\begin{figure}[!h]
    \centering
    \begin{subfigure}[t]{0.55\textwidth}
        \captionsetup{justification=raggedright, singlelinecheck=false, skip=-1pt, position=top}
        \caption{MBB beam}
        \includegraphics[width=\linewidth]{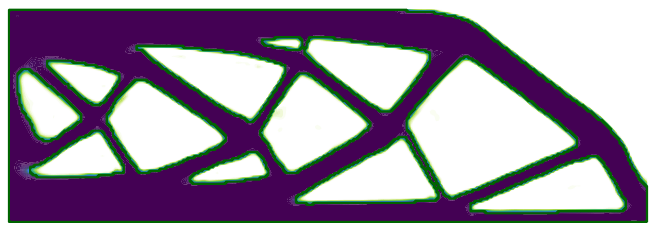}
        \label{fig interface 1}
    \end{subfigure}
    \vspace{-1em}
    \begin{subfigure}[t]{0.3\textwidth}
        \captionsetup{justification=raggedright, singlelinecheck=false, skip=-1pt, position=top}
        \caption{Cantilever beam}
        \includegraphics[width=\linewidth]{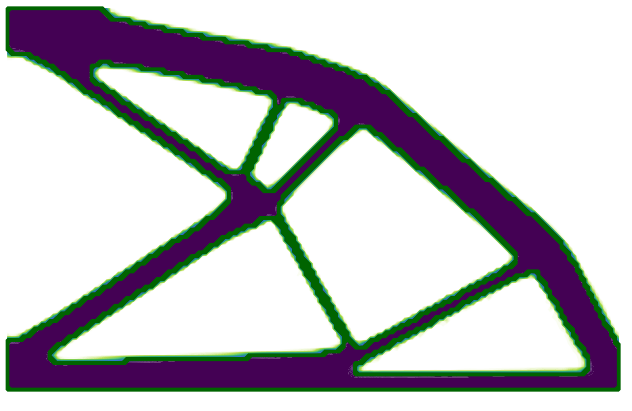}
        \label{fig interface 2}
    \end{subfigure}
    \caption{\textbf{Interface extraction in final topologies:} Examples from (a) the MBB beam and (b) the cantilever beam are visualized with the interface pixels highlighted in green.}
    \label{fig interface}
\end{figure}

To quantify the interface fraction, we apply image processing techniques to the final topologies of all examples. We first binarize the topology using a thresholding technique to remove any gray pixels. The threshold value is iteratively adjusted until the resulting binary structure satisfies the prescribed volume fraction constraint.
Morphological operations of erosion and dilation are then applied to the binarized structure to identify interface pixels (green) as illustrated in \Cref{fig interface} for the MBB beam and cantilever beam examples.

\section{Preliminary 3D Results} \label{appendix 3D}

\begin{figure}[!t]
    \centering
    \begin{subfigure}[t]{0.75\textwidth}
        \captionsetup{justification=raggedright, singlelinecheck=false, skip=-1pt, position=top}
        \caption{3D benckmark examples:}
        \includegraphics[width=\linewidth]{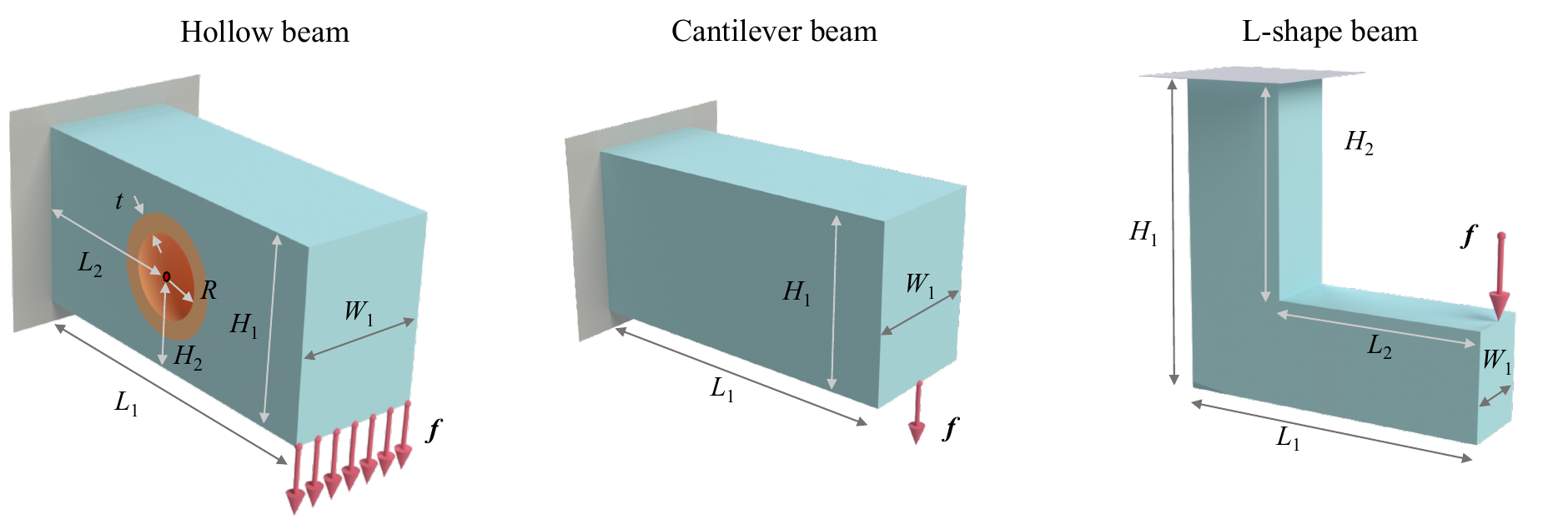}
        \label{fig 3D 1}
    \end{subfigure}
    \vspace{-1em}

    \begin{subfigure}[t]{0.75\textwidth}
        \captionsetup{justification=raggedright, singlelinecheck=false, skip=-1pt, position=top}
        \caption{Optimized topologies:}
        \includegraphics[width=\linewidth]{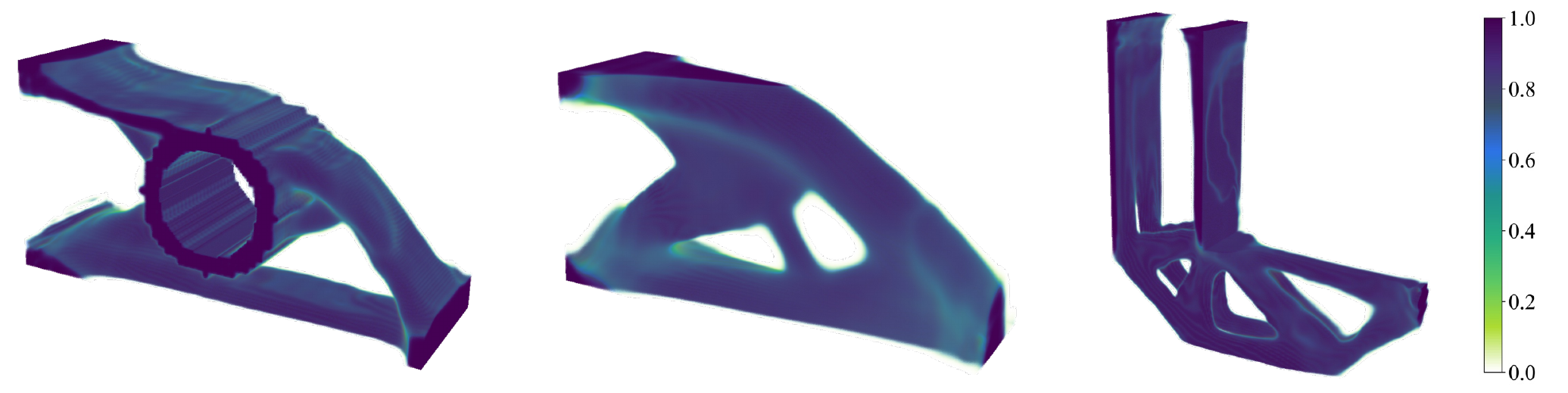}
        \label{fig 3D 2}
    \end{subfigure}
    \vspace{-1em}

    \caption{\textbf{3D CM with the proposed PIGP framework:} 
    (a) geometric schematics of the benchmark problems and 
    (b) the corresponding optimized topologies.}
    \label{fig 3D}
\end{figure}

We extended our PIGP framework to 3D CM benchmark problems. The geometric configurations are shown in \Cref{fig 3D 1}, and the corresponding parameters, including dimensions and applied external loads are summarized in \Cref{tab 3D}. For the hollow beam example, a solid cylinder is added around the hole as a design constraint. The same adaptive grid strategy and curriculum training for volume fraction are adopted. 

The optimized topologies are presented in \Cref{fig 3D 2}. The results demonstrate that design constraints can be enforced effectively using GPs. However, limited GPU memory restricts the number of GP sampling points, leading to rough surfaces along cylindrical holes. The optimized cantilever beam exhibits fewer structural features, likely due to the NN’s bias toward low-frequency patterns, even with PGCAN. In contrast, the L-shaped beam shows more complex internal structures than the cantilever beam.  

\begin{table*}[!h]
    \centering
    \renewcommand{\arraystretch}{1.5}
    \small
    \setlength\tabcolsep{8pt}
    \resizebox{\textwidth}{!}{%
    \begin{tabular}{l|c|c|c|c|c|c|c|c|c} 
    \hline
    \textbf{3D Example} & $H_1$ ($\mathrm{mm}$) & $L_1$ ($\mathrm{mm}$) & $W_1$ ($\mathrm{mm}$) & $H_2$ ($\mathrm{mm}$) & $L_2$ ($\mathrm{mm}$) & $t$ ($\mathrm{mm}$) & $R$ ($\mathrm{mm}$) & $f$ ($\mathrm{N}$ or $\mathrm{N/mm}$) & \textbf{$\psi_f$} \\ 
    \hline
    Hollow beam           & 30   & 60 & 20 & 15 & 30 & 3 & 7  & $0.01$       & 0.3 \\
    Cantilever beam    & 30  & 60 & 20 & --   & -- & -- & --   & $0.1$ & 0.2 \\
    L-shape beam       & 60 & 60 & 15  & 40  & 40 & -- & -- & $0.1$ & 0.2 \\
    \hline
    \end{tabular}
    }
    \caption{\textbf{Parameter values for the 3D CM benchmark examples in \Cref{fig 3D}:} Listed are the geometric dimensions, applied external forces, and target volume fractions for the five cases. All dimensions are given in $\mathrm{mm}$. The symbol $f$ denotes the external force magnitude, expressed in $\mathrm{N}$ for point loads and in $\mathrm{N/mm}$ for distributed loads.}
    \label{tab 3D}
\end{table*}

From the preliminary study, we find that extending the framework from 2D to 3D poses the following main challenges: (1) high computational cost due to the large number of variables (three displacement and six stress/strain components), (2) GPU memory demand, and (3) reduced accuracy of finite difference or Monte Carlo schemes for gradient and integral evaluations. Our ongoing modifications to the PIGP framework aim to address these issues.

\section{Projection and  Binarization} \label{appendix proj}

The projection function $P(\cdot)$, defined in \Cref{eq proj rho}, is employed to suppress intermediate densities and reduce the gray area. To ensure a fair comparison, we incorporated this projection function into the original SIMP implementation based on the 88-line MATLAB code \citep{andreassen_efficient_2011}. As noted in \cite{ferrari_new_2020}, several adjustments are necessary to obtain reasonable topologies: (1) the gradient of the projection function with respect to $\rho(\cdot)$ must be included in the sensitivity analysis; (2) the threshold density $\rho_t$ must be solved at each iteration using a Newton method to ensure the volume fraction constraint; and (3) a continuation scheme for the sharpness parameter $\beta$ is required, where $\beta$ starts at 2 and is doubled every 25 iterations until reaching a final value of $\beta = 8.0$. In our PIGP framework, we simply fix $\beta = 8.0$ for the entire training process rather than employing a continuation scheme.

\begin{figure}[!h]
    \centering
    \begin{subfigure}[t]{0.4\textwidth}
        \captionsetup{justification=raggedright, singlelinecheck=false, skip=-1pt, position=top}
        \caption{\textbf{SIMP} ($\Filter$ $2$) without $P(\cdot)$}
        \includegraphics[width=\linewidth]{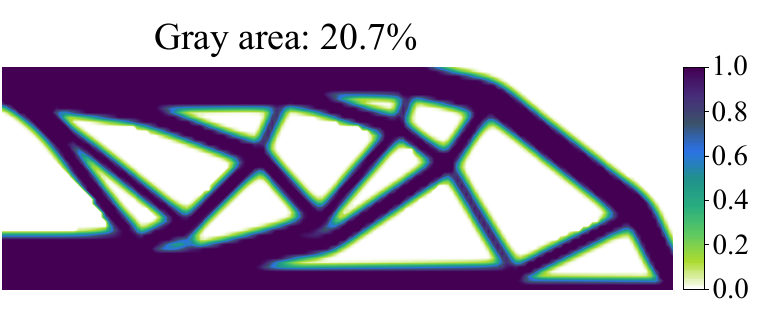}
        \label{fig proj 1}
    \end{subfigure}
    \vspace{-1em}
    \begin{subfigure}[t]{0.4\textwidth}
        \captionsetup{justification=raggedright, singlelinecheck=false, skip=-1pt, position=top}
        \caption{\textbf{SIMP} ($\Filter$ $2$) with $P(\cdot)$}
        \includegraphics[width=\linewidth]{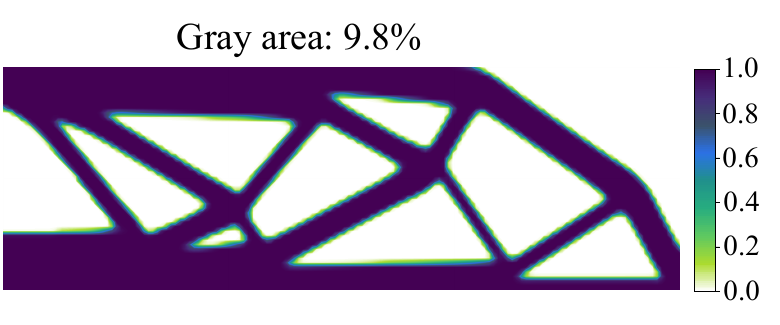}
        \label{fig proj 2}
    \end{subfigure}
    \caption{\textbf{Comparison of topologies using SIMP:} MBB beam topologies from \textbf{SIMP} ($\Filter$ $2$) (a) without the Heaviside projection function and (b) with projection.}

    \label{fig proj}
\end{figure}

The resulting topologies from SIMP ($\Filter$ $2$), with and without the projection function, are shown in \Cref{fig proj}. The projection function significantly reduces the gray area fraction from $20.7\%$ to $9.8\%$, demonstrating its effectiveness in eliminating intermediate densities.

\begin{table*}[!h]
    \centering
    \renewcommand{\arraystretch}{1.5}
    \small
    \setlength\tabcolsep{5pt}
    \caption{\textbf{Summary of compliance values:} Median compliance ($\mathrm{mJ}$) for the original and binarized topologies across the five examples, comparing \textbf{SIMP} ($\Filter$ $2$ and $4$) with \textbf{PIGP} ($\Res$ $36$ and $18$).}
    \begin{tabular}{l|cc|cc|cc|cc}
        \hline
        \multirow{2}{*}{\textbf{Example}}
        & \multicolumn{2}{c|}{\textbf{SIMP} ($\Filter$ $2$)} 
        & \multicolumn{2}{c|}{ \textbf{SIMP} ($\Filter$ $4$)} 
        & \multicolumn{2}{c|}{\textbf{PIGP} ($\Res$ $36$)} 
        & \multicolumn{2}{c}{ \textbf{PIGP} ($\Res$ $18$)} \\
        \cline{2-9}
        & Original & Binarized  
        & Original & Binarized 
        & Original & Binarized  
        & Original & Binarized  \\
        \hline
        MBB              &  1.891 &  1.848 &  1.994 &  1.909 &  1.924 &  1.873  &  1.949 &  1.892 \\
        Cantilever       &  0.712  &  0.695   &  0.739  &  0.704 &  0.709  &  0.695  &  0.736  &  0.724  \\
        Uniformly loaded &  0.284  &  0.245  &  0.448  &  0.255  &  0.243  &  0.238  &  0.247  &  0.261  \\
        L-shape          &  1.757  &  1.686  &  1.874  &  1.716  &  1.759  &  1.744  &  1.784  &  1.752  \\
        Hollow           &  0.528  &  0.523   &  0.538  &  0.526  &  0.530  &  0.523  &  0.535 & 0.524 \\
        \hline
    \end{tabular}
    \label{tab comp binary}
\end{table*}

To remove the influence of gray regions on compliance, all topologies are binarized using a dynamic thresholding algorithm where the threshold density is iteratively adjusted to achieve the target volume fraction. Compliance values before and after binarization are compared in \Cref{tab comp binary} for SIMP ($\Filter$ $2$ and $4$) and PIGP ($\Res$ $36$ and $18$). As shown in \Cref{tab comp binary}, compliance generally decreases slightly after binarization with the exception of SIMP ($\Filter$ $4$) in the Uniformly loaded beam where the original version has a large gray area fraction. Previous studies have reported that NN-based TO methods often show lower compliance before thresholding, whereas SIMP tends to achieve lower compliance after binarization \citep{sanu_neural_2024}. In our results, no clear trend is observed regarding which method yields lower compliance for binarized topologies. For example, PIGP ($\Res$ $36$) achieves a compliance of $0.238$ for the uniformly loaded beam, lower than SIMP ($\Filter$ $2$), while in the other examples the values are equal or higher. Overall, compliance values from binarized topologies remain very close across the two methods.

\section{Sensitivity of Grid Resolution} \label{appendix grid sensitivity}

\begin{figure*}[!h]
    \centering
    \includegraphics[width = 0.95\textwidth]{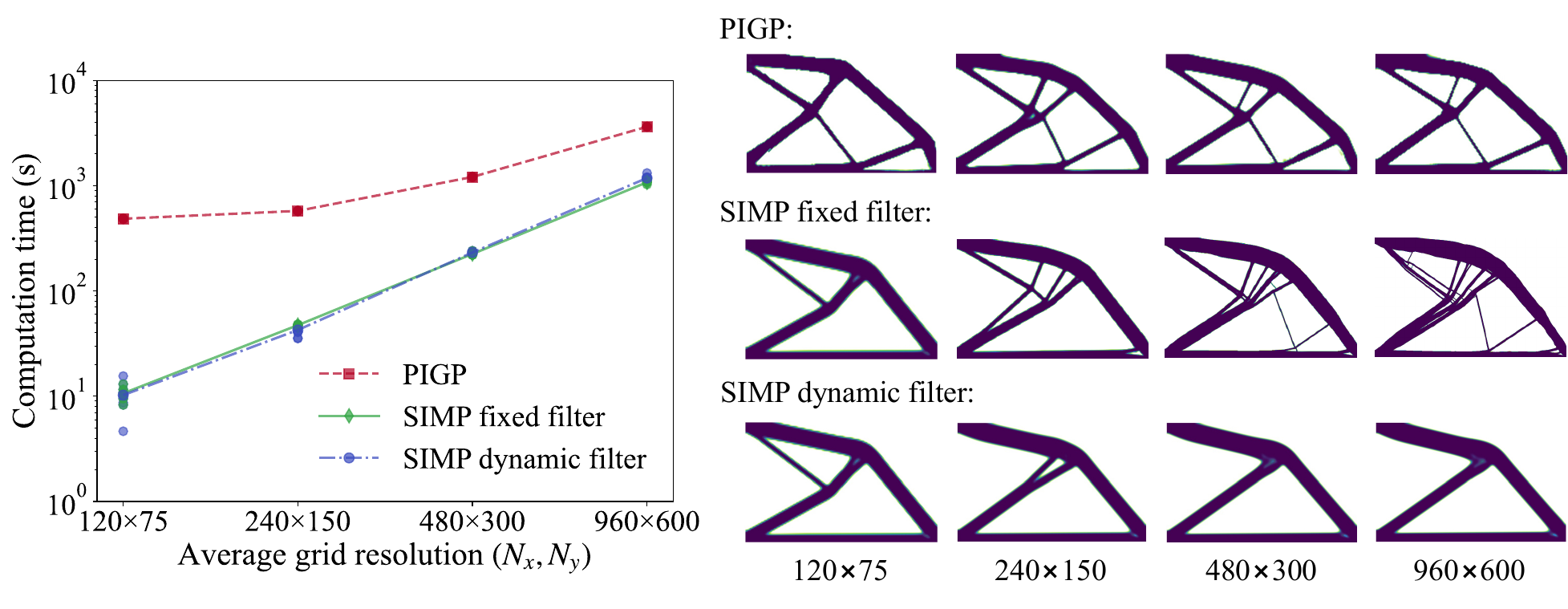}
    \caption{\textbf{Computational efficiency of PIGP compared with SIMP:} Total computation times for different average grid resolutions, together with the corresponding optimized topologies.}
    \label{fig mesh dep}
\end{figure*}

We evaluate the total computation time ($\sec$) and the sensitivity of optimized results to grid resolution using the cantilever beam example. As shown in \Cref{fig mesh dep}, the average grid resolution $(N_x, N_y)$ is refined from coarse $(120,75)$ to super fine $(960,600)$ for both our PIGP approach and the SIMP method. For SIMP, two cases are considered: (1) the filter radius fixed at $2$ for all mesh resolutions, and (2) the filter radius successively doubled from the coarse-grid value of $2$ up to $16$ for the finest grid. In contrast, our PIGP framework consistently uses $\Res = 36$. The total computational times for PIGP and both SIMP cases are summarized in \Cref{fig mesh dep}.

The results demonstrate that PIGP topologies are largely insensitive to adaptive grid resolution, consistent with the observations in \Cref{fig grid 1}.
In contrast, SIMP with a fixed filter radius exhibits pronounced mesh dependence. Computation time comparisons further confirm that FEM-based SIMP remains more efficient than PIGP at all grid resolutions, although the gap narrows on finer meshes. This trend highlights the potential of our framework for large-scale problems, which is a key direction of our future work. Importantly, our goal is not to surpass SIMP in efficiency, but to establish a robust mesh-free TO framework capable of reliably solving benchmark compliance minimization problems.

\bibliographystyle{sn-mathphys-num}
\bibliography{References}
\end{document}